\documentclass{article} % For LaTeX2e

\usepackage{iclr2026_conference,times}
\usepackage{amsmath,amsfonts,bm}

\def\eqref#1{equation~\ref{#1}}
\def\1{\bm{1}}

\DeclareMathAlphabet{\mathsfit}{\encodingdefault}{\sfdefault}{m}{sl}
\SetMathAlphabet{\mathsfit}{bold}{\encodingdefault}{\sfdefault}{bx}{n}

\usepackage{hyperref}
\usepackage{url}
\usepackage{booktabs}
\usepackage{caption}
\usepackage{tcolorbox}
\usepackage{multirow}

\usepackage[dvipsnames]{xcolor}
\usepackage{graphicx}    % 插入图片
\usepackage{listings}    % 代码块
\usepackage{xcolor}      % 代码高亮配色
\usepackage{enumitem}
\usepackage{float}
\usepackage{longtable}

\title{OSWorld-MCP: Benchmarking MCP Tool \\ Invocation In Computer-Use Agents}

\author{\textbf{Hongrui Jia}$^{1,2,\dagger}$, \quad \textbf{Jitong Liao}$^{2,\dagger}$, \quad \textbf{Xi Zhang}$^{2,\dagger}$, \quad \textbf{Haiyang Xu}$^{2,*}$, \quad \textbf{Tianbao Xie}$^{2}$, \\ \textbf{Chaoya Jiang}$^{1}$, \quad \textbf{Ming Yan}$^{2,*}$, \quad \textbf{Si Liu}$^{3}$, \quad \textbf{Wei Ye}$^{1,*}$, \quad \textbf{Fei Huang}$^{2}$ \\
$^1$ Peking University \qquad $^2$  Tongyi Lab, Alibaba Group \qquad $^3$  Beijing Zhongguancun Academy \\
{\tt\small\{jiahongrui, wye\}@pku.edu.cn, shuofeng.xhy@alibaba-inc.com} \\
}

\iclrfinaltrue
\begin{document}

\maketitle

\let\thefootnote\relax\footnotetext{\noindent$^\dagger$Equal contribution. $^*$Corresponding author.}

\begin{abstract}
With advances in decision-making and reasoning capabilities, multimodal agents show strong potential in computer application scenarios. Past evaluations have mainly assessed GUI interaction skills, while tool invocation abilities, such as those enabled by the Model Context Protocol (MCP), have been largely overlooked. Comparing agents with integrated tool invocation to those evaluated only on GUI interaction is inherently unfair.
We present OSWorld-MCP, the first comprehensive and fair benchmark for assessing computer-use agents' tool invocation, GUI operation, and decision-making abilities in a real-world environment. We design a novel automated code-generation pipeline to create tools and combine them with a curated selection from existing tools. Rigorous manual validation yields 158 high-quality tools (covering 7 common applications), each verified for correct functionality, practical applicability, and versatility.
Extensive evaluations of state-of-the-art multimodal agents on OSWorld-MCP show that MCP tools generally improve task success rates (e.g., from 8.3\% to 20.4\% for OpenAI
o3 at 15 steps, from 40.1\% to 43.3\% for Claude 4 Sonnet at 50 steps), underscoring the importance of assessing tool invocation capabilities. However, even the strongest models have relatively low tool invocation rates, Only 36.3\%, indicating room for improvement and highlighting the benchmark's challenge.
By explicitly measuring MCP tool usage skills, OSWorld-MCP deepens understanding of multimodal agents and sets a new standard for evaluating performance in complex, tool-assisted environments. Our code, environment, and data are publicly available at \href{https://osworld-mcp.github.io}{https://osworld-mcp.github.io}.

\end{abstract}

\section{Introduction}
Large Language Models (LLMs) such as GPT-5~\citep{gpt5}, DeepSeek-R1~\citep{guo2025deepseek}, and Qwen3~\citep{yang2025qwen3} have dramatically advanced reasoning and decision-making capabilities. Building on these advances, recent Large Multimodal Models (LMMs) are able to address complex computer-use tasks, which has stimulated considerable research interest~\citep{qin2025ui,lai2025computerrl,song2025coact,luo2025gui,lu2025uir1, lu2025uis1, ye2025mobile,wang2024mobile,wang2025mobile,zhu2025internvl3,wang2024mobile1}. Consequently, how to reliably and robustly benchmark different LMMs in GUI-driven scenarios has become a key open question. Existing evaluation frameworks ~\citep{xie2024osworld,abhyankar2025osworld,osworld_verified,bonatti2024windows,kuntz2025harm,rawles2024androidworld} primarily focus on assessing a model's ability to perform GUI-based operations, by predefining a set of user-interface actions (\textit{e.g., click, type}, and \textit{drag}) and 
allowing the model to autonomously decide how to complete the task.
% 现有的framework没有直接评估工具调用能力，重点放在了GUI操作上。只需要描述一下benchmark。

% Although numerous benchmarks have emerged for evaluating GUI agents, most of them overlook an important capability: the ability to invoke external tools, such as Model Context Protocol (MCP) ~\citep{anthropic_mcp_intro}. MCP is an open standard designed to connect AI applications to external systems. With MCP, computer-use agents can link to data sources (e.g., local files, databases), tools (e.g., search engines, calculators), and workflows (e.g., specialized prompts), enabling them to access critical information and perform tasks. Completing tasks via MCP is often more convenient. For example, as shown in \ref{fig:fig1}, when a computer-use agent receives the instruction "Please help me install the autoDocstring extension in VS Code", a GUI-based approach would require at least four steps, whereas the MCP tool can accomplish it in just one step — offering greater convenience and robustness. In fact, some agents~\citep{lai2025computerrl, song2025coact} have already incorporated autonomous tool invocation, achieving impressive results. However, directly comparing such agents against others in a benchmark that only measures GUI interaction is inherently unfair. At present, there is still a lack of comprehensive and fair benchmarks that assess GUI operation capabilities, tool invocation capabilities, and decision-making abilities of computer-use agents in an integrated manner.

\begin{figure*}[t]
    \centering
    \includegraphics[width=0.92\textwidth]{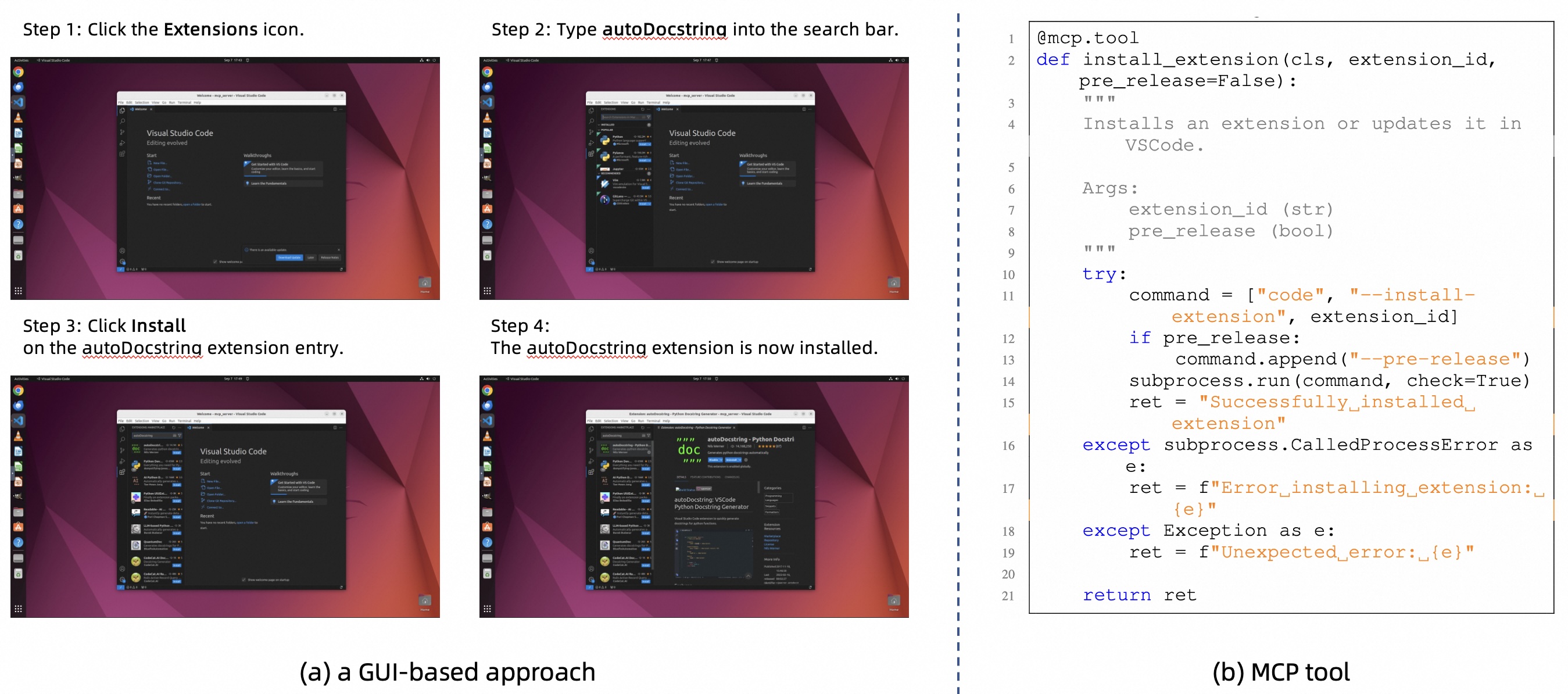}
    % \vspace{-1.6ex}
    \caption{Comparison of completing the instruction \textit{``Please help me install the autoDocstring extension in VS Code''} via GUI operations (a) and the MCP Tool (b).}
    % \caption{Comparison of completing the instruction \textit{``Please help me install the autoDocstring extension in VS Code''} via GUI operations and MCP Tool. (a): The illustration of GUI operations — at least four distinct actions. (b): The illustration of the MCP Tool code.}
     % the minimum number of steps required to accomplish this task via
     % for completing the same instruction
     % This comparison demonstrates that the MCP Tool offers a more convenient alternative to traditional GUI-based operations
    % \vspace{-4ex}
    \label{fig:fig1}
\end{figure*}

Although many benchmarks have been proposed for evaluating GUI agents, most neglect a crucial capability: the ability to invoke external tools such as the Model Context Protocol (MCP)~\citep{anthropic_mcp_intro}. MCP is an open standard that connects AI applications with external systems. By using MCP, computer-use agents can access diverse resources, including data sources such as local files and databases, and tools such as search engines and calculator,
% and workflows such as specialized prompts, 
thereby enabling them to obtain critical information and complete tasks more effectively.
In many cases, performing a task through MCP is more efficient than relying solely on GUI operations.
For instance, as illustrated in Figure~\ref{fig:fig1}, when an agent is instructed to install the autoDocstring extension in VS Code, a GUI-based approach may require at least four steps, whereas an MCP tool can achieve the same result in a single step, offering both greater efficiency and higher robustness.
Several recent agents~\citep{lai2025computerrl, song2025coact} have already incorporated autonomous tool invocation and have achieved notable performance gains.
However, it is inherently inequitable to compare such agents with others that assess only GUI interaction.
% However, comparing such agents with others using benchmarks that assess only GUI interaction is inherently inequitable.
At present, there remains a lack of comprehensive and equitable benchmarks that jointly evaluate GUI operation skills, tool invocation capabilities, and the decision-making competence of computer-use agents in an integrated framework.

% \begin{figure}[htbp]
%     \centering
%     % 左侧图片 0.45宽度
%     \begin{minipage}[t]{0\textwidth}
%         \centering
%         \includegraphics[width=\textwidth]{iclr2026/Fig1_left.jpg} % 换成你的图片路径
%         \caption{GUI}
%     \end{minipage}
%     \hfill
%     % 右侧代码 0.45宽度
%     \begin{minipage}[t]{0.6\textwidth}
%         \centering
%         \lstinputlisting[language=Python, caption={MCP Tool}]{iclr2026/install_extension_mcp_tool.py} % 也可以用 lstlisting 环境直接写代码
%     \end{minipage}
% \end{figure}

% To address this gap, we propose OSWorld-MCP, the first comprehensive benchmark that evaluating the tool invocation capabilities for computer-use agents.
To bridge this gap, we introduce OSWorld-MCP, the first comprehensive and fair benchmark designed to evaluate the tool invocation capabilities of computer-use agents.Our primary motivation is to establish a unified standard for fair comparison of tool utilization abilities across different models, addressing the current lack of consistency in tool sets and evaluation metrics.
%
% Specifically,
% our OSWorld-MCP is built up a widely used real-world computer-use benchmark OSWorld~\citep{xie2024osworld}.
% Based on the OSWorld, as shown in Figure~\ref{fig:fig6}, we introduce a curated set of \textbf{158 high-quality tools} covering 7 common applications such as LibreOffice and VS Code, ensuring a diverse and realistic testing environment.
% These tools can be applied to a total of 250 tasks in OSWorld, including 155 that involve challenging multi-round tool invocations, reflecting real-world complexity.
%
%
Built upon a widely used real-world computer-use environment OSWorld~\citep{xie2024osworld}, OSWorld-MCP significantly extends its capabilities by incorporating a curated set of \textbf{158 high-quality MCP tools}. 
These tools cover \textbf{7} common applications such as LibreOffice Writer and VS Code, ensuring a diverse and realistic testing environment (Figure~\ref{fig:fig6}). 
{A number of 25 non-target tools are included, serving as distractors in the OSWorld-MCP tasks.}
%
% {\color{red}Besides, our designed MCP Tools can be applied to a total of 69\% tasks, including 43\% tasks that involve challenging multi-round tool invocations, reflecting real-world complexity.} %证明tool的可用性%
Besides, our designed MCP tools are applicable to 250 tasks, accounting for 69\% of the entire benchmark, which underscores their broad applicability. Notably, 153 tasks, representing 42\% of the benchmark, involve challenging multi-round tool invocations. Even the strongest model, Claude 4 Sonnet, achieves an accuracy of 0 when relying solely on GUI operations for tasks requiring four rounds of tool invocation, and only 16.7\% accuracy when MCP tools are introduced. These results highlight the challenging of our benchmark and the broad utility of our tools in diverse scenarios.
% Besides, our designed MCP tools are applicable to 250 tasks (69\% of the entire benchmark), highlighting their broad utility.
% % demonstrate good versatility and usability, 
% Notably, 153 tasks (42\% of the total) involve challenging multi-round tool invocations,
% % reflecting not only real-world complexity but also the high versatility of our tools.
% highlighting the challenging of our benchmark and the broad utility of our tools in diverse scenarios.

%These issues underscore the comprehensiveness and quality of our MCP tools in addressing real-world computer-use scenarios.

% During the evaluation, a key feature of OSWorld-MCP is its dynamic interaction between GUI operations and tool usage.
Another distinguishing feature of OSWorld-MCP is its dynamic interaction between GUI operations and tool usage. 
Specifically, at every step of a task, the agent can autonomously choose between our MCP tools and direct GUI actions (\textit{e.g., click} and \textit{type}) to interact with the graphical interface.
% for directly interacting with the graphical interface.
%
% By integrating both GUI operations and MCP-based tool invocation, 
With this setting,
OSWorld-MCP can provide a balanced and thorough assessment of LMM capabilities in hybrid decision-making skills. 
% MCP tool invocation and
% GUI operations, with particular emphasis on 
Here, the decision-making involves not only selecting the correct tools, but also choosing the most efficient execution path when both GUI operations and MCP tools are required to accomplish the task.
Besides, to provide a more nuanced evaluation, we introduce two new metrics alongside task accuracy: 
% Tool Invocation Rate (TIR), measuring the proportion of tasks successfully completed using MCP tools, and Average Completion Steps (ACS), quantifying task completion efficiency. 
\textit{Tool Invocation Rate} (TIR) and \textit{Average Completion Steps} (ACS).
TIR measures the proportion of tasks successfully completed using MCP tools, offering insights into an agent's tool utilization propensity, while ACS quantifies task completion efficiency.
% , reflecting the agent's overall performance.
% These metrics offer deeper insights into an agent's tool utilization capabilities and overall task performance.
%
In conclusion,
compared with existing text-based tool-use benchmarks~\citep{liu2025mcpeval, mo2025livemcpbench, gao2025mcp, wang2025mcp} and the above GUI relevant benchmarks, OSWorld-MCP has significant advantages in evaluating real-world computer-use scenarios. 
% It requires agents to interpret visual GUI information, perform GUI operations, invoke appropriate tools, and demonstrate the ability to effectively chain multiple tools.
It challenges agents to interpret visual GUI information, perform GUI operations, invoke appropriate tools, and effectively chain multiple tools.
% – skills crucial in practical applications.
This combination of features makes OSWorld-MCP a more comprehensive and realistic benchmark for evaluating computer-use capabilities.

To develop the above-mentioned high-quality MCP tools, we design an automated code generation pipeline comprising three modules: the Code Generation Module, the Code Filter Module, and the Tool Wrap Module. By leveraging the advanced reasoning capabilities of OpenAI o3~\citep{openai_o3_o4mini_2025b}, this pipeline produces 72 functional tools. These tools are then combined with those curated from existing MCP servers~\citep{lai2025computerrl}, followed by a fine-grained manual verification procedure to remove functionally redundant items and highly task-specific ones that lack relevance to real-world applications. 
% This process results in a curated collection of 158 high-quality tools, each verified to be practically usable and consistent with the intended task difficulty.
Through this process, we obtain a curated collection of 158 high‑quality tools, each verified to be readily usable and aligned with the designed task difficulty.
With this comprehensive tool set and our OSWorld-MCP, 
we conduct a detailed experimental study and analysis on a range of state-of-the-art LMMs and multimodal frameworks. Three key findings are identified during experiments: 
(1) \textbf{MCP tools enhance agent accuracy and efficiency compared to the GUI-only setting.} For example, the success rate of OpenAI o3 increases from 8.3\% to 20.4\% at 15 steps.
(2) \textbf{The tool invocation rate positively correlates with performance.} We also observe that tool invocation rates for multimodal agents remain relatively low, indicating the significant potential in the tool utilization capabilities of current LMMs and multimodal frameworks. 
(3) \textbf{The composition of multiple tools remains a significant challenge.} Performance declines on complex task involving more tools. In the other hands, agent struggle on selecting the correct tool from a full list. 
% }

% with the best-performing LMM achieving a success rate of 45.0 \%. At the same time, we observe that multimodal agents tend to have relatively low tool invocation rates, which range from 9.5\% to 37.7\%. This indicates that the tool utilization capability of current LMMs and multimodal frameworks still has significant room for improvement.

Our contributions are as follows:

% \vspace{-0.8ex}
\begin{itemize}[leftmargin=1em, itemsep=0pt]

\item We introduce OSWorld-MCP, a comprehensive, fair, and novel benchmark for evaluating computer-use agents that integrates 158 high-quality MCP tools (covering 7 common apps) with GUI operations in real-world scenarios.
% Covering 7 common applications, these tools enable a realistic assessment of agent capabilities 
% integrating MCP tool invocation with GUI operations for evaluating computer-use agents.
It bridges the gap between pure-GUI and text-based tool-use evaluations, offering a holistic and realistic assessment of computer-use capabilities.
% It bridges pure-GUI and text-based evaluations, offering a holistic assessment of real-world computer-use capabilities.

% \item  {\color{blue} We introduce OSWorld-MCP, a comprehensive, fair, and novel dynamic benchmark that uniquely evaluates MCP tool invocation alongside GUI proficiency and decision-making. It bridges the gap between existing pure-GUI and text-based tool-use benchmarks, providing a more holistic and realistic assessment of computer-use capabilities.}

% \item We present OSWorld-MCP, a comprehensive and fair benchmark for evaluating multimodal agents on computer-use tasks. It uniquely combines the assessment of tool invocation with GUI proficiency and decision-making, bridging the gap between existing pure-GUI and text-based tool-use evaluations.

\item We propose a novel pipeline combining automated code generation with rigorous manual curation to create MCP tools, enhancing our benchmark's evaluation depth and fairness.
% the depth and breadth of our benchmark's evaluation capabilities while ensuring its fairness. 
We also introduce new metrics (\textit{i.e.}, TIR and ACS) for nuanced assessment of agents' tool utilization propensity.
% \vspace{-0.8ex}
% Besides, we propose new evaluation metrics (\textit{i.e.}, TIR and ACS), which provides a more nuanced assessment of an agent's tool utilization propensity.

\item Our extensive experiments indicate that (1) MCP tools improve agent metrics; (2) higher tool invocation correlates with higher accuracy; (3) combining tools introduces significant challenges.

\end{itemize}
% \vspace{-2ex}
% Our principal findings are as follows:

% \begin{itemize}
% \item Effective utilization of MCP tools markedly enhances the accuracy of large multimodal models in completing OSWorld tasks and reduces the average number of steps required for completion.

% \item Tool invocation rate and accuracy exhibit a positive correlation, while the tool invocation rate shows no clear relationship with the average number of completion steps.

% \item As task difficulty increases, MCP tools consistently enable large multimodal models to complete tasks with greater efficiency and accuracy. Nevertheless, integrating MCP tools imposes a greater challenge on these models than combining graphical user interface operations.
% \end{itemize}

% In summary, we introduce a comprehensive and fair benchmark for multimodal computer-use agents that incorporates both GUI operations and MCP tool invocation. In addition, we present novel insights to guide future research on intelligent computer-use systems.

\section{Related Work}
% \vspace{-1ex}
\subsection{Benchmarks for Multimodal Agents}
% \vspace{-1ex}
Current benchmarks~\citep{deng2023mind2web,lu2024weblinx,kapoor2024omniact,zhou2023webarena,koh2024visualwebarena,drouin2024workarena,tian2024mmina,bonatti2024windows,xie2024osworld} for multimodal agents primarily focus on evaluating their ability to complete tasks through GUI-based operations. Static benchmarks such as Mind2Web~\citep{deng2023mind2web}, WebLinx~\citep{lu2024weblinx}, and OmniAct~\citep{kapoor2024omniact} rely on manually collected static datasets to assess agent performance. These static benchmarks are built upon pre-defined trajectories of GUI actions, which make them incapable of evaluating alternative action paths that may arise when tool invocation is introduced. In contrast, dynamic interactive benchmarks operate in open-ended environments and provide reward signals upon task completion, enabling a more flexible and open-ended evaluation of agents. Notable examples of dynamic benchmarks for specific environments include WebArena~\citep{zhou2023webarena}, VisualWebArena~\citep{koh2024visualwebarena}, WorkArena~\citep{drouin2024workarena}, MMInA~\citep{tian2024mmina}, WindowsAgentArena~\citep{bonatti2024windows}, and OSWorld~\citep{xie2024osworld}. However, existing dynamic interactive benchmarks typically predefine only GUI actions for the agent to use, and therefore lack a comprehensive and fair evaluation framework that jointly measures multimodal agents' tool invocation, GUI interaction, and decision-making capabilities.

\subsection{Model Context Protocol}
% \vspace{-1ex}
With the rapid development of multimodal agents, increasing attention has been paid to their tool invocation capabilities~\citep{song2025coact,lai2025computerrl}. Introduced by Anthropic in November 2024, the Model Context Protocol (MCP) is a JSON-RPC–based client–server interface designed for secure context ingestion and structured tool invocation. MCP provides a standardized, model-agnostic interface that enables AI applications to connect to external systems such as tools, data resources, and workflows, thereby facilitating the integration of large language models with external data sources and utilities~\citep{anthropic_mcp_intro}. This standardization addresses challenges arising from fragmented and highly customized integrations. MCP supports flexible plug-and-play tooling, secure infrastructure integration, and cross-LLM vendor compatibility~\citep{ehtesham2025survey}. Several contemporary MCP-related benchmarks for LLM evaluation have recently emerged~\citep{gao2025mcp,liu2025mcpeval,mo2025livemcpbench,wang2025mcp}. MCPEval~\citep{liu2025mcpeval} and MCP-Radar~\citep{gao2025mcp} cover only a limited set of MCP servers, typically no more than a few dozen tools, which restricts task diversity. LiveMCPBench~\citep{mo2025livemcpbench} employs large language model–based evaluation, an approach that is not well suited to tasks requiring real-time knowledge.  MCP-Bench~\citep{wang2025mcp} defines its tasks based on available tools, introducing constraints that create a gap between benchmark tasks and truly open-ended real-world problems. Currently, there is no benchmark that comprehensively evaluates multimodal agents in terms of GUI interaction, tool invocation, and decision-making capabilities within an integrated and fair framework.

\section{OSWorld-MCP Bench}
% \vspace{-1ex}

\subsection{Overview} 

\begin{figure*}[t]
    \centering
    \includegraphics[width=0.9\textwidth]{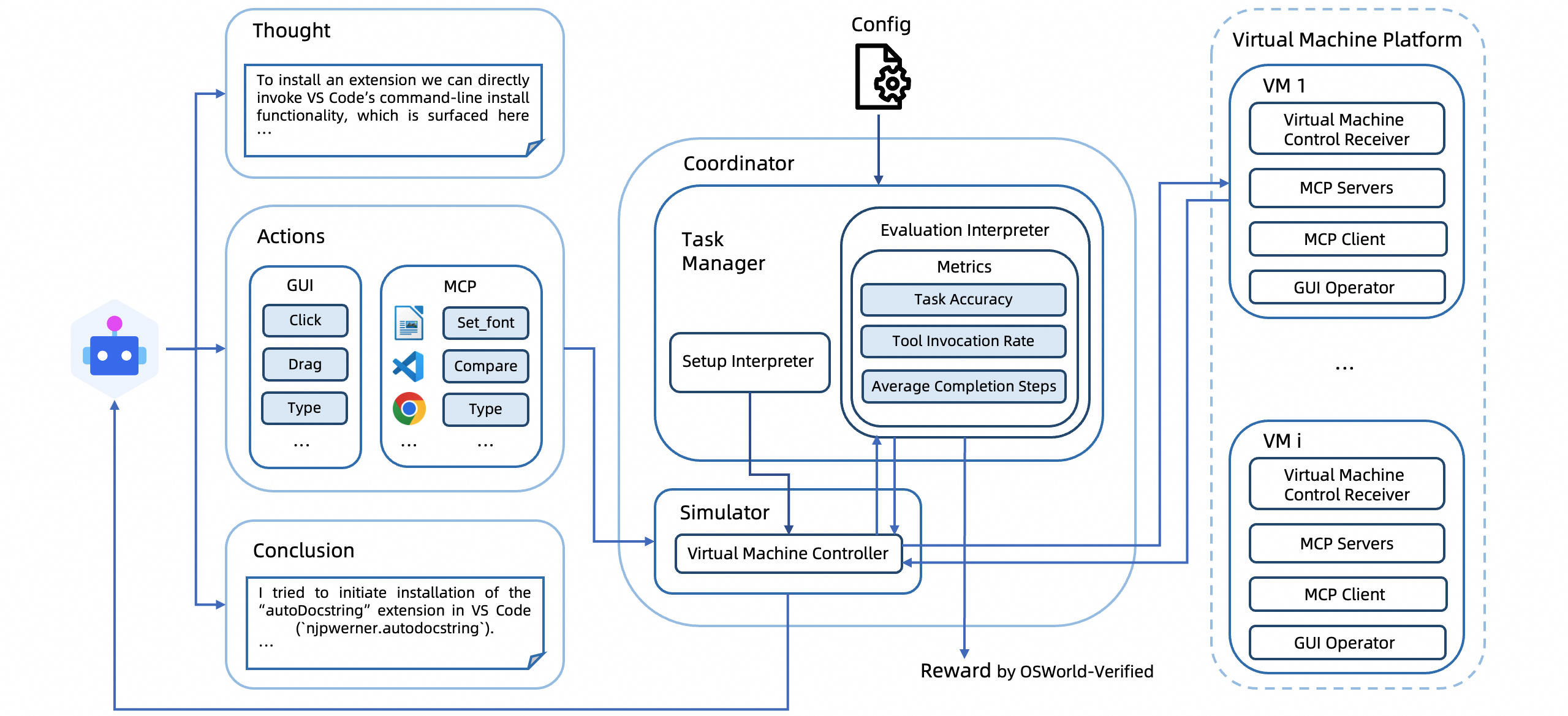}
    % % \vspace{-2ex}
    % \caption{Overview of the OSWorld-MCP framework. During evaluation, multimodal agents can autonomously decide at each step whether to use the most appropriate GUI action or invoke the most suitable MCP tool to advance task execution. We adopt OSWorld-Verified for automated verification of task outcomes, ensuring the accuracy and reliability of the evaluation.}
    \caption{Overview of the OSWorld-MCP framework.}
    % \vspace{-2ex}
    \label{fig:fig2}
    % \vspace{-2ex}
\end{figure*}

% We present OSWorld-MCP, a comprehensive benchmark for evaluating computer-use agents. OSWorld-MCP is built upon the OSWorld benchmark, a widely adopted dynamic and interactive evaluation suite for assessing multimodal agent performance in realistic computing environments (Ubuntu, Windows, macOS). OSWorld spans nine applications — Chromium, GIMP, LibreOffice Suite, OS, Thunderbird, VLC, and Visual Studio Code — and covers 369 real-world tasks, leveraging both Graphical User Interface (GUI) and Command Line Interface (CLI) interactions. As illustrated in Figure \ref{fig:fig2}, OSWorld-MCP extends OSWorld by complementing its GUI-operation evaluation with effective assessment of multimodal agents' tool invocation and decision-making capabilities. This includes the ability to decide between GUI and MCP actions, as well as selecting the most suitable MCP tool for a given task in real-world scenarios. In the following section, we first describe our automated tool generation pipeline for constructing the tools used in OSWorld-MCP. We then detail our tool collection and filtering process, ensuring the selection of high-quality tools for building the OSWorld-MCP dataset. Finally, we define the evaluation metrics of OSWorld-MCP and outline how these metrics enable a holistic evaluation of model performance.

We propose OSWorld-MCP, a comprehensive benchmark for evaluating computer-use agents. OSWorld-MCP is built upon the OSWorld benchmark, which is a widely used dynamic and interactive evaluation framework designed to assess the performance of multimodal agents in realistic computing environments, including Ubuntu, Windows, and macOS. OSWorld covers nine applications 
% (Chromium, GIMP, LibreOffice Suite, operating system utilities, Thunderbird, VLC, and Visual Studio Code)
and consists of a total of 369 real-world tasks that involve interaction through both graphical user interfaces (GUI) and command-line interfaces (CLI). As illustrated in Figure \ref{fig:fig2}, OSWorld-MCP enables effective assessment of multimodal agents in authentic scenarios, capturing their tool invocation capability, GUI operation skills, and decision-making competence. Decision-making assessment includes the ability to choose between GUI and MCP pathways, as well as the ability to select the most appropriate MCP tool for a given task. In the following section, we first introduce our automated pipeline for tool generation and the procedures for collecting and filtering tools used to construct OSWorld-MCP. We then present a detailed analysis of the MCP tools we produce, demonstrating the high quality and rational design of the OSWorld-MCP tool set. Finally, we describe the evaluation metrics defined for the OSWorld-MCP dataset and outline how these metrics enable comprehensive performance measurement of different models.

\subsection{Tools Generation and filter} 
\label{subsection:tool collection and filter}
% \vspace{-1ex}
% Existing MCP servers generally offer tools that are relatively simple, often with overlapping functionalities, and provide only a limited number of usable tools for specific software application scenarios. As a result, there is a lack of high-quality, readily usable MCP tools. To address this issue, we designed an automated tool generation pipeline in which the user only needs to provide the task to be performed, and our LLM automatically generates a tool to accomplish it, as shown in Figure \ref{fig:fig3}.(b). Based on OpenAI o3~\citep{openai_o3_o4mini_2025b}, we generated tools for each task in OSWorld. Drawing inspiration from CoAct's prompting approach~\citep{song2025coact}, we constructed our own prompt for tool generation and instructed OpenAI o3 to generate feasible tool-based solutions whenever possible, enabling task completion through these tools. We then selected those tools that successfully solved their respective tasks, resulting in a total of 145 new tools. Finally, we designed prompts for o3 (as shown in Figure \ref{fig:fig3}.(a).) to automatically wrap these tools into an MCP server for deployment and use.

\begin{figure*}[t]
    \centering
    \includegraphics[width=0.9\textwidth]{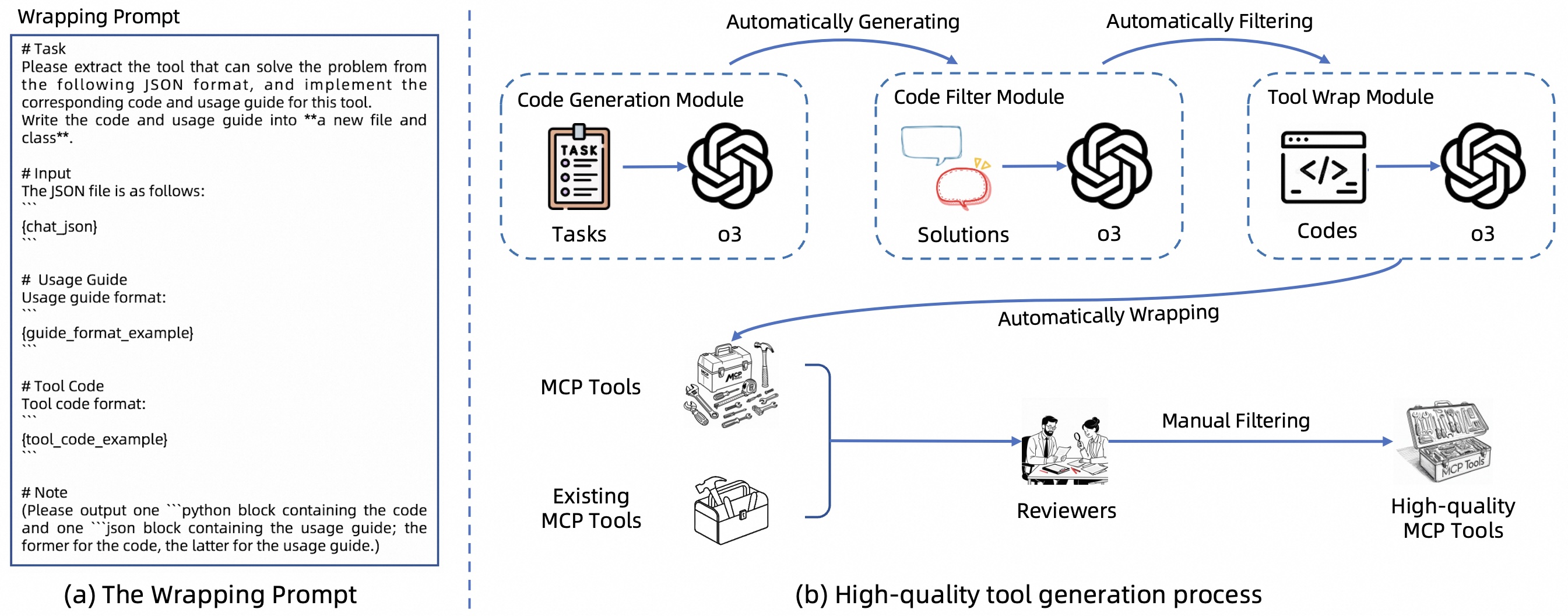}
    % \vspace{-2ex}
    % \caption{Illustration of our tool generation process. (a) shows the prompt we designed for automating the wrapping of code into MCP Tools using OpenAI o3. (b) illustrates the workflow for generating high-quality tools.}
    \caption{Illustration of our tool generation process. (a) Prompt for wrapping code into MCP tools with OpenAI o3. (b) Workflow for generating high-quality tools.}
    \label{fig:fig3}
    % \vspace{-3.5ex}
\end{figure*}

Existing MCP servers generally offer tools that are relatively simple and often have overlapping functionalities. Consequently, there is a shortage of high-quality MCP tools that can be readily applied in practical scenarios. To address this limitation, we design an automated tool generation pipeline composed of three modules: the Code Generation Module, the Code Filter Module, and the Tool Wrap Module. In the Code Generation Module, the user only needs to specify the target task, and the module automatically generates code to accomplish it. Specifically, we employ OpenAI o3~\citep{openai_o3_o4mini_2025b} to produce code-based solutions for every task in OSWorld. Following the prompting strategy of CoAct~\citep{song2025coact}, we develop our own prompt and instruct OpenAI o3 to generate, whenever feasible, functional code solutions capable of completing the tasks. In the Code Filter Module, we use OpenAI o3 to summarize usable code obtained from multi-turn interactions. This summarized code is then applied to solve the corresponding tasks, and any code successfully completing the tasks is retained. Through this process, we obtain seventy-two verified solutions. In the Tool Wrap Module, we provide OpenAI o3 with a prompt that automates the packaging of these verified code solutions into 72 MCP tools, as illustrated in Figure~\ref{fig:fig3}.(a).
% Existing MCP servers typically provide tools that are relatively simple, often with overlapping functionalities. As a result, there is a lack of high-quality, readily applicable MCP tools. To address this issue, we designed an automated tool generation pipeline consisting of three modules: the Code Generation Module, the Code Filter Module, and the Tool Wrap Module. In the Code Generation Module, the user only needs to provide the target task, and the module automatically generates code to solve it. Specifically, we used OpenAI o3 to produce code-based solutions for every task in OSWorld. Following the prompting strategy of CoAct, we constructed our own prompt and instructed OpenAI o3 to generate, whenever possible, functional code solutions that could be used to complete the tasks. In the Code Filter Module, we used OpenAI o3 to summarize the usable code produced during multi-turn interactions. This summarized code was then applied to solve the corresponding tasks. We retained the code that successfully completed the tasks, resulting in a total of 72 verified solutions. In the Tool Wrap module, we provided o3 with a prompt that automated the process of packaging these verified code solutions into 72 MCP tools, as shown in Figure \ref{fig:fig3}.(a).

In addition, we carefully curate 192 tools from existing MCP servers. Since some generate or imported tools are tailored to solve a single specific task and thus unsuitable for real-world use, we conduct a manual inspection of all 264 collected tools to remove such task-specific items as well as functionality duplicates. Each tool is independently evaluated by at least two reviewers with extensive GUI agent development experience, and is retained only if both reviewers deemed it qualified. After this manual filtering process, we obtain 158 high-quality tools that are both applicable and valuable in real-world computing environments.
% \scalebox{0.9}[0.9]{

\subsection{Tools Analysis} 
\label{tool_analysis}

\begin{figure*}[tb]
    \centering
    \includegraphics[width=0.96\textwidth]{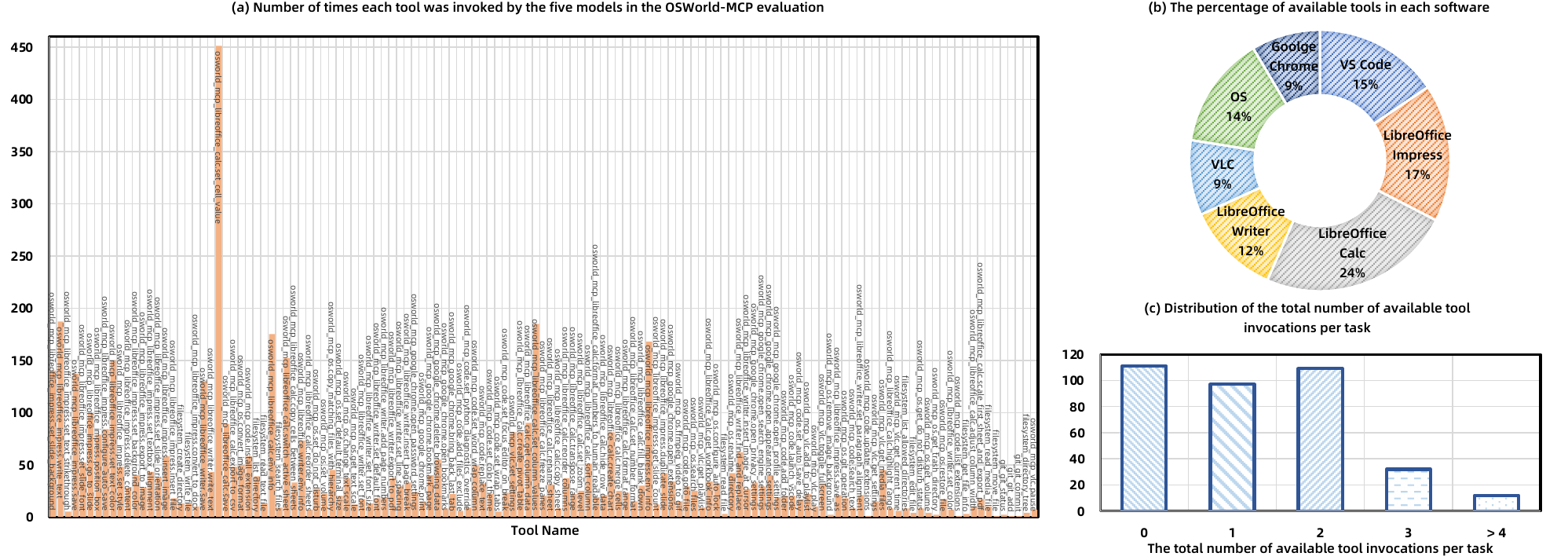}
    % \vspace{-2ex}
    \caption{(a) illustrates the total number of times each tool was invoked by the five models in a single OSWorld-MCP evaluation. (b) depicts the distribution of our 158 high‑quality tools across different usage scenarios for various OSWorld environments. Due to software version constraints within OSWorld, MCP servers were not developed for GIMP and Thunderbird. (c) presents the distribution of the total available tool invocations per task.}
    % \vspace{-2ex}
    \label{fig:fig6}
    % \vspace{-4ex}
\end{figure*}

We analyze the composition of the 158 high-quality tools designed for practical real-world use, with their distribution across application scenarios shown in Figure~\ref{fig:fig6}.(b). To further ensure that these tools have a tangible positive impact on task completion in realistic settings, we conduct an additional manual validation and find that 133 tools effectively contribute to improving task efficiency, while the remaining 25 tools originate from existing external MCP servers.

To verify that the effective tools can be actively utilized by models, we evaluate five state-of-the-art large multimodal models, including Qwen2.5-VL-72B-Instruct~\citep{bai2025qwen2} and Claude 4 Sonnet~\citep{anthropicClaudeSonnet}, on OSWorld-MCP, allowing them at each step to autonomously choose between performing GUI operations and invoking any of the 158 high-quality tools. The results indicate that 131 tools are invoked at least once during evaluation. The remaining two tools, which are manually re-verified for usability, are hypothesized to be absent from model usage due to the complexity of the associated tasks, which likely discouraged models from attempting to invoke them. Tool invocation frequencies are presented in Figure~\ref{fig:fig6}.(a).

We also conduct a manual analysis of all 361 OSWorld-MCP tasks (8 Google Drive tasks excluded) to annotate the set of available tools for each task and to record the total number of invocations of these tools across evaluations. The distribution of total available tool invocations per task is shown in Figure~\ref{fig:fig6}.(c). An available tool is defined as one whose invocation can make task execution substantially more efficient. Based on this definition, OSWorld-MCP is classified into two subsets: \textbf{Tool-Beneficial} Tasks, which include tasks for which at least one available tool can improve efficiency (250 tasks), and \textbf{Non-Tool-Beneficial} Tasks, which include tasks for which no available tool improves efficiency (111 tasks).

These findings demonstrate that the tools in OSWorld-MCP are genuinely relevant to real-world needs and are not artificially tailored for specific benchmark tasks. Furthermore, many tasks can be completed more efficiently through multiple tool invocations, requiring models to select appropriate tools and to exhibit strong decision-making capabilities. This confirms the soundness of our tool design and underscores the challenging nature of the benchmark.

\subsection{Metrics} 

Building on the evaluation of GUI operation capabilities, we introduce three metrics in OSWorld-MCP to assess a multimodal agent's tool invocation and decision-making abilities: Task Accuracy, Tool Invocation Rate, and Average Completion Steps.

\textbf{Task Accuracy. }
Similar to OSWorld, we use accuracy as an overall performance indicator to measure how well a multimodal agent completes tasks. This metric jointly reflects the agent's decision-making ability, tool invocation skills, and GUI interaction performance.

\textbf{Tool Invocation Rate (TIR). }
As described in Section \ref{subsection:tool collection and filter}, human reviewers classify each task into one of two categories: Tool-Beneficial Tasks or Non-Tool-Beneficial Tasks.Let $N_t$ be the total number of Tool-Beneficial Tasks, and $n_t$ the number of such tasks in which the agent invoked a tool and successfully completed the task during evaluation. Let $N_g$ be the total number of Non-Tool-Beneficial Tasks, and $n_g$ the number of such tasks in which the agent did not invoke a tool and successfully completed the task. We define TIR as:

% \vspace{-3ex}
\begin{equation}
\textrm{TIR} = {(n_t + n_g)}/{(N_t + N_g)}
\end{equation}
% \vspace{-3ex}

% When computing TIR within the Tool-Beneficial Task subset ($N_g = 0$), the formula reduces to $\textrm{TIR} = {n_t}/{N_t}$,
% which measures the proportion of Tool-Beneficial Tasks where the agent correctly invoked a tool and succeeded. When computing TIR within the Non-Tool-Beneficial Task subset ($N_t = 0$), the formula reduces to $\textrm{TIR} = {n_g}/{N_g}$, which measures the proportion of Non-Tool-Beneficial Tasks where the agent correctly avoided tool invocation and succeeded. 
TIR can effectively indicate the agent's ability to decide when a tool should or should not be invoked.
% TIR thus serves as an effective indicator of the agent's ability to decide when a tool should or should not be invoked.
% For each task, two experienced GUI agent developers manually reviewed it and listed all applicable tools from our curated set. As noted in Section \ref{subsection:tool collection and filter}, we removed tools with overlapping functionality, so the listed tools are unique utility combinations rather than interchangeable alternatives. After manual review, 232 tasks were identified as ones that could be completed more efficiently by invoking tools. Figure X \ref{} shows the distribution of tool-use frequencies for these tasks. Let $N_t$ be the number of tool-beneficial tasks, and $n_t$ be the number of these tasks in which the agent actually invoked a tool during evaluation. Let $N_g$ be the number of tasks for which tool invocation would not yield a more efficient solution, and $n_g$ the number of such tasks where the agent indeed did not use a tool. We define $TIR$ as:

% \begin{equation}
%     TIR=\frac{n_t + n_g}{N_t + N_g}
% \end{equation}
% $TIR$ effectively measures a multimodal agent's ability to utilize tools in relevant scenarios.

\textbf{Average Completion Steps (ACS). }
This metric measures the average number of steps an agent takes to complete a task. For $N$ tasks, if the number of execution steps for task $i$ is $S_i$, the Average Completion Steps is computed as:
% \vspace{-2ex}
\begin{equation}
\textrm{ACS} = {\sum_{i=1}^{N}{S_i}}/{N}    
\end{equation}
% \vspace{-3ex}

ACS reflects decision-making efficiency: the more accurate the decisions, the higher the rate of correct tool usage, and the more frequently the agent selects efficient tools, the lower ACS will be.

\section{Experiments}

\subsection{Setup}

We evaluate a series of state-of-the-art Large Multimodal Models (LMMs), including Qwen2.5-VL-72B-Instruct~\citep{bai2025qwen2}, Qwen3-VL-Plus~\citep{qwen3_vl_blog}, Seed1.5-VL~\citep{guo2025seed1}, Claude 4 Sonnet~\citep{anthropicClaudeSonnet}, OpenAI o3~\citep{openai_o3_o4mini_2025b}, and Gemini-2.5-Pro~\citep{comanici2025gemini}, by running a computer-use agent on real-world tasks in OSWorld-MCP involving nine different software applications.
To facilitate comparison of performance differences across models, we standardized our evaluation using the GUI‑Owl agent configuration. This may lead to some performance fluctuations for certain models under the original OSWorld configuration.
% Our evaluation adopts the GUI-Owl agent configuration. 
Specifically, at each step, the core LMM performs visual perception of the current interface, generates a corresponding thought, and proposes the next action along with a reasoning summary. This reasoning history informs subsequent planning during task execution. 
% for all models.
Besides, we also evaluate a multi-agent framework Agent-S2.5~\citep{agents25},
% Experiments on multi-agent frameworks are also conducted, involving Agent-S2.5~\citep{agents25} and Mobile-Agent-v3~\citep{ye2025mobile}. 
and adopt OpenAI o3 as the main generation model with UI-TARS-1.5-72B as the grounding model.
For each task, the agent is restricted to a fixed maximum number of steps to either complete the task or determine that it is infeasible. The temperature parameter is set to 1.0.

% Our evaluation adopts the GUI-Owl agent configuration, where the core LMM performs visual perception of the current interface, generates a corresponding thought, and proposes the next action with reasoning at each step. This reasoning history informs subsequent planning during task execution. For each task, the agent is restricted to a fixed maximum number of steps, with the temperature parameter set to 1 for all models.

In the original OSWorld configuration, LMMs may only use a predefined set of 11 basic GUI operations, including \textit{key, type, mouse\_move, click, drag, right\_click, middle\_click, double\_click, scroll, wait and terminate}, to complete tasks.
With the introduction of MCP tools in our OSWorld-MCP, the LMM can, autonomously decide whether to invoke any MCP Tool or perform a GUI operation at each action step.
In details, we employ the 158 high-quality MCP tools curated in Section \ref{subsection:tool collection and filter}. 
Since providing all 158 tools simultaneously would result in excessively long input contexts,
% and severely impair decision-making performance
% at each action-prediction step
we apply Retrieval-Augmented Generation (RAG) to select only the tools relevant to the current application. 
The model then chooses from these filtered MCP tools or GUI operations.

\subsection{Main Results}
\label{main-res}
% As shown in Table \ref{table:main-results}, we evaluate the performance of five different foundation end-to-end models and two agentic frameworks OSWorld-MCP with our curated high-quality tools.
% Among the end-to-end agent, Claude 4 Sonnet achieves the highest accuracy in OSWorld-MCP, reaching 38.6 (15 steps) and 45.0 (50 steps), respectively. It attains the highest accuracy (28.7) in the original OSWorld configuration, while maintaining high Tool Invocation Rates (TIR) in both the Tool-Beneficial Tasks and Non-Tool-Beneficial Tasks subsets. 
% %
% Qwen-2.5-VL-72B exhibits the smallest accuracy improvement in OSWorld-MCP among all models, which indicates the lack of its tool-calling ability.
% %
% In addition, Seed1.5-VL records the lowest Average Completion Steps (ACS) in OSWorld-MCP, at only 10.0 (15 steps), while Claude 4 Sonnet reaches 20.1 (50 steps), indicating a better tool invocation and decision-making capabilities of the two models. 
% %
% Our findings also show that current LMMs generally exhibit low TIR in the OSWorld-MCP benchmark. Even the highest — achieved by Seed1.5-VL — is only 23.6\%, while the lowest, Qwen2.5-VL-72B-Instruct, is just 11.4\%, highlighting persistent challenges in invoking tools correctly and efficiently.
% %
% For multi-agent frameworks, the Agent-S2.5 reach the highest performance in both 15 steps and 50 steps settings with 41.9 and xxx accuracy, respectively. Most agents obtain improvement in accuracy and execution steps when equipped with MCP tools, indicating the effectiveness of external tools to address challenging tasks.

As shown in Table \ref{table:main-results}, we evaluate six advanced end-to-end models and one agent-based frameworks under both the original OSWorld configuration and OSWorld-MCP with our curated high-quality tools, using maximum step limits of 15 and 50 respectively. 
% We also conduct evaluation with a maximum step limit of 50, and the results are reported in Table \ref{table:main-results-50} in the Appendix \ref{sec:more-result}. 
Due to fluctuations in the experimental results, we conducted three runs for each model or framework under each configuration. The results reported in Tables \ref{table:main-results} are the averages over these three runs.

Among the end-to-end models, Claude 4 Sonnet achieves the highest accuracy in OSWorld-MCP for both step limits, reaching 35.3 (15 steps) and 43.3 (50 steps) respectively among LMMs. Claude 4 Sonnet also records the highest tool invocation rate in OSWorld-MCP. Seed-VL1.5 and Claude 4 Sonnet achieve the lowest average completion steps (ACS) at 15 and 50 steps respectively, with values of 10.2 and 20.1. 
% Since these two models also obtain the highest accuracies among the five models, this indicates that Claude 4 Sonnet and Seed-VL1.5 possess strong decision-making capabilities. 
Our results also reveal that current large multimodal models generally have low tool invocation rates in OSWorld-MCP: even the highest, achieved by Claude 4 Sonnet, is only 36.3 percent, while the lowest, from Qwen2.5-VL-72B-Instruct, is 10.9 percent. This shows that these models still face challenges in invoking tools both correctly and efficiently.
%
% For the multi-agent frameworks, Agent-S2.5 delivers the best performance at both step limits, achieving accuracies of 42.8 and 51.4. Most agents show improvements in both accuracy and execution efficiency when equipped with MCP tools, indicating that external tools are highly effective for handling challenging tasks.
%
For the multi-agent frameworks, Agent-S2.5 achieves an accuracy of 42.1 at the 15-step limit and 49.5 at the 50-step limit, confirming its effectiveness in handling challenging tasks.
% {\color{red}For the multi-agent frameworks, Agent-S2.5 achieves the highest accuracy (42.8 and 51.4) at both step limits, and most agents improve in accuracy and efficiency with MCP tools, confirming their effectiveness for challenging tasks.}

\begin{table*}[btp]
\centering
\caption{Performance on OSWorld-MCP (15 Steps).}
\vspace{-1ex}
\resizebox{0.96\textwidth}{!}{
\begin{tabular}{l l c lll lll lll}
\toprule
\multirow{2}{*}{\textbf{Agent Model}} & 
\multirow{2}{*}{\textbf{Actions}} & \multirow{2}{*}{\textbf{Steps}} & 
\multicolumn{3}{c}{\textbf{Tool-Beneficial Tasks}} & 
\multicolumn{3}{c}{\textbf{Non-Tool-Beneficial Tasks}} & 
\multicolumn{3}{c}{\textbf{Overall}} \\
& & & 
\multicolumn{1}{c}{\textbf{Accuracy}} & \multicolumn{1}{c}{\textbf{TIR}} & \multicolumn{1}{c}{\textbf{ACS}} & 
\multicolumn{1}{c}{\textbf{Accuracy}} & \multicolumn{1}{c}{\textbf{TIR}} & \multicolumn{1}{c}{\textbf{ACS}} & 
\multicolumn{1}{c}{\textbf{Accuracy}} & \multicolumn{1}{c}{\textbf{TIR}} & \multicolumn{1}{c}{\textbf{ACS}} \\
\midrule
\multicolumn{12}{c}{\textit{Open Models}} \\
\midrule
\multirow{4}{*}{Qwen2.5-VL} 
 & GUI & 15 & 10.1 & - & 13.0 & 14.4 & - & 12.9 & 11.4 & - & 13.0 \\
 & + MCP & 15  & 14.7$_{\;4.6\uparrow}$ & 11.6 & 13.6$_{\;0.6\uparrow}$ & 18.3$_{\;3.9\uparrow}$ & 16.5 & 13.1$_{\;0.2\uparrow}$ & 15.8$_{\;4.4\uparrow}$ & 13.1 & 13.5$_{\;0.5\uparrow}$  \\
 & GUI & 50 & 12.3 & - & 30.6 & 17.7 & - & 30.4 & 13.9 & - & 30.5 \\
 & + MCP & 50 & 11.7$_{\;0.6\downarrow}$ & 7.3 & 38.6$_{\;8.0\uparrow}$ & 21.6$_{\;3.9\uparrow}$ & 18.9 & 34.2$_{\;3.8\uparrow}$ & 14.8$_{\;0.9\uparrow}$ & 10.9 & 37.2$_{\;6.7\uparrow}$  \\
 \multirow{4}{*}{Qwen3-VL} 
 & GUI & 15 & 24.5 & - & 11.5 & 27.3 & - & 11.7 & 25.4 & - & 11.6 \\
 & + MCP& 15 & 30.5$_{{\;6.0\uparrow}}$ & 23.1 & 10.2$_{{\;1.5\downarrow}}$ & 33.0 $_{{\;5.7\uparrow}}$ & 27.6 & 11.1$_{{\;0.6\downarrow}}$ & 31.3$_{{\;5.9\uparrow}}$ & 24.5 & 10.5$_{{\;1.1\downarrow}}$ \\
 & GUI & 50 & 31.9 & - & 25.5 & 38.0 & - & 25.7 & 33.8 & - & 25.6 \\
 & + MCP& 50 & 37.0 $_{{\;5.1\uparrow}}$ & 27.7 & 20.2$_{{\;5.3\downarrow}}$ & 43.8$_{{\;5.8\uparrow}}$ & 33.3 & 23.3$_{{\;2.4\downarrow}}$ & 39.1$_{{\;5.3\uparrow}}$ & 29.5 & 21.1$_{{\;4.5\downarrow}}$ \\
\midrule
\multicolumn{12}{c}{\textit{Proprietary Models}} \\
\midrule
\multirow{4}{*}{Gemini-2.5-Pro} 
 & GUI & 15  & 6.3 & - & 13.7 & 9.8 & - & 14.1 & 7.4 & - & 13.8 \\
 & + MCP & 15 & 24.9$_{\;18.6\uparrow}$ & 20.4 & 10.2$_{\;3.5\downarrow}$ & 9.9$_{\;0.1\uparrow}$ & 8.7 & 13.9$_{\;0.2\downarrow}$ & 20.5$_{\;13.1\uparrow}$ & 16.8 & 11.4$_{\;2.4\downarrow}$  \\
 & GUI & 50 & 11.5 & - & 39.7 & 17.5 & - & 41.7 & 13.3 & - & 40.3  \\
 & + MCP & 50 & 29.5$_{\;18.0\uparrow}$ & 23.2 & 25.2$_{\;14.5\downarrow}$ & 21.6$_{\;4.1\uparrow}$ & 17.7 & 40.2$_{\;1.5\downarrow}$ & 27.2$_{\;13.9\uparrow}$ & 21.5 & 29.7$_{\;10.6\downarrow}$ \\
\multirow{4}{*}{OpenAI o3} 
 & GUI & 15  & 8.5 & - & 13.8 & 7.8 & - & 14.4 & 8.3 & - & 14.0 \\
 & + MCP & 15  & 25.4$_{\;16.9\uparrow}$ & 21.3 & 10.5$_{\;3.3\downarrow}$ & 9.1$_{\;1.3\uparrow}$ & 6.3 & 14.1$_{\;0.3\downarrow}$ & 20.4$_{\;12.1\uparrow}$ & 16.7 & 11.6$_{\;2.4\downarrow}$  \\
 & GUI & 50 & 10.7 & - & 43.9 & 17.4 & - & 46.8 & 12.8 & - & 44.8 \\
 & + MCP & 50 & 28.9$_{\;18.2\uparrow}$ & 24.8 & 27.0$_{\;16.9\downarrow}$ & 16.8$_{\;0.6\downarrow}$ & 12.3 & 43.7$_{\;3.1\downarrow}$ & 25.2$_{\;12.4\uparrow}$ & 21.0 & 32.1$_{\;12.7\downarrow}$ \\
\multirow{4}{*}{Seed1.5-VL} 
 & GUI & 15  & 27.3 & - & 10.6 & 29.3 & - & 11.5 & 27.9 & - & 10.9 \\
 & + MCP & 15  & 31.4$_{\;4.1\uparrow}$ & 21.7 & 9.6$_{\;1.0 \downarrow }$ & 33.3$_{\;4.0\uparrow}$ & 32.7 & 11.5$_{\;0.0 - }$ & 32.0$_{\;4.1\uparrow}$ & 25.1 & 10.2$_{\;0.7\downarrow}$ \\
 & GUI & 50 & 31.2 & - & 22.4 & 40.2 & - & 26.8 & 34.0 & - & 23.8 \\
 & + MCP & 50 & 36.1$_{{\;4.9\uparrow}}$ & 22.7 & 20.8$_{{\;1.6\downarrow}}$ & 43.6$_{{\;3.4\uparrow}}$ & 43.2 & 27.8$_{{\;1.0\uparrow}}$ & 38.4$_{{\;4.4\uparrow}}$ & 29.0 & 23.0$_{{\;0.8\downarrow}}$ \\
\multirow{4}{*}{Claude-4-Sonnet} 
 & GUI & 15  & 29.0 & - & 11.8 & 32.9 & - & 12.0 & 30.2 & - & 11.9 \\
 & + MCP & 15 & 35.6$_{{\;6.6\uparrow}}$  & 29.7  & 9.8 $_{{\;2.0\downarrow}}$ & 34.5$_{{\;1.6\uparrow}}$ & 30.9 & 11.7$_{{\;0.3\downarrow}}$ & 35.3$_{{\;5.1\uparrow}}$ & 30.0 & 10.4$_{{\;1.5\downarrow}}$ \\
 & GUI & 50 & 38.2 & - & 24.2 & 44.1 & - & 25.6 & 40.1 & - & 24.7 \\
 & + MCP& 50 & 42.4$_{{\;4.2\uparrow}}$ & 35.3 & 18.8$_{{\;5.4\downarrow}}$ & 45.5$_{{\;1.4\uparrow}}$ & 38.7 & 22.8$_{{\;2.8\downarrow}}$ & 43.3$_{{\;3.2\uparrow}}$ & 36.3 &  20.1$_{{\;3.6\downarrow}}$ \\
\midrule
\multicolumn{12}{c}{\textit{Multi-agent Models}} \\
\midrule

 \multirow{4}{*}{Agent-S2.5} 
 & GUI & 15 & 35.8 & - & 11.4 & 38.6 & - & 11.2 & 36.7 & - & 11.3  \\
 & + MCP & 15 & 42.9$_{{\;7.1\uparrow}}$ & 28.1 & 9.6$_{{\;1.8\downarrow}}$ & 40.4$_{{\;1.8\uparrow}}$ & 34.2 & 10.9$_{{\;0.3\downarrow}}$ & 42.1$_{{\;5.4\uparrow}}$ & 30.0 & 10.0$_{{\;1.3\downarrow}}$ \\
 & GUI & 50 & 46.9 & - & 21.1 & 47.3 & - & 8.4 & 47.1 & - & 20.2 \\
 & + MCP & 50 & 49.4$_{{\;2.5\uparrow}}$ & 32.9 & 16.3$_{{\;4.8\downarrow}}$ & 49.6$_{{\;2.3\uparrow}}$ & 40.5 & 18.7$_{{\;0.3\uparrow}}$ & 49.5$_{{\;2.4\uparrow}}$ & 35.3 & 17.0 $_{{\;3.2\downarrow}}$  \\
 
% \multirow{2}{*}{Mobile-Agent-v3} 
%  & GUI & 15  & 28.9 & - & 17.7 & 34.3 & - & 24.1 & 30.8 & - & 19.7 \\
%  & + MCP& 15  & 24.8$_{{\;4.1\downarrow}}$ & 0.0 & 22.1$_{{\;4.4\uparrow}}$ & 40.5$_{{\;6.2\uparrow}}$ & 36.9 & 24.7$_{{\;0.6\uparrow}}$ & 29.9$_{{\;0.9\downarrow}}$ & 11.4 & 22.9$_{{\;3.2\uparrow}}$ \\
\bottomrule
\end{tabular}
}
\vspace{-2ex}
\label{table:main-results}
\end{table*}

\textbf{Impact of Tool Invocation on Accuracy and Efficiency.}
A comparison with the GUI-only setting shows that, with the exception of Qwen2.5-VL, all six end-to-end models and both agent frameworks achieve higher accuracies and lower ACS after the introduction of MCP tools. Qwen2.5-VL, despite achieving a slight accuracy improvement, shows increased ACS. It indicates poor tool invocation capability and weak decision-making ability, leading to longer average completion times.
% Its tool invocation rate (TIR) in Tool-Beneficial Tasks is only 6.8 at 15 steps and 7.2 at 50 steps, suggesting poor tool usage and decision-making. This likely leads to the invocation of tools that do not aid efficient execution and sometimes even incorrect tools, resulting in longer task completion times.

Among the remaining models and frameworks, Gemini-2.5-Pro exhibits the most significant improvement. At 15 steps, its overall accuracy rises from 7.4 to 20.5, while ACS decreases from 13.8 to 11.4. 
% In Tool-Beneficial Tasks, its accuracy increases from 8.8 to 26.3, and its tool invocation rate reaches 20.0 percent, ranking just below Claude 4 Sonnet and Agent-S2.5.
%
Furthermore, a comparison of Tool-Beneficial and Non-Tool-Beneficial Tasks shows that, except for Qwen2.5-VL-72B-Instruct, all models demonstrate substantial accuracy improvement in Tool-Beneficial Tasks after the introduction of MCP tools. The largest gain is observed in Gemini-2.5-Pro at 15 steps, where accuracy increases from 6.3 to 24.9. In addition, accuracy changes for Non-Tool-Beneficial Tasks are minor. 
% For these tasks, some models such as Qwen2.5-VL and Gemini-2.5-Pro achieve slight performance gains, while others experience small decreases. 
Two possible factors may explain these results: First, Non‑Tool‑Beneficial Tasks refer only to tasks without efficiency‑enhancing tools, yet they may still include tools that, while not improving efficiency, have a positive effect on task completion. Such tools can make it easier for the model to solve the problem, thereby increasing Acc.
Second, for tasks that contain no tools beneficial to the task at all, the tools provided can help the model rule out irrelevant solution paths, making it easier to execute the task along the correct path and thus improving task Accuracy.
This also explains why ACS often decreases on Non‑Tool‑Beneficial Tasks.
% First, some models, including Qwen2.5-VL, may inherently have weaker tool invocation capabilities, so MCP tools have limited impact on their performance in these tasks. Second, output variability in some models may cause slight accuracy increases in Non-Tool-Beneficial Tasks.
% %
From these results, we derive the following conclusion:

\begin{tcolorbox}[colframe=blue!75!black,colback=blue!5]
  % \emph{\textbf{Finding 1:} Effective use of MCP tools has a significant positive impact on improving LMM accuracy in completing OSWorld tasks and on reducing average completion steps.}
  % \vspace{-0.5ex}
  \emph{\textbf{Finding 1:} MCP tools significantly enhance LMMs' performance in computer-use, improving accuracy and reducing completion steps for most models. The effectiveness varies across different LMMs, indicating disparities in tool utilization capabilities.}
  % \vspace{-0.5ex}
  \label{finding1}
\end{tcolorbox}

\begin{figure*}[htbp]
    \centering
    % \vspace{-2ex}
    \includegraphics[width=0.95\textwidth]{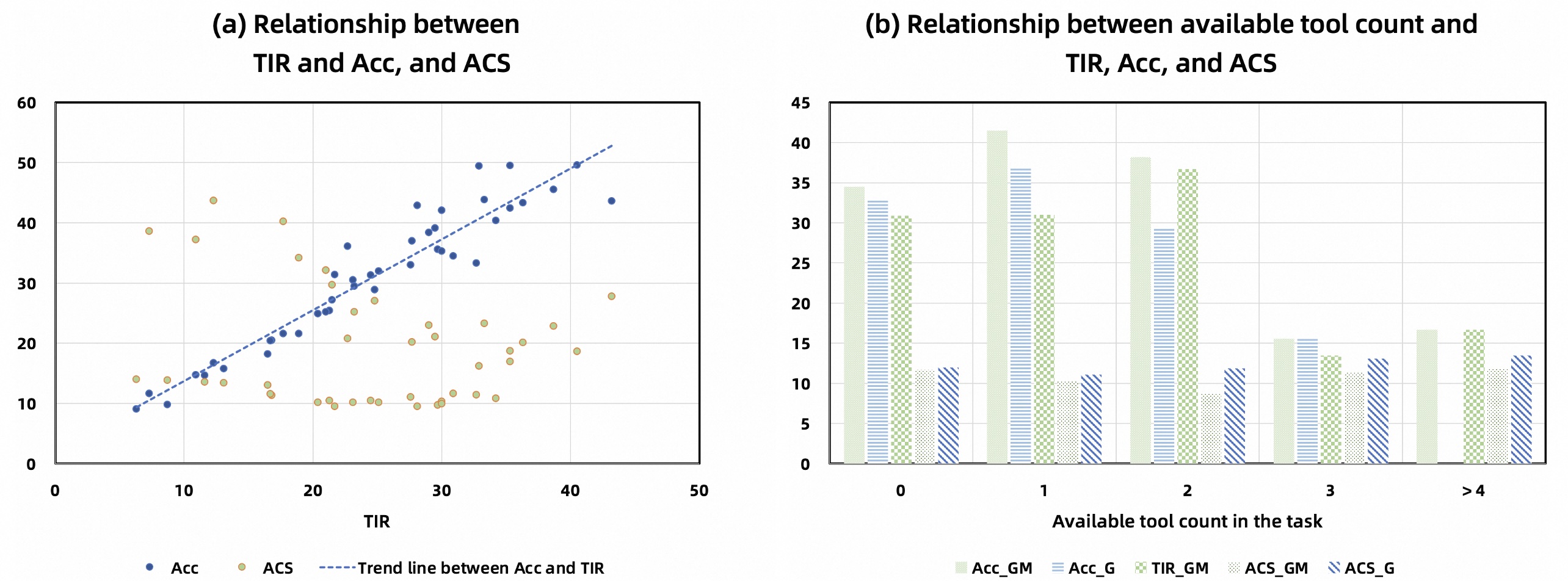}
    % \caption{(a): The relationships between TIR, Acc, and ACS. (b): The variations in TIR, Acc, and ACS of Claude-4-Sonnet across task sets with different numbers of available tools. The suffix GM denotes a setting in which both GUI operations and MCP tools are available, while the suffix G denotes a setting in which only GUI operations are permitted.}
    \caption{(a): The relationships between TIR, Acc, and ACS. (b): The performance of Claude-4-Sonnet across task sets with different numbers of available tools. GM: GUI + MCP , G: GUI Only.}
    \label{fig:fig5}
    % \vspace{-2ex}
\end{figure*}

\textbf{Impact of Tool Invocation Rate on Accuracy and Average Completion Steps.}
Our experiments reveal that, across different models, the tool invocation rate (TIR) and accuracy (Acc) generally exhibit a positive correlation, whereas Average Completion Steps (ACS) show no obvious correlation with TIR. We computed TIR, Acc, and ACS for each model under varying maximum step limits, across Tool-Beneficial Tasks, Non-Tool-Beneficial Tasks, and the entire task set. These results were aggregated into a single chart (Figure \ref{fig:fig5}.(a)).
As shown in the figure, for a given model, higher TIR values tend to correspond to higher accuracies, indicating a clear positive relationship between tool invocation and task success. 
Notably, this relationship remains stable regardless of differences in step limits or task sets. This strongly supports the soundness of our MCP tool design.
In contrast, ACS does not show a clear correlation with TIR. Further analysis of ACS–TIR patterns across different task sets and step limits suggests that, under the same settings, an increase in TIR can sometimes coincide with a decrease in ACS. We hypothesize two possible reasons for this phenomenon:
a). TIR reflects the proportion of correct tool invocations. A higher TIR indicates a higher proportion of correct tool usage, which can enable the model to complete tasks more efficiently.
b). The complexity of OSWorld-MCP tasks varies significantly, and the overall task completion rate remains relatively low. When the maximum step limit is raised, models tend to make more attempts in solving complex tasks, which can counteract or obscure the efficiency gains that come from correct tool usage.
% However, ACS does not display a consistent relationship with TIR. Further analysis of ACS–TIR patterns across different task sets and step limits suggests that, in some settings, an increase in TIR can coincide with a decrease in ACS. We indicate this is due to two complementary factors: a). TIR measures the proportion of correct tool use. A higher TIR implies a greater proportion of correct tool invocations. As established in Finding 1, correct tool use can both improve task accuracy and enable more efficient task completion. b). Task complexity in OSWorld-MCP varies widely. Overall task completion rates remain relatively low, and when the maximum step limit is raised, models tend to make more attempts to solve difficult tasks, which may offset or mask the efficiency gains from correct tool usage.
% \vspace{-0.5ex}
\begin{tcolorbox}[colframe=blue!75!black,colback=blue!5]
  % \emph{\textbf{Finding 2:} Tool invocation rate and accuracy show a positive correlation, but TIR has no clear relationship with Average Completion Steps.}
  % \vspace{-0.5ex}
  \emph{\textbf{Finding 2:} Tool Invocation Rate (TIR) positively correlates with task accuracy, but its relationship with ACS is complex and non-linear, suggesting that the impact of tool use on efficiency depends on various factors including task difficulty and model-specific strategies.}
  \label{finding2}
  \vspace{-2ex}
\end{tcolorbox}
% \vspace{-0.5ex}

% Tool Invocation Rate (TIR) positively correlates with task accuracy, but its relationship with ACS is complex and non-linear, suggesting that the impact of tool use on efficiency depends on various factors including task difficulty and model-specific strategies.

\textbf{Impact of the number of available tools.}
As shown in Figure \ref{fig:fig5}.(b), we conducted experiments using the best-performing model from the previous evaluations, Claude 4 Sonnet, on the manually annotated task set described in Section \ref{tool_analysis}. The tasks were grouped according to the number of available tools. 
% Claude 4 Sonnet was tested under two configurations: one allowing only GUI operations and the other allowing both GUI operations and MCP tool usage. 
For each configuration, we computed Acc, TIR, and ACS.

Firstly, in the results for the GUI-only configuration, we observe that as the number of available tools increases, Acc tends to decrease and ACS tends to increase. This indicates that task difficulty rises with the number of available tools.
% Firstly, in the results for the GUI-only configuration, we observed that as the number of available tools increased, Acc tended to decrease and ACS tended to increase. This is because a greater number of available tools corresponds to more complex tasks, and without access to MCP tools, such tasks require longer GUI operation sequences and place higher demands on the model's step-by-step GUI operation accuracy.
Secondly,in the results for the GUI plus MCP Tools configuration, we found that when the number of available tools was relatively small, Acc increased and ACS decreased. However, as the number of available tools grew, both Acc and TIR dropped sharply, and ACS gradually rose. A possible explanation is that with fewer available tools, the tasks are relatively easier, and the model is more likely to select the tools that are useful for the task, thereby completing it with higher accuracy and efficiency. When the number of available tools increases, the tasks become more complex. Even though more efficient tools might be available, these tasks often require multiple-tool combinations, making it difficult for the model to accurately select the most relevant tools, leading to reduced task accuracy and higher average completion steps.
Thirdly,
when comparing the GUI-only configuration with the GUI plus MCP Tools configuration, the latter consistently achieved lower ACS and higher accuracy overall. However, in cases where ACS was similar in both configurations, we found that the GUI plus MCP Tools configuration sometimes resulted in lower accuracy. We suspect that the use of MCP tool combinations is more challenging for LMMs than combining GUI operations.

\begin{table*}[tbp]
\centering
\caption{Performance of Gemini-2.5-Pro on OSWorld-MCP under different configurations.}
\vspace{-1ex}
% \caption{The performance of Gemini-2.5-Pro on OSWorld-MCP is evaluated under three settings. Base (MCP) refers to the same configuration as used in Section \ref{main-res}. \textit{w/} Tools Shuffle is based on the Base (MCP) setting but with the tool list randomly shuffled. \textit{w/o} Tools RAG is based on the Base (MCP) setting with the removal of Tools RAG tailored to the currently operated software.}
\resizebox{0.96\textwidth}{!}{
\begin{tabular}{l l lll lll lll}
\toprule
\multirow{2}{*}{\textbf{Agent Model}} & 
\multirow{2}{*}{\textbf{Settings}} & 
\multicolumn{3}{c}{\textbf{Tool-Beneficial Tasks}} & 
\multicolumn{3}{c}{\textbf{Non-Tool-Beneficial Tasks}} & 
\multicolumn{3}{c}{\textbf{Overall}} \\
& & 
\multicolumn{1}{c}{\textbf{Accuracy}} & \multicolumn{1}{c}{\textbf{TIR}} & \multicolumn{1}{c}{\textbf{ACS}} & 
\multicolumn{1}{c}{\textbf{Accuracy}} & \multicolumn{1}{c}{\textbf{TIR}} & \multicolumn{1}{c}{\textbf{ACS}} & 
\multicolumn{1}{c}{\textbf{Accuracy}} & \multicolumn{1}{c}{\textbf{TIR}} & \multicolumn{1}{c}{\textbf{ACS}} \\

\midrule
\multirow{3}{*}{Gemini-2.5-Pro} 
 & Base (MCP) & 24.9 & 20.4 & 10.2 & 9.9 & 8.7 & 13.9 & 20.5 & 16.8 & 11.4 \\
 & w/ Tools Shuffle & 24.7$_{{\; 0.2 \downarrow}}$ & 19.2$_{{\; 1.2 \downarrow}}$ & 10.4$_{{\; 0.2 \uparrow}}$ & 18.0$_{{\; 8.1 \uparrow}}$ & 17.1$_{{\; 8.4 \uparrow}}$ & 13.9$_{{\; 0.0 - }}$ & 22.7$_{{\; 2.2 \uparrow}}$ & 18.6$_{{\; 1.8 \uparrow}}$ & 11.5$_{{\; 0.1 \uparrow}}$ \\
 & w/o Tools RAG & 18.0$_{{\; 6.9 \downarrow}}$ & 12.4$_{{\; 8.0 \downarrow}}$ & 8.7$_{{\; 1.5 \downarrow}}$ & 9.9$_{{\; 0.0 - }}$ & 9.9$_{{\; 1.2 \uparrow}}$ & 9.2$_{{\; 4.7 \downarrow}}$ & 15.5$_{{\; 5.0 \downarrow}}$ & 11.6$_{{\; 5.2 \downarrow}}$ & 8.8$_{{\; 2.6 \downarrow}}$ \\

\bottomrule
\end{tabular}
}
\label{table:ablation-results}
\vspace{-3ex}
\end{table*}

% \vspace{-0.5ex}
\begin{tcolorbox}[colframe=blue!75!black,colback=blue!5]
  % \emph{\textbf{Finding 3:} As task difficulty increases, the MCP Tools consistently helps LMMs to complete tasks more efficiently and accurately. However, combining MCP Tools presents a greater challenge to LMMs than combining GUI operations.}
  % \vspace{-0.5ex}
  \emph{\textbf{Finding 3:} MCP tools generally improve performance in complex tasks, but their efficacy diminishes in extremely complex scenarios requiring tool combinations. This indicates that combining multiple tools is more challenging than combining GUI operations.}
  % \vspace{-0.5ex}
\end{tcolorbox}
% \vspace{-0.5ex}

\subsection{Ablation Study}

To more comprehensively evaluate the tool invocation and decision-making capabilities of LMMs, we conducted a series of ablation studies.
We selected Gemini-2.5-Pro, the model with the largest accuracy gain from MCP tools over the GUI-only setting, as the test subject for these experiments.

\textbf{Impact of the Number of Callable Tools on Model Accuracy.}
In the default OSWorld-MCP setup, the set of tools available at each step is a subset obtained via Retrieval-Augmented Generation (RAG), filtered to match the current application in use.
To investigate the impact of the number of callable tools on model performance, we removed the RAG filtering and allowed the model to choose freely from all 158 tools for each task.
As shown in Table \ref{table:ablation-results}, removing RAG led to a noticeable performance drop: the overall Acc decreased from 20.5 to 15.5.
% We attribute this to the fact that descriptions of all 158 tools result in an excessively long tool context, which can seriously impair the model's ability to identify the correct tool, thereby disrupting accurate tool invocation.
We attribute this to the fact that descriptions of all 158 tools result in excessively long tool contexts, which markedly reduce the model's tendency to use tools, thereby impairing accurate tool invocation.
For Tool‑beneficial Tasks, removing RAG reduced Acc from 24.9 to 18.0 and TIR from 20.4 to 12.4, indicating a pronounced decline in the model's inclination to invoke tools for problem solving.
In contrast, for Non‑tool‑beneficial Tasks, although Acc remained unchanged after removing RAG, TIR increased from 8.7 to 9.9, suggesting that the model became more inclined to employ the GUI rather than tools to solve the tasks.

\textbf{Impact of Tool Description Order in Prompts.}
% In the default OSWorld-MCP setup, tool descriptions are provided to the LMM in alphabetical order.
% To study the effect of tool description order on model performance, we randomly shuffled the descriptions before passing them to the LMM and re-evaluated on OSWorld-MCP.
% As shown in Table \ref{table:ablation-results}, randomizing the order caused the overall Acc to drop from 23.8 to 22.7.
% This result indicates that the ordering of tool descriptions within the prompt has a measurable impact on model performance in OSWorld-MCP.
In the default OSWorld‑MCP configuration, tool descriptions are provided to the LMM in alphabetical order.
To investigate the impact of tool description ordering on model performance, we randomly shuffled the descriptions prior to passing them to the LMM and re‑evaluated on OSWorld‑MCP.
As shown in Table \ref{table:ablation-results}, random ordering increased overall Acc from 20.5 to 22.7.
This result indicates that the ordering of tool descriptions in the prompt has a substantial effect on model performance in OSWorld‑MCP.
While the model's performance on Tool‑beneficial Tasks was nearly unchanged before and after shuffling, differences were considerable for Non‑tool‑beneficial Tasks.
This may be because, with fewer tools, the model tends to invoke the corresponding tool when one is available, whereas in the absence of available tools, the description order may implicitly suggest alternative solution strategies.
For consistency in evaluation, tool descriptions are presented to the LMM in lexicographical order in OSWorld‑MCP.

% \input{iclr2026/tables/abl1}
% \input{iclr2026/tables/abl2}

% \vspace{-0.5ex}
\subsection{Case Study}
% \vspace{-0.5ex}

% 是否加RAG，所有工具都加入和只加入对应Server进行对比

% TODO: prompt 是否 shuffle？每次调用都 shuffle。工具顺序，按字典序

% TODO：只保留 GUI

% chart，每个 app 的比例

% In order to demonstrate the tool invocation capability of the GUI Agent, we present a complex example in LibreOffice Calc: \textit{Copy the ``Revenue" column along with the header to a new sheet named ``Sheet2"}. This task requires creating a new sheet and copying the specified column into another sheet. For the end-to-end Gemini2.5-Pro agent, the task is accomplished by utilizing the tools for creating a sheet and copying a sheet, followed by employing the tool to switch to Sheet2 to ensure the operation has been completed, as illustrated in Figure \ref{fig:mcp-gemini-steps}. For the agent not equipped with MCP tools, it struggles to select the specific column within the spreadsheet, leading to task failure. This demonstrates that the inclusion of MCP helps the agent accomplish tasks with precision, showcasing how tool calls can serve as a valuable complement to the agent's capabilities. A more detailed analysis of this task is provided in Appendix \ref{appendix:more-cases}.

In order to demonstrate the tool invocation capability of the GUI Agent, we present a complex example in LibreOffice Calc: \textit{Copy the ``Revenue'' column along with the header to a new sheet named ``Sheet2''}. This task requires creating a new sheet and copying the specified column. The end-to-end Gemini-2.5-Pro agent accomplishes this by utilizing tools for creating a sheet and copying data, then switching to Sheet2 to verify the operation, as illustrated in Figure \ref{fig:mcp-gemini-steps}. In contrast, the agent without MCP tools fails to select the specific column, demonstrating how tools can serve as a valuable complement to the agent's capabilities. More cases are analyzed in Appendix \ref{appendix:more-cases}.

% \input{iclr2026/chapter/chap5_analysis}

% \section{Experiments}
% \subsection{setup} 
% \subsection{Main results} 
% \subsection{Ablation Study} 
% \section{Results and Analysis} 
\section{Conclusion} 

We introduce OSWorld-MCP, a fair and comprehensive benchmark for evaluating Large Multimodal Models (LMMs) in computer-use scenarios by jointly assessing graphical user interface (GUI) operation skills and MCP tool-invocation capabilities. Using an automated pipeline combined with meticulous manual validation, we construct a high-quality and diverse collection of MCP tools that supports realistic and balanced evaluation within the OSWorld framework. Experiments on eight state-of-the-art LMMs demonstrate that tool invocation can substantially improve robustness and efficiency, while also revealing trade-offs between usage frequency and overall performance. Looking ahead, extending OSWorld-MCP to more complex, dynamic, and collaborative environments, as well as incorporating human-centred evaluation metrics, will further advance the development of general-purpose, efficient, and trustworthy computer-use agents.

\section*{Ethics statement}

The OSWorld-MCP dataset is constructed entirely from publicly available software environments and tasks in OSWorld. All MCP tools included in the benchmark are either automatically generated and manually validated by the authors or selected from existing open-source MCP servers, ensuring that no proprietary, confidential, or personally identifiable information is included. The dataset contains only synthetic interaction records between multimodal agents and computer application environments; no human subject data are collected. We release OSWorld-MCP solely for research and educational purposes to advance the evaluation of multimodal agents in realistic computer-use scenarios. Researchers using this dataset should comply with all applicable laws, institutional guidelines, and license terms. The authors bear responsibility for ensuring that the dataset is free of harmful or unethical content and that its use will not compromise privacy or security.

% \section*{Reproducibility statement}

% We will release all resources necessary to reproduce our work, including the OSWorld-MCP dataset, the complete set of 158 validated MCP tools, and the automated code-generation pipeline used to create them. Detailed documentation will be provided to ensure that researchers can replicate our experiments under the same conditions, including task definitions, maximum step settings, and all metric implementations (Accuracy, Tool Invocation Rate, and Average Completion Steps). All resources will be hosted in an open-access repository upon publication.
\bibliography{iclr2026_conference}
\bibliographystyle{iclr2026_conference}

\appendix
\newpage
\section{Appendix}

\subsection{More Cases}
\label{appendix:more-cases}

We present and analyze additional tool call examples to demonstrate how tool calls enable agents to complete diverse and complex tasks more efficiently and accurately.

\newcommand{\stepcaption}[1]{%
  \vspace{0.4em} % TODO: can we use \vspace?
  \parbox[t]{\linewidth}{%
    % \ttfamily
    \scriptsize
    \raggedright
    #1
  }
  % \vspace{0.2em} % TODO: can we use \vspace?
}

\subsubsection{Case: VsCode Settings}

The goal of the task is ``Please help me modify VS Code setting to hide all `\_\_pycache\_\_' folders in the explorer view''. Executed actions are as follows:

\begin{enumerate}
    \item \texttt{osworld\_mcp\_code.add\_files\_exclude}
    \begin{itemize}
        \item \texttt{pattern="**/\_\_pycache\_\_"}
    \end{itemize}

    \item \texttt{computer\_use}
    \begin{itemize}
        \item \texttt{action="left\_click"}
        \item \texttt{x=1131.0}
        \item \texttt{y=712.5}
    \end{itemize}
    
    \item \texttt{computer\_use}
    \begin{itemize}
        \item \texttt{action="terminate"}
        \item \texttt{status="success"}
    \end{itemize}
\end{enumerate}

\begin{figure}[!htbp]
    \centering
    \setlength{\tabcolsep}{4pt}

    \begin{tabular}{cc}
        % Step 0
        \begin{minipage}{0.48\linewidth}
            \stepcaption{Initial state: {\texttt{reset()}}}
            \includegraphics[width=\linewidth]{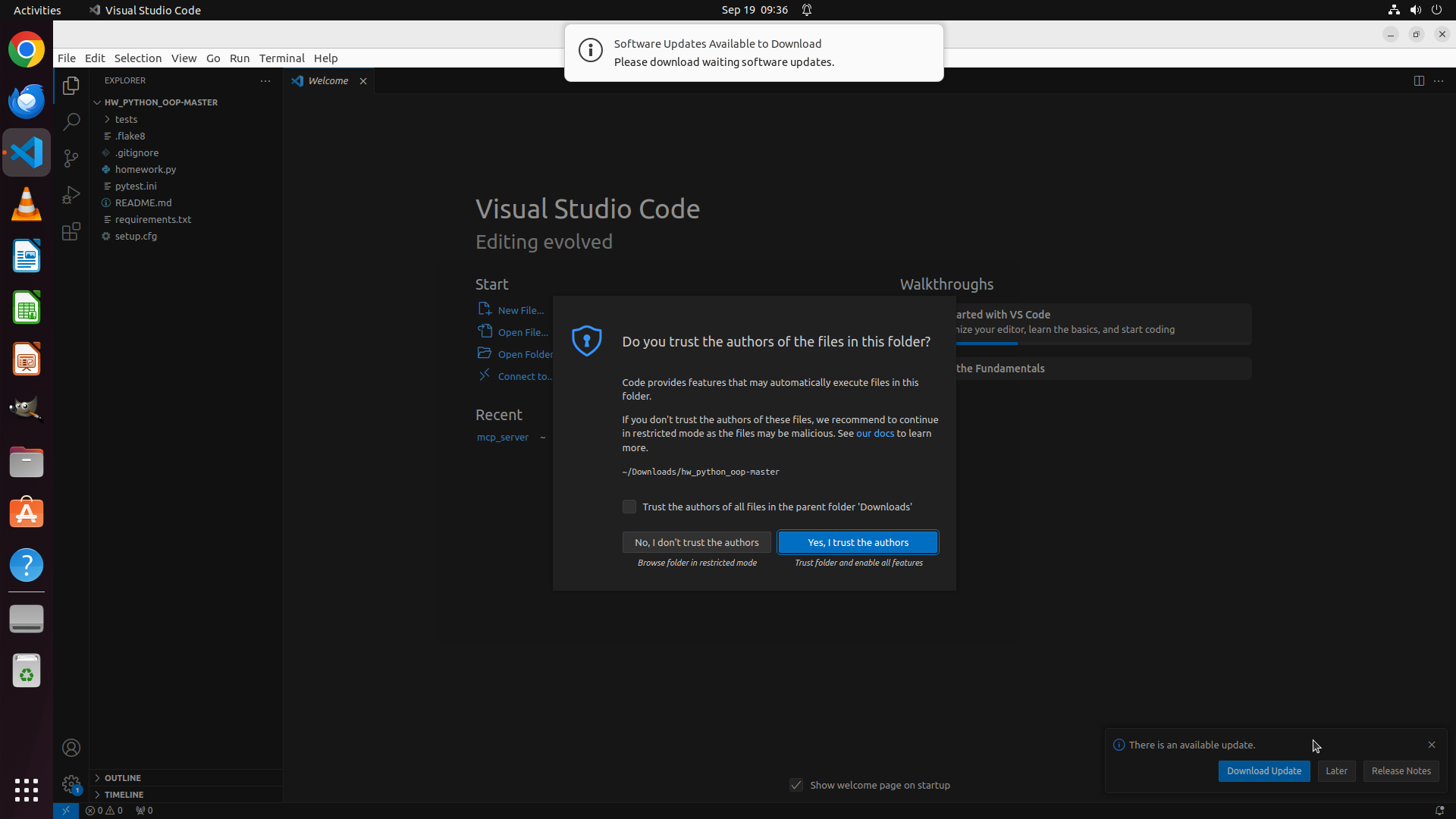}
        \end{minipage} &
        % Step 1
        \begin{minipage}{0.48\linewidth}
            \stepcaption{Step 1: {\texttt{osworld\_mcp\_code.add\_files\_exclude(p...}}}
            \includegraphics[width=\linewidth]{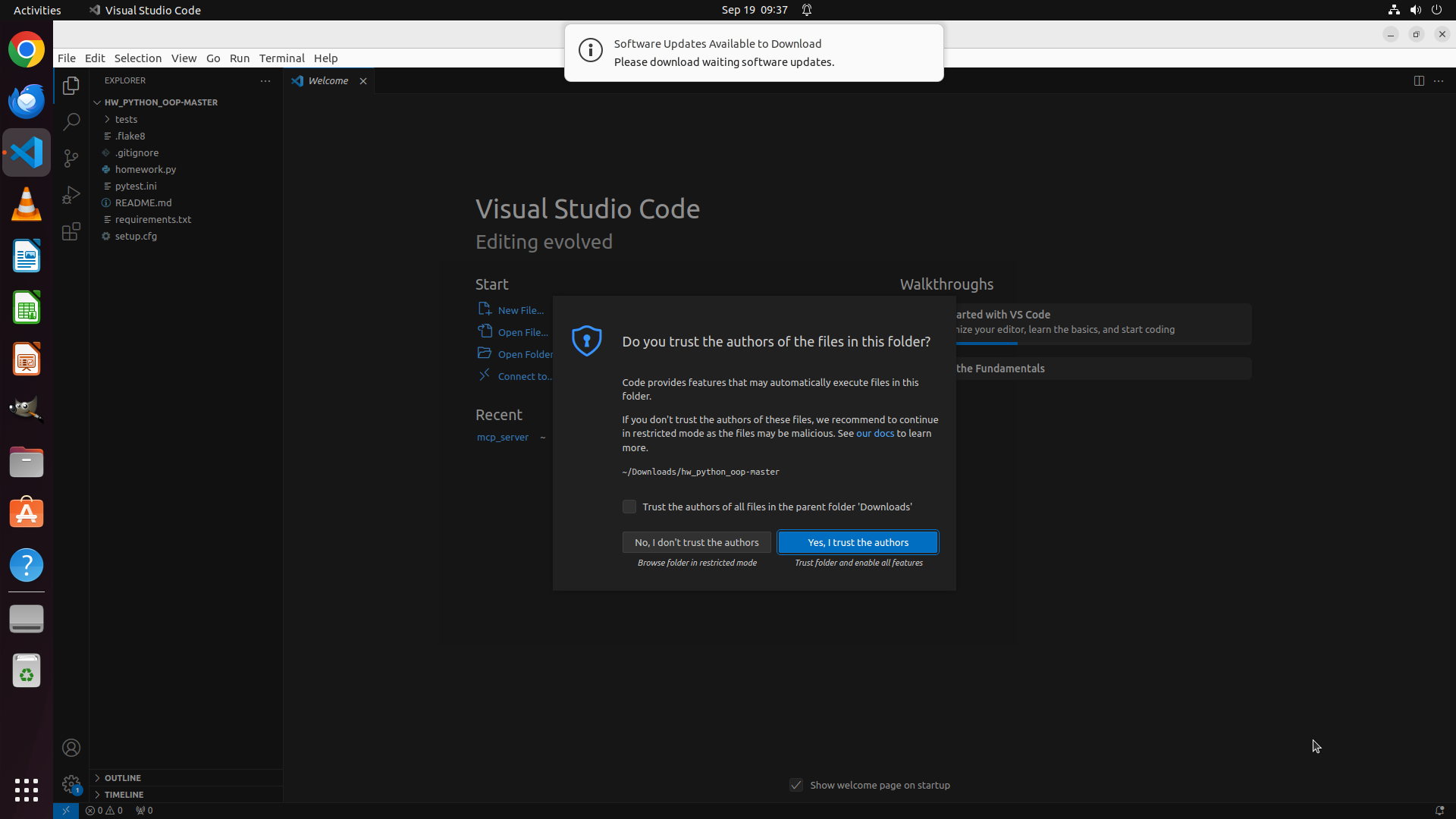}
        \end{minipage} \\
        % Step 2
        \begin{minipage}{0.48\linewidth}
            \stepcaption{Step 2: {\texttt{computer\_use(action="left click", x=...}}}
            \includegraphics[width=\linewidth]{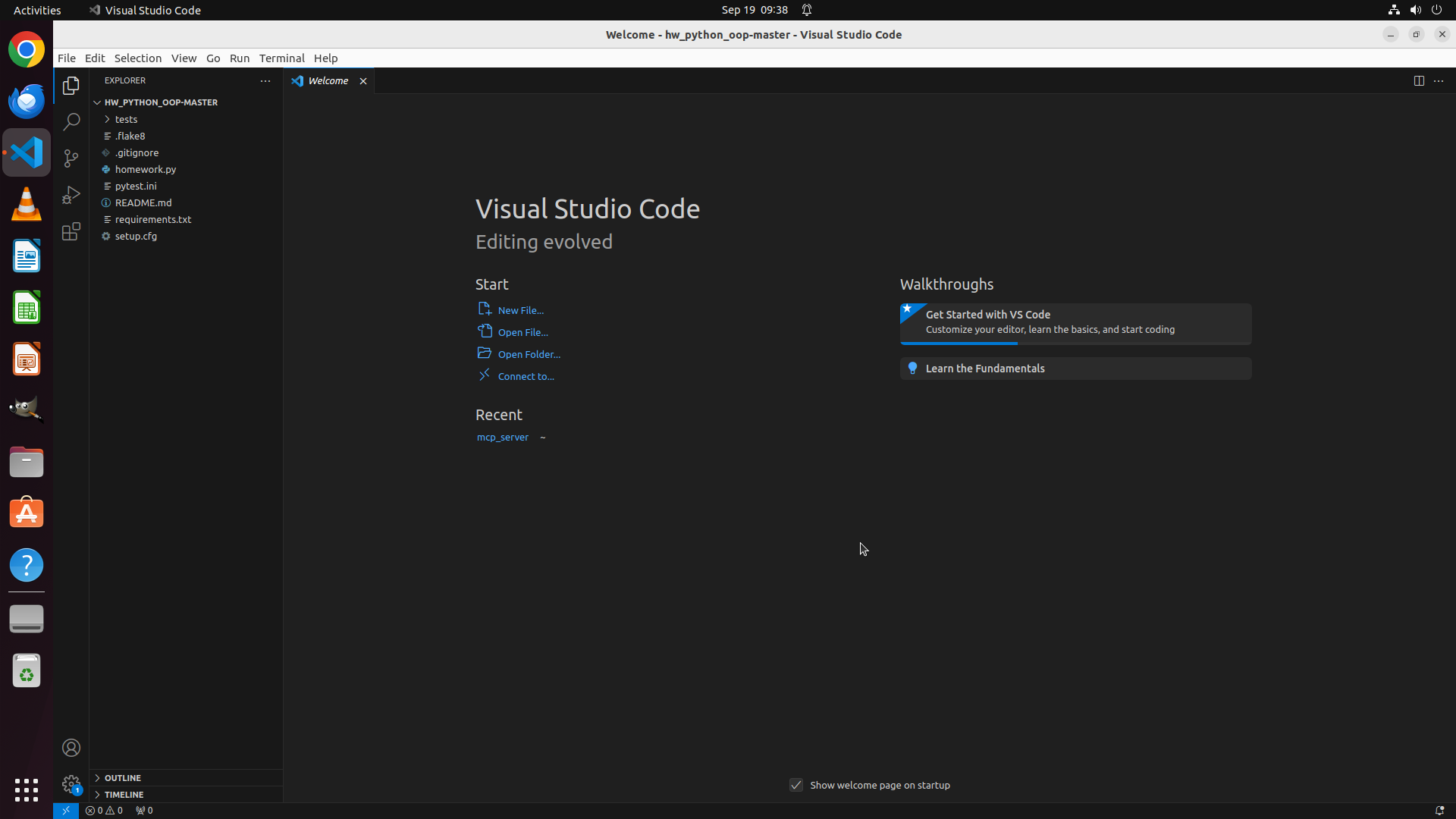}
        \end{minipage} &
        % Step 3
        \begin{minipage}{0.48\linewidth}
            \stepcaption{Step 3: {\texttt{computer\_use(action="terminate", sta...}}}
            \includegraphics[width=\linewidth]{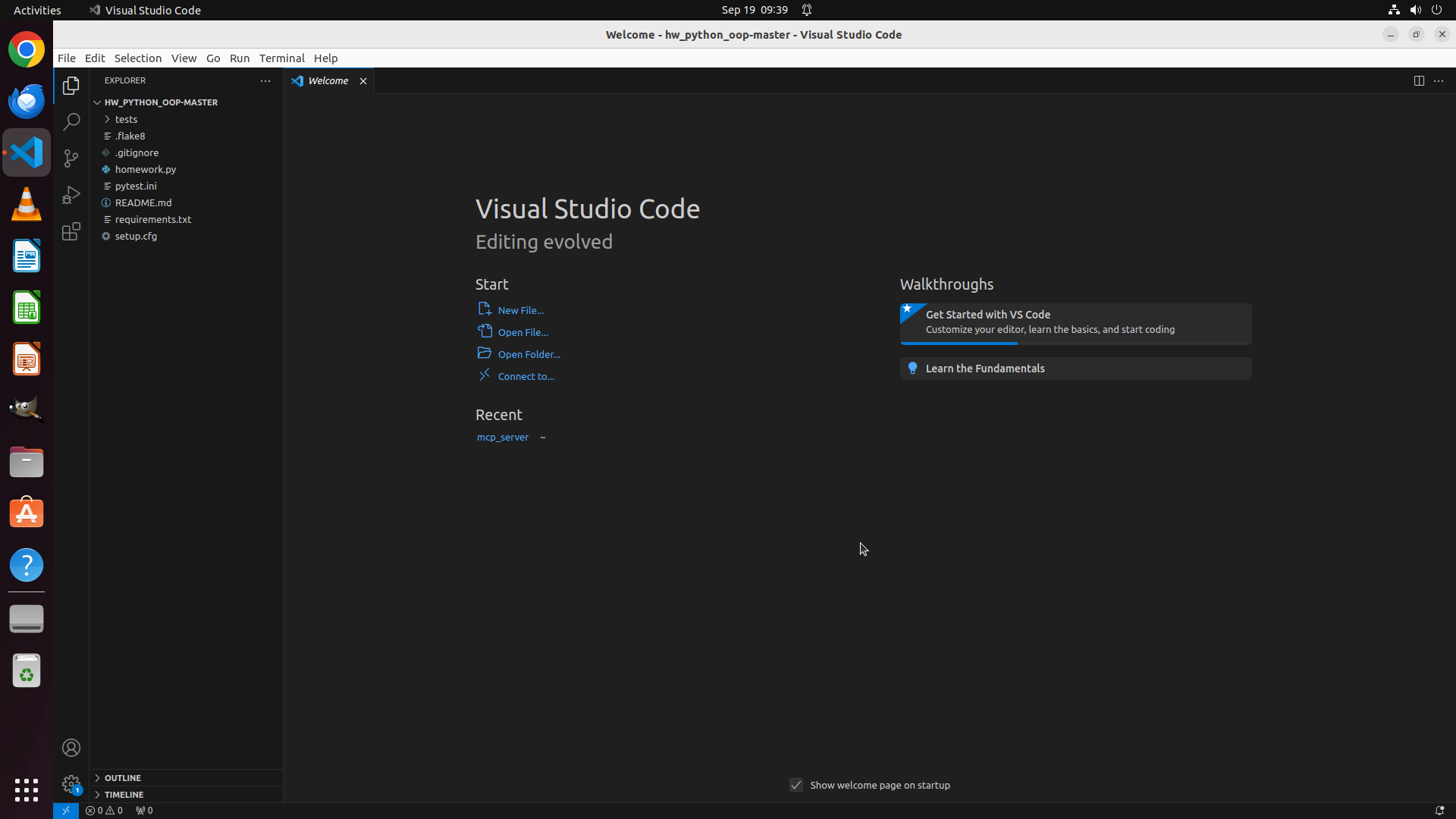}
        \end{minipage} \\
    \end{tabular}

    \caption{Please help me modify VS Code setting to hide all ``\_\_pycache\_\_'' folders in the explorer view.}
    \label{fig:mcp-code-steps}
\end{figure}

Figure \ref{fig:mcp-code-steps} showcases the executing process. According to the evaluation program, the specific folder is hidden in the explorer view.

% \newpage

\subsubsection{Case: Filesystem Operation}

The goal of the task is ``I have an image of my receipt located in /home/user. I'm looking to transform it into a PDF file. Can you assist me with this task? Save the resulting PDF as `receipt.pdf' on the desktop''. Executed actions are as follows:

\begin{enumerate}
    \item \texttt{osworld\_mcp\_os.convert\_image\_format}
    \begin{itemize}
        \item \texttt{image\_path="/home/user/receipt.jpg"}
        \item \texttt{output\_format="pdf"}
        \item \texttt{output\_path="/home/user/Desktop/receipt.pdf"}
    \end{itemize}

    \item \texttt{filesystem\_list\_directory}
    \begin{itemize}
        \item \texttt{path="/home/user"}
    \end{itemize}

    \item \texttt{osworld\_mcp\_os.convert\_image\_format}
    \begin{itemize}
        \item \texttt{image\_path="/home/user/OIP.jpg"}
        \item \texttt{output\_format="pdf"}
        \item \texttt{output\_path="/home/user/Desktop/receipt.pdf"}
    \end{itemize}
    
    \item \texttt{computer\_use}
    \begin{itemize}
        \item \texttt{action="terminate"}
        \item \texttt{status="success"}
    \end{itemize}
\end{enumerate}

% The agent utilized tool to convert image to other format like PDF, while the tool response indicated there are no such file or directory. By listing the file in the home path, the correct file name is obtained by agent, and the 再次的结果 eventually produce desired PDF file from the original image, as demonstrated in Figure xxx.

\begin{figure}[!htbp]
    \centering
    \setlength{\tabcolsep}{4pt}

    \begin{tabular}{cc}
        % Step 0
        \begin{minipage}{0.48\linewidth}
            \stepcaption{Initial state: {\texttt{reset()}}}
            \includegraphics[width=\linewidth]{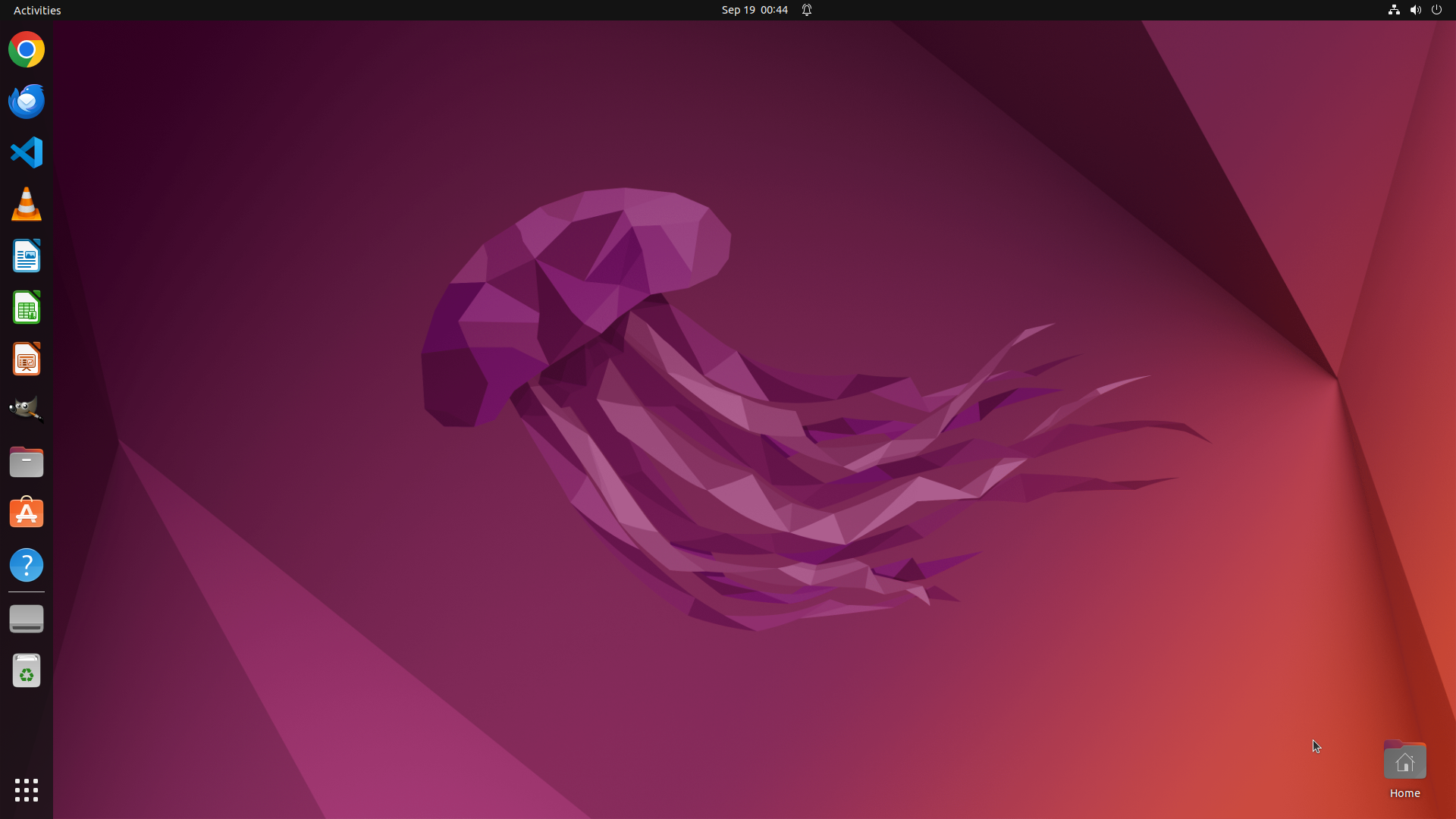}
        \end{minipage} &
        % Step 1
        \begin{minipage}{0.48\linewidth}
            \stepcaption{Step 1: {\texttt{osworld\_mcp\_os.convert\_image\_format(...}}}
            \includegraphics[width=\linewidth]{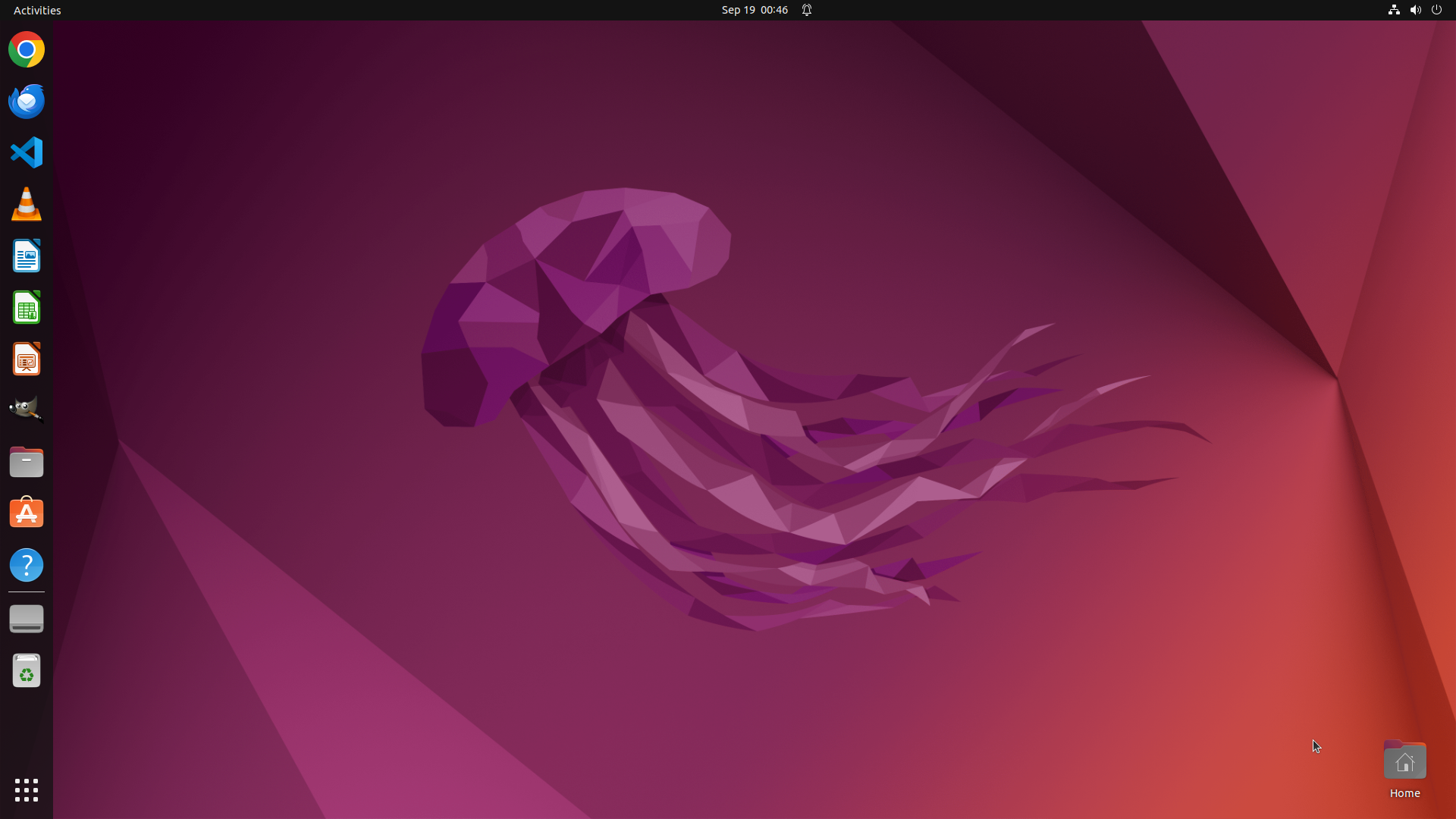}
        \end{minipage} \\
        % Step 2
        \begin{minipage}{0.48\linewidth}
            \stepcaption{Step 2: {\texttt{filesystem\_list\_directory(path="/hom...}}}
            \includegraphics[width=\linewidth]{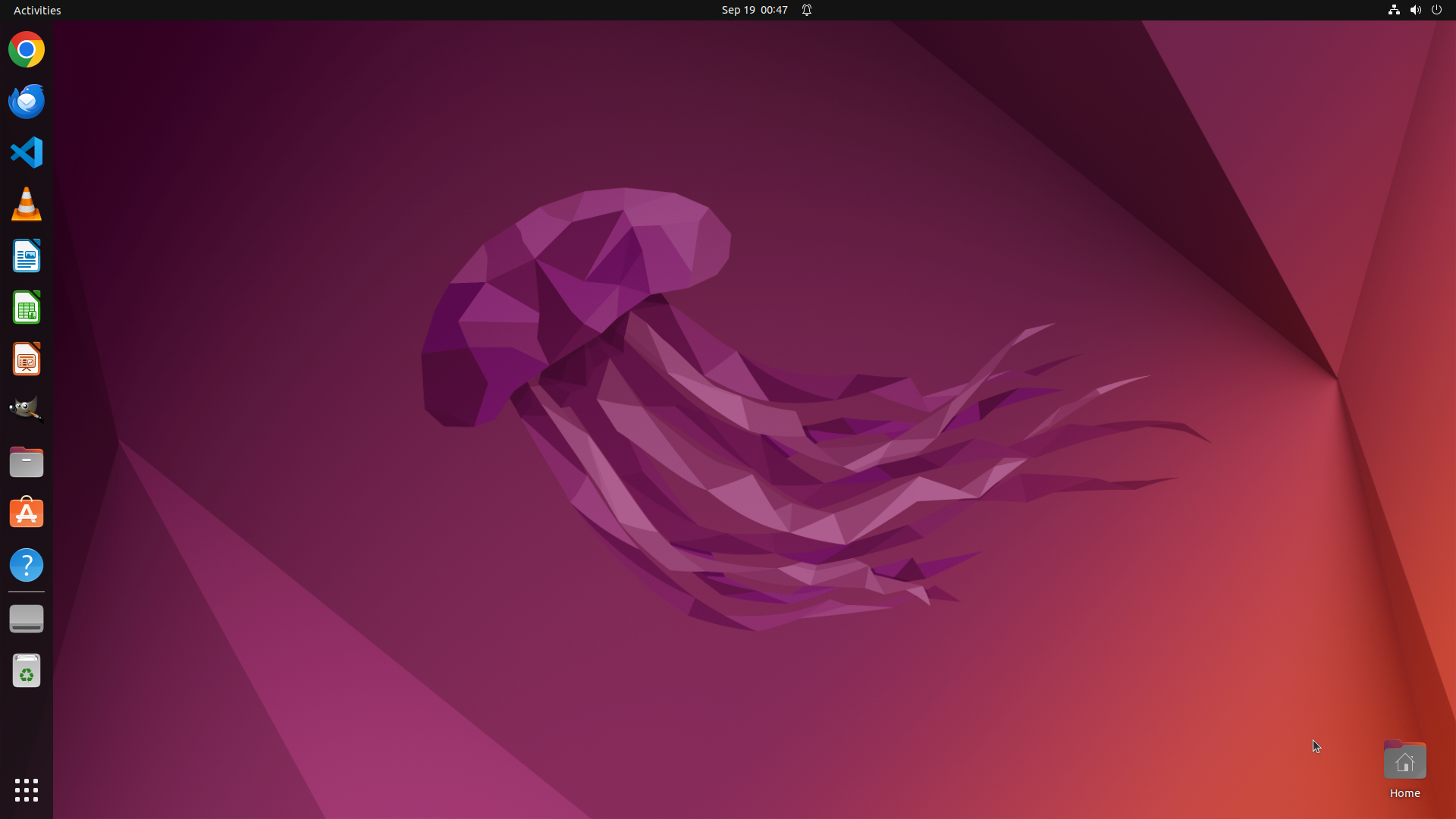}
        \end{minipage} &
        % Step 3
        \begin{minipage}{0.48\linewidth}
            \stepcaption{Step 3: {\texttt{osworld\_mcp\_os.convert\_image\_format(...}}}
            \includegraphics[width=\linewidth]{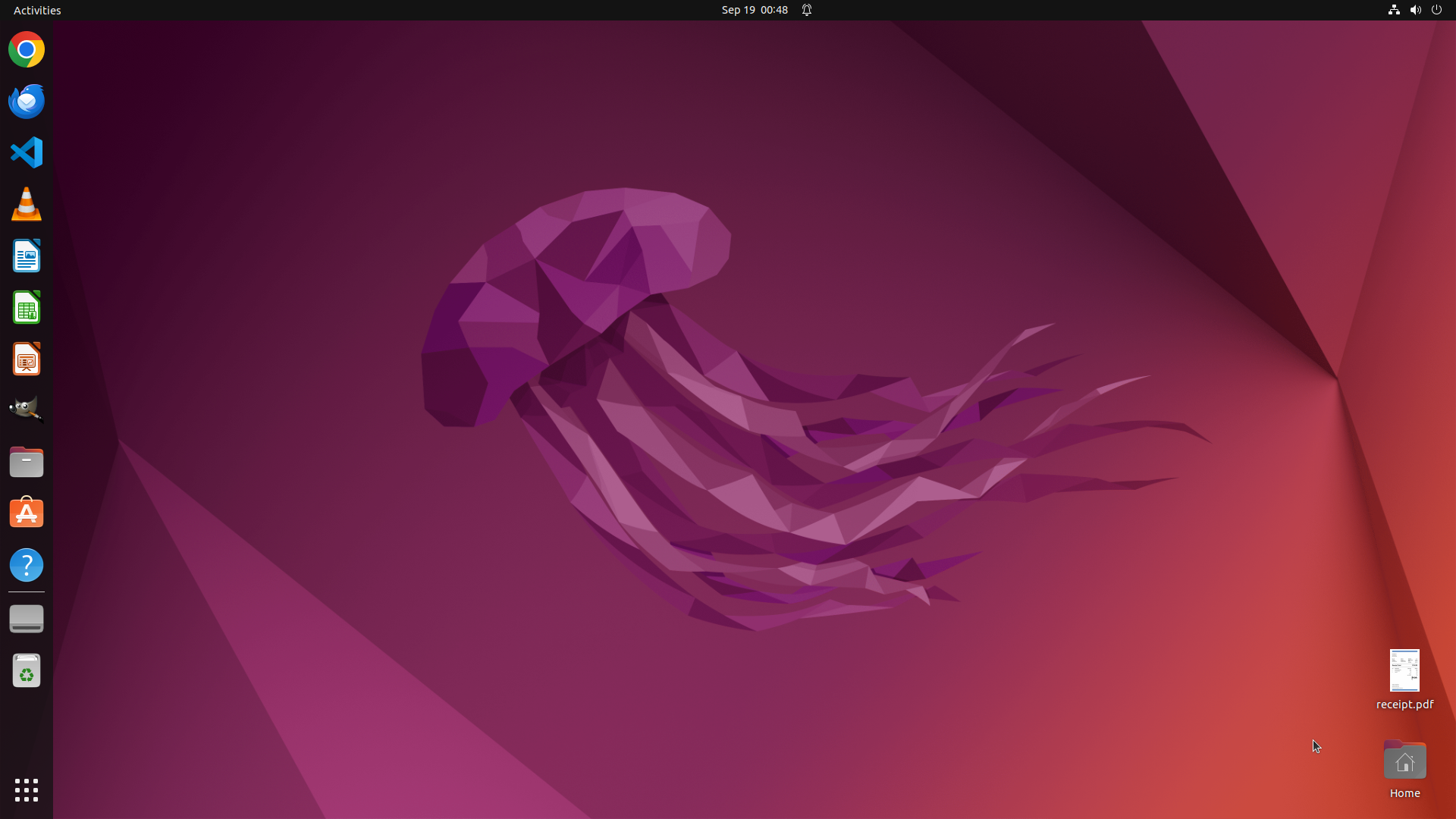}
        \end{minipage} \\
        % Step 4
        \begin{minipage}{0.48\linewidth}
            \stepcaption{Step 4: {\texttt{computer\_use(action="terminate", sta...}}}
            \includegraphics[width=\linewidth]{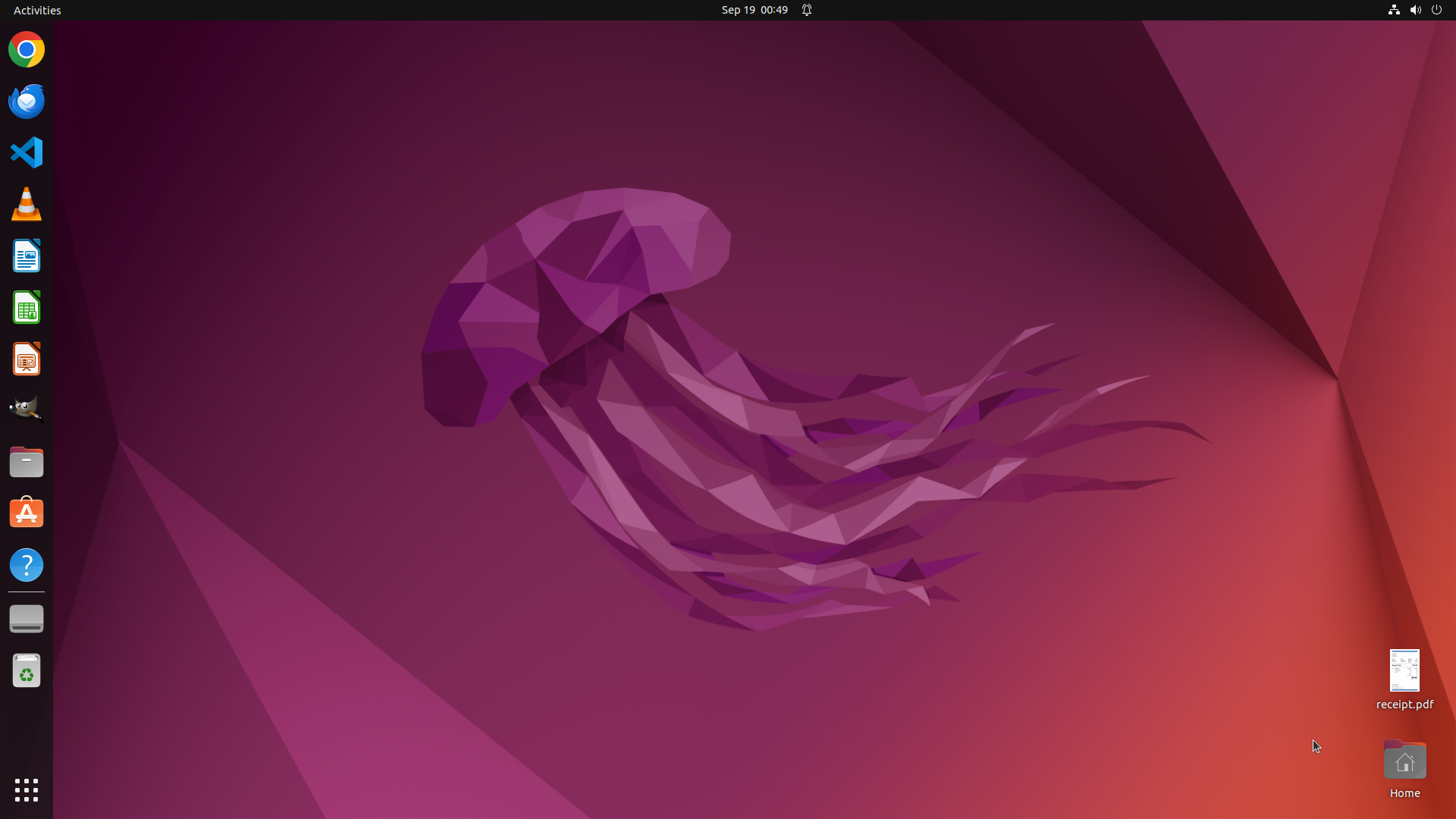}
        \end{minipage} &
    \end{tabular}

    \caption{I have an image of my receipt located in /home/user. I'm looking to transform it into a PDF file. Can you assist me with this task? Save the resulting PDF as ``receipt.pdf'' on the desktop.}
    \label{fig:mcp-os-steps}
\end{figure}

The agent first utilized the tool to convert an image to PDF format, but the tool response indicated the specified file or directory did not exist. By listing the files in the home directory path, the agent obtained the correct filename. The subsequent conversion attempt successfully produced the desired PDF file from the original image, as demonstrated in Figure \ref{fig:mcp-os-steps}.

% \newpage

\subsubsection{Case: Slide Duplication}

The goal of the task is ``Please duplicate the last two slides and insert the copies in alternating order, so the sequence becomes: original slide A, original slide B, then duplicated slide A, duplicated slide B''. Executed actions are as follows:

\begin{enumerate}
    \item \texttt{osworld\_mcp\_libreoffice\_impress.get\_slide\_count}

    \item \texttt{osworld\_mcp\_libreoffice\_impress.duplicate\_slide}
    \begin{itemize}
        \item \texttt{slide\_index=24}
        \item \texttt{target\_position=25}
    \end{itemize}

    \item \texttt{osworld\_mcp\_libreoffice\_impress.get\_slide\_count}

    \item \texttt{osworld\_mcp\_libreoffice\_impress.duplicate\_slide}
    \begin{itemize}
        \item \texttt{slide\_index=24}
        \item \texttt{target\_position=25}
    \end{itemize}

    \item \texttt{osworld\_mcp\_libreoffice\_impress.duplicate\_slide}
    \begin{itemize}
        \item \texttt{slide\_index=23}
        \item \texttt{target\_position=24}
    \end{itemize}

    \item \texttt{osworld\_mcp\_libreoffice\_impress.duplicate\_slide}
    \begin{itemize}
        \item \texttt{slide\_index=24}
        \item \texttt{target\_position=25}
    \end{itemize}

    \item \texttt{osworld\_mcp\_libreoffice\_impress.save}
    
    \item \texttt{computer\_use}
    \begin{itemize}
        \item \texttt{action="terminate"}
        \item \texttt{status="success"}
    \end{itemize}
\end{enumerate}

\begin{figure}[!htbp]
    \centering
    \setlength{\tabcolsep}{4pt}

    \begin{tabular}{cc}
        % Step 0
        \begin{minipage}{0.48\linewidth}
            \stepcaption{Initial state: {\texttt{reset()}}}
            \includegraphics[width=\linewidth]{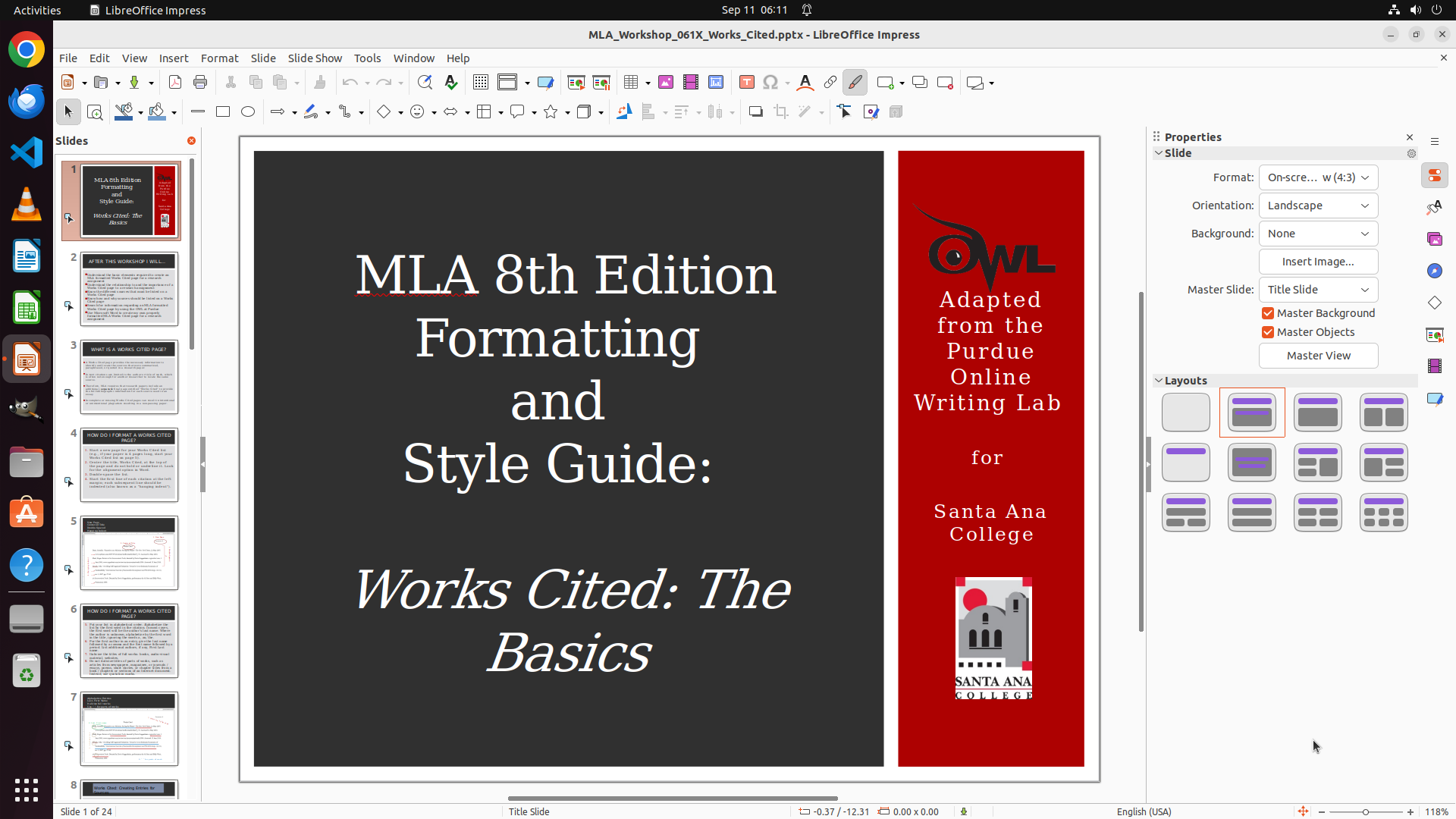}
        \end{minipage} &
        % Step 1
        \begin{minipage}{0.48\linewidth}
            \stepcaption{Step 1: {\texttt{osworld\_mcp\_libreoffice\_impress.get\_...}}}
            \includegraphics[width=\linewidth]{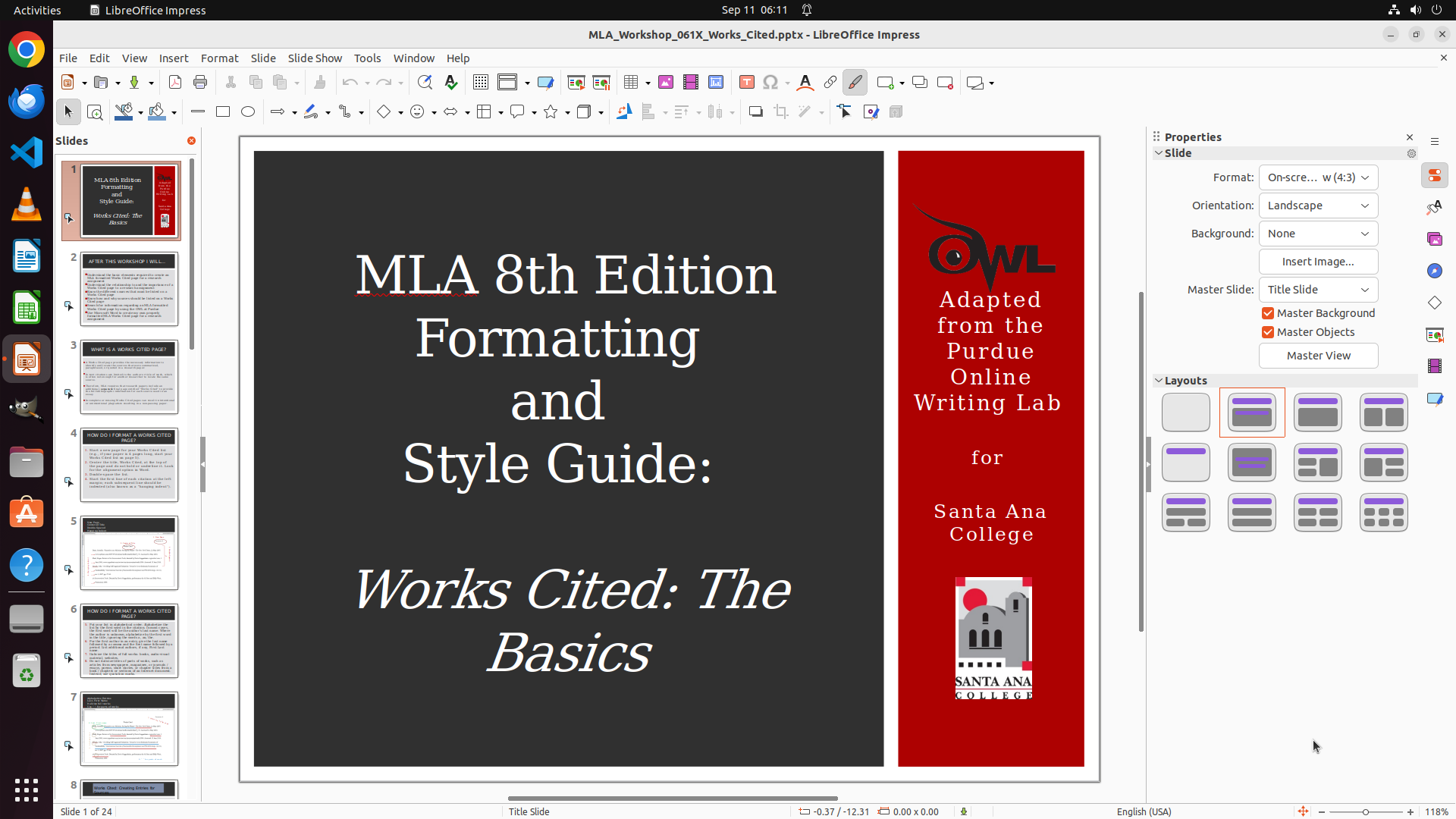}
        \end{minipage} \\
        % Step 2
        \begin{minipage}{0.48\linewidth}
            \stepcaption{Step 2: {\texttt{osworld\_mcp\_libreoffice\_impress.dupl...}}}
            \includegraphics[width=\linewidth]{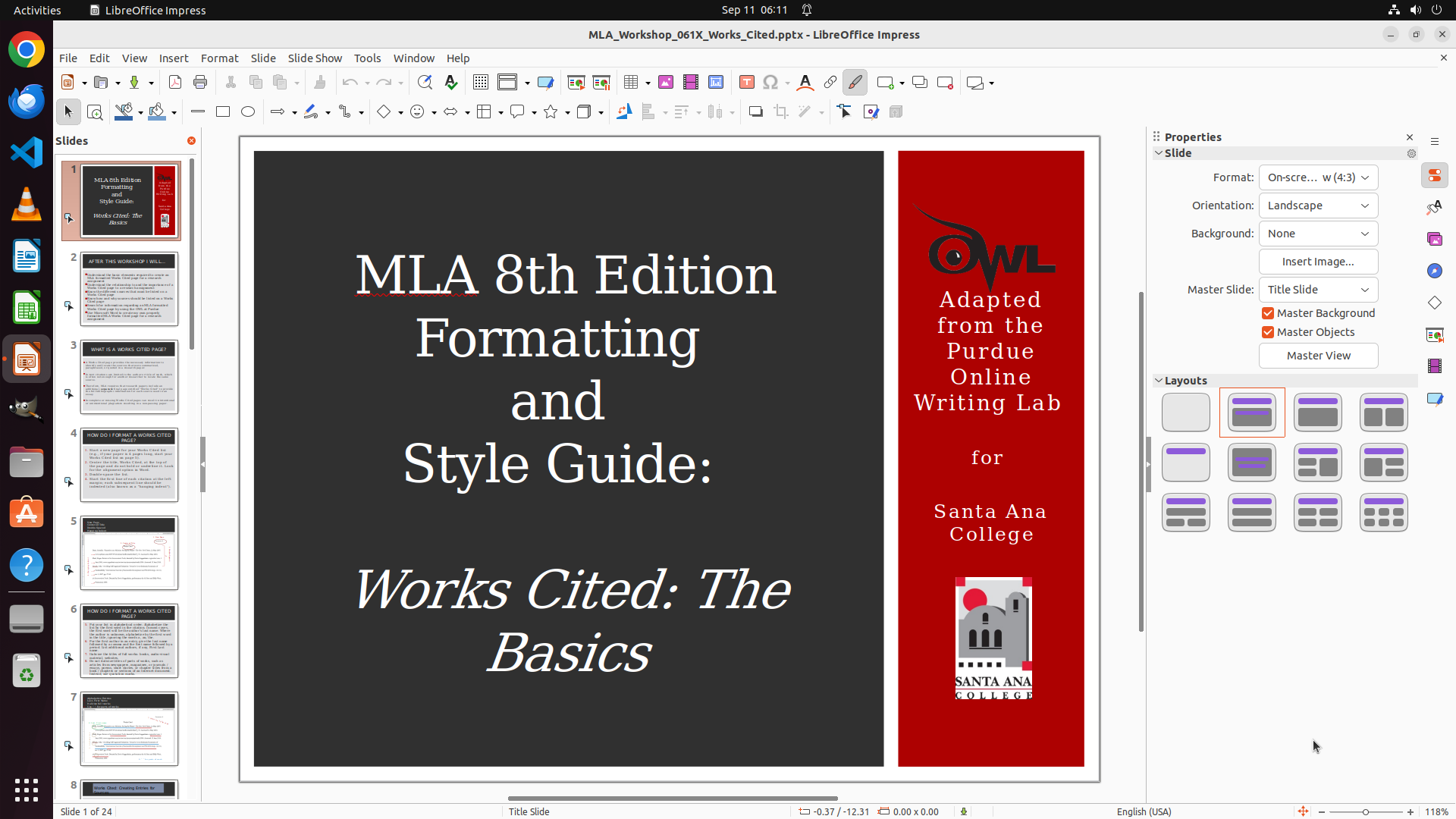}
        \end{minipage} &
        % Step 3
        \begin{minipage}{0.48\linewidth}
            \stepcaption{Step 3: {\texttt{osworld\_mcp\_libreoffice\_impress.get\_...}}}
            \includegraphics[width=\linewidth]{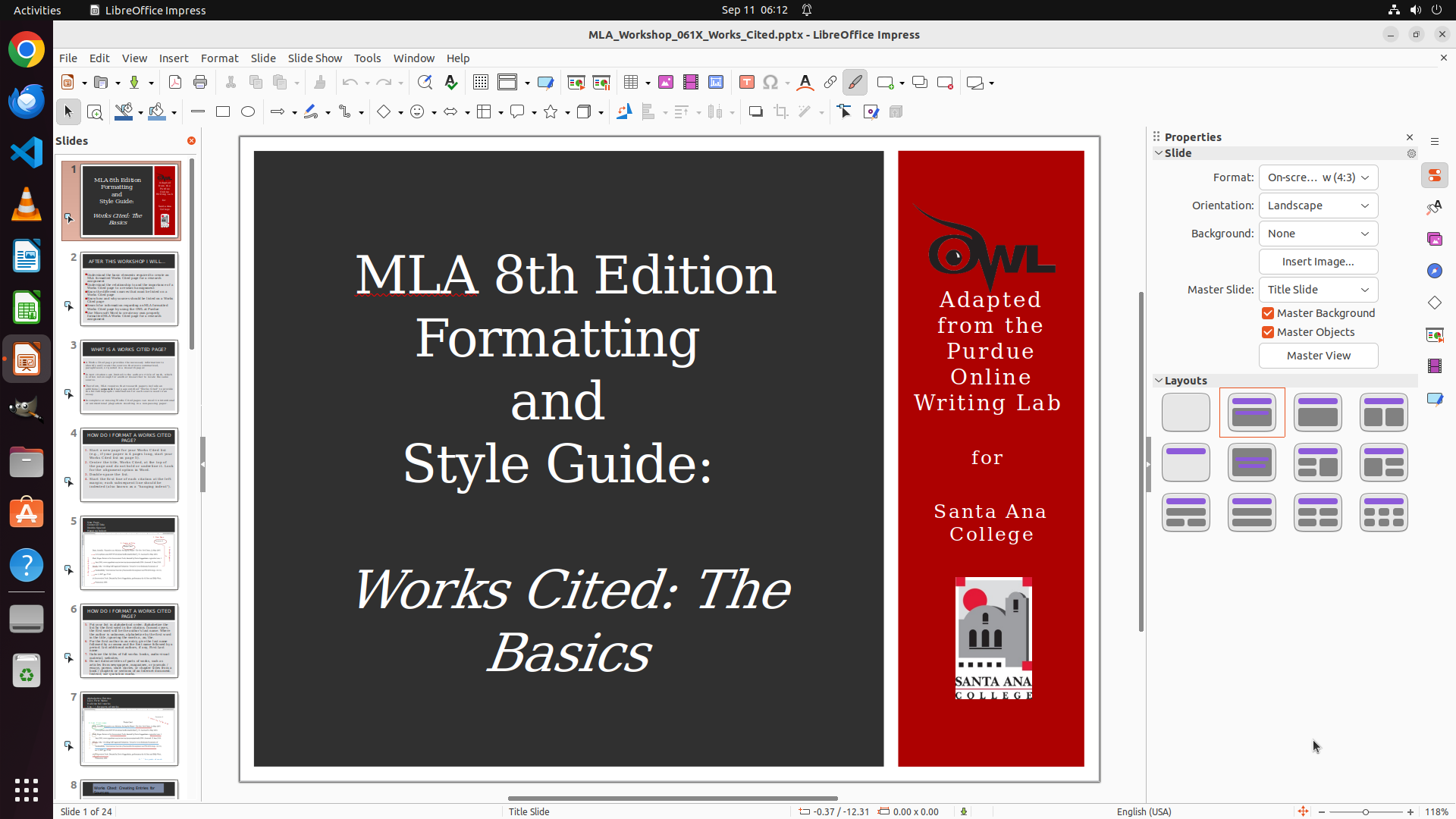}
        \end{minipage} \\
        % Step 4
        \begin{minipage}{0.48\linewidth}
            \stepcaption{Step 4: {\texttt{osworld\_mcp\_libreoffice\_impress.dupl...}}}
            \includegraphics[width=\linewidth]{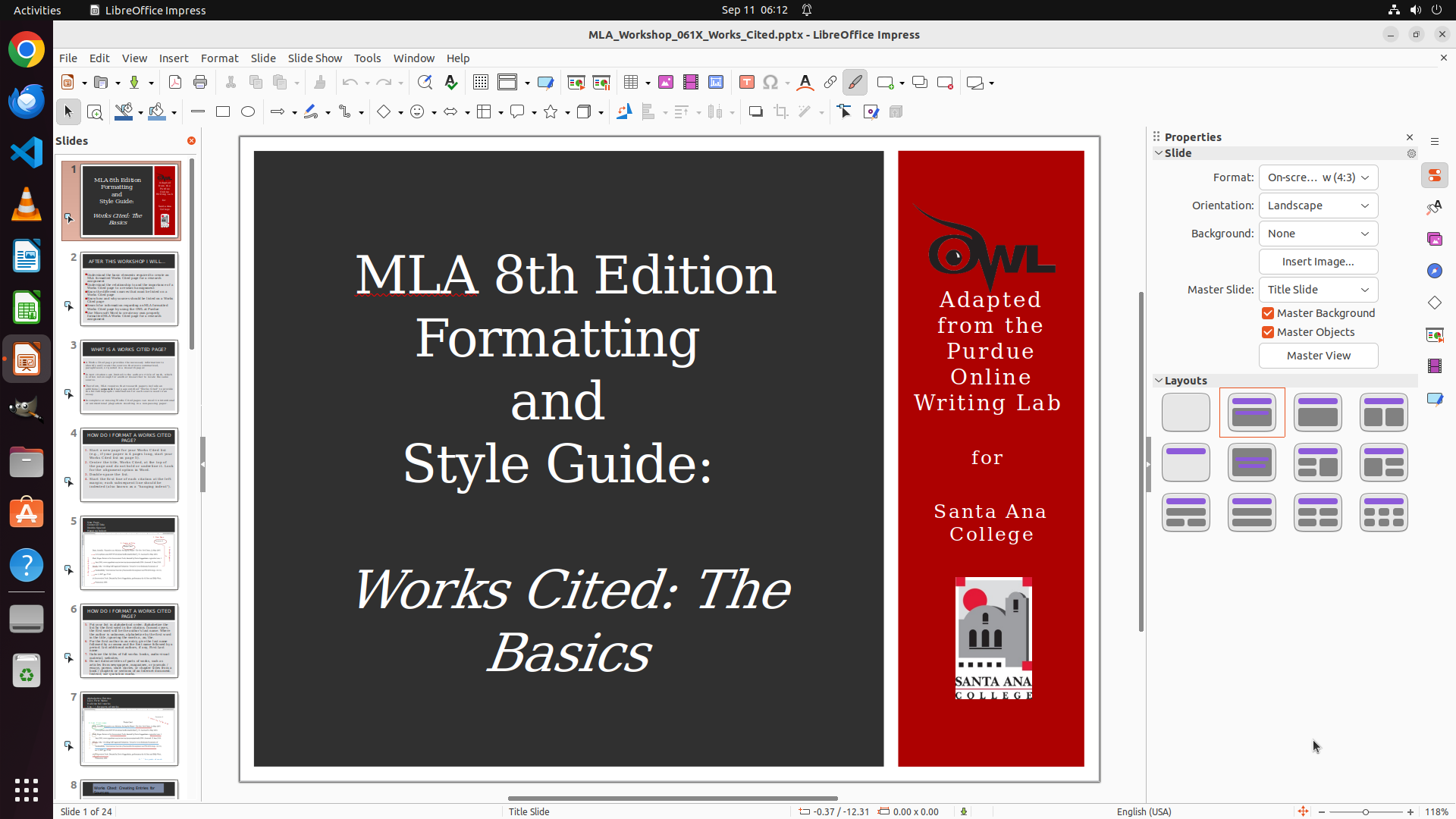}
        \end{minipage} &
        % Step 5
        \begin{minipage}{0.48\linewidth}
            \stepcaption{Step 5: {\texttt{osworld\_mcp\_libreoffice\_impress.dupl...}}}
            \includegraphics[width=\linewidth]{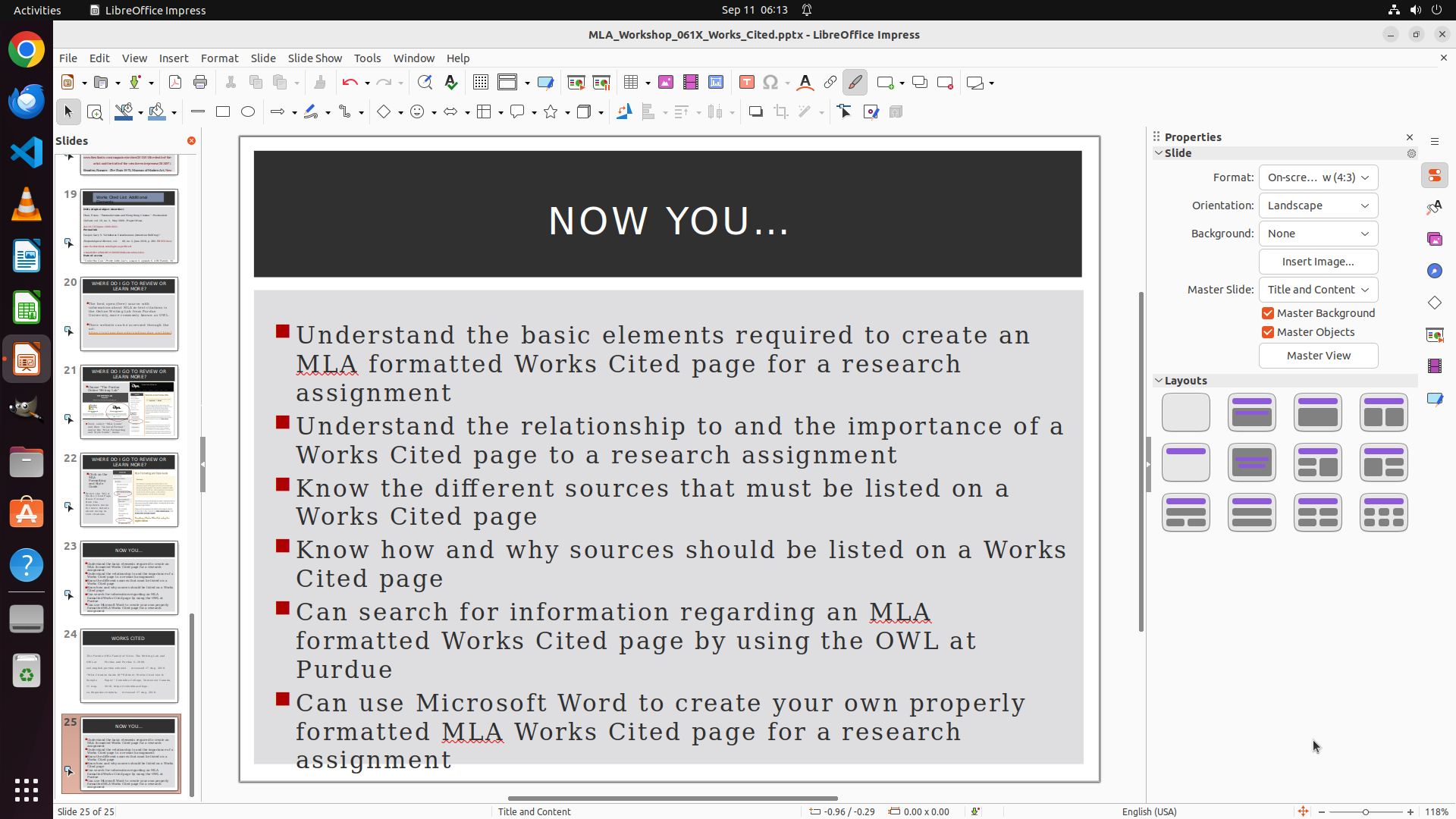}
        \end{minipage} \\
        % Step 6
        \begin{minipage}{0.48\linewidth}
            \stepcaption{Step 6: {\texttt{osworld\_mcp\_libreoffice\_impress.dupl...}}}
            \includegraphics[width=\linewidth]{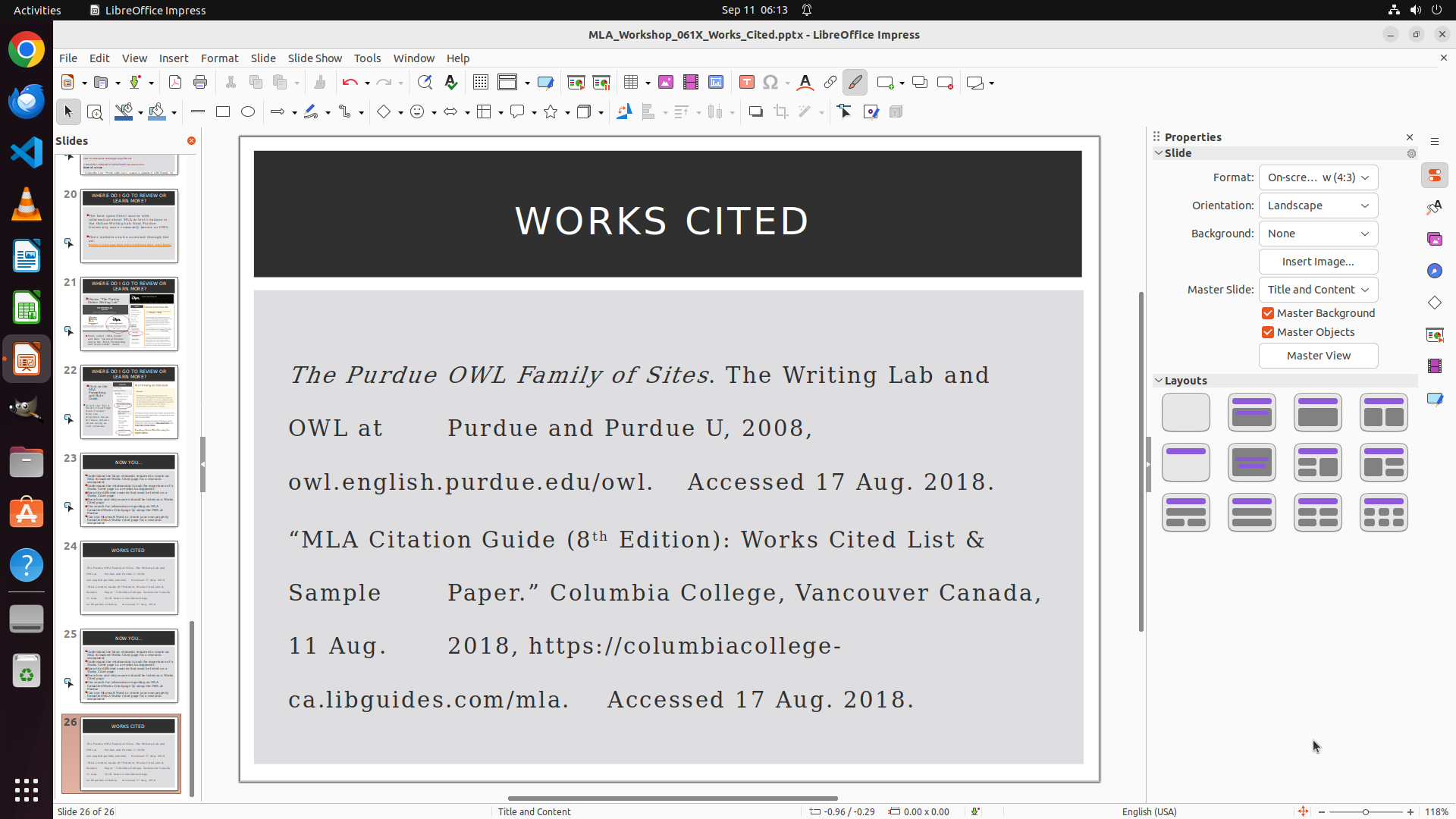}
        \end{minipage} &
        % Step 7
        \begin{minipage}{0.48\linewidth}
            \stepcaption{Step 7: {\texttt{osworld\_mcp\_libreoffice\_impress.save...}}}
            \includegraphics[width=\linewidth]{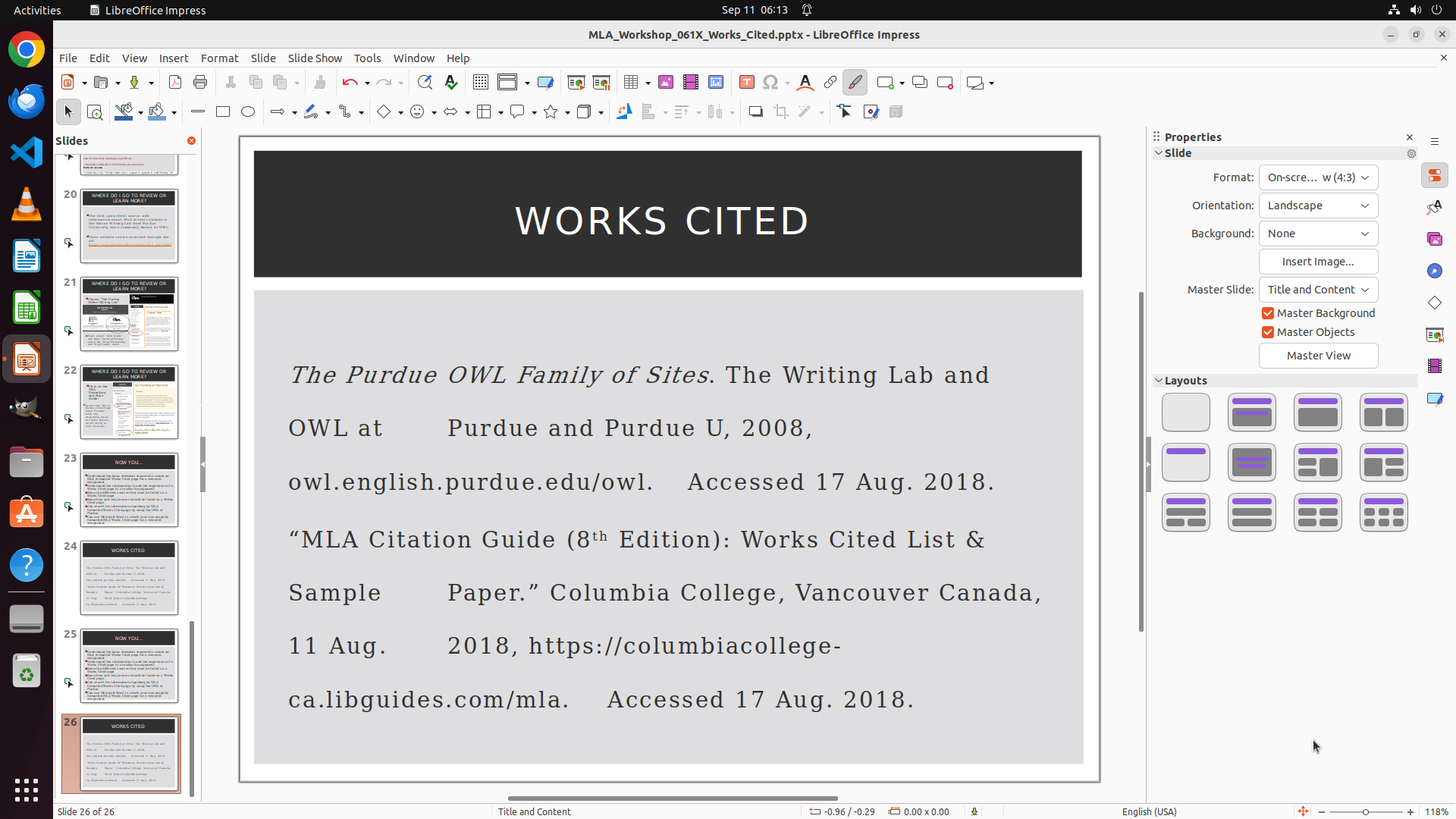}
        \end{minipage} \\
        % Step 8
        \begin{minipage}{0.48\linewidth}
            \stepcaption{Step 8: {\texttt{computer\_use(action="terminate", sta...}}}
            \includegraphics[width=\linewidth]{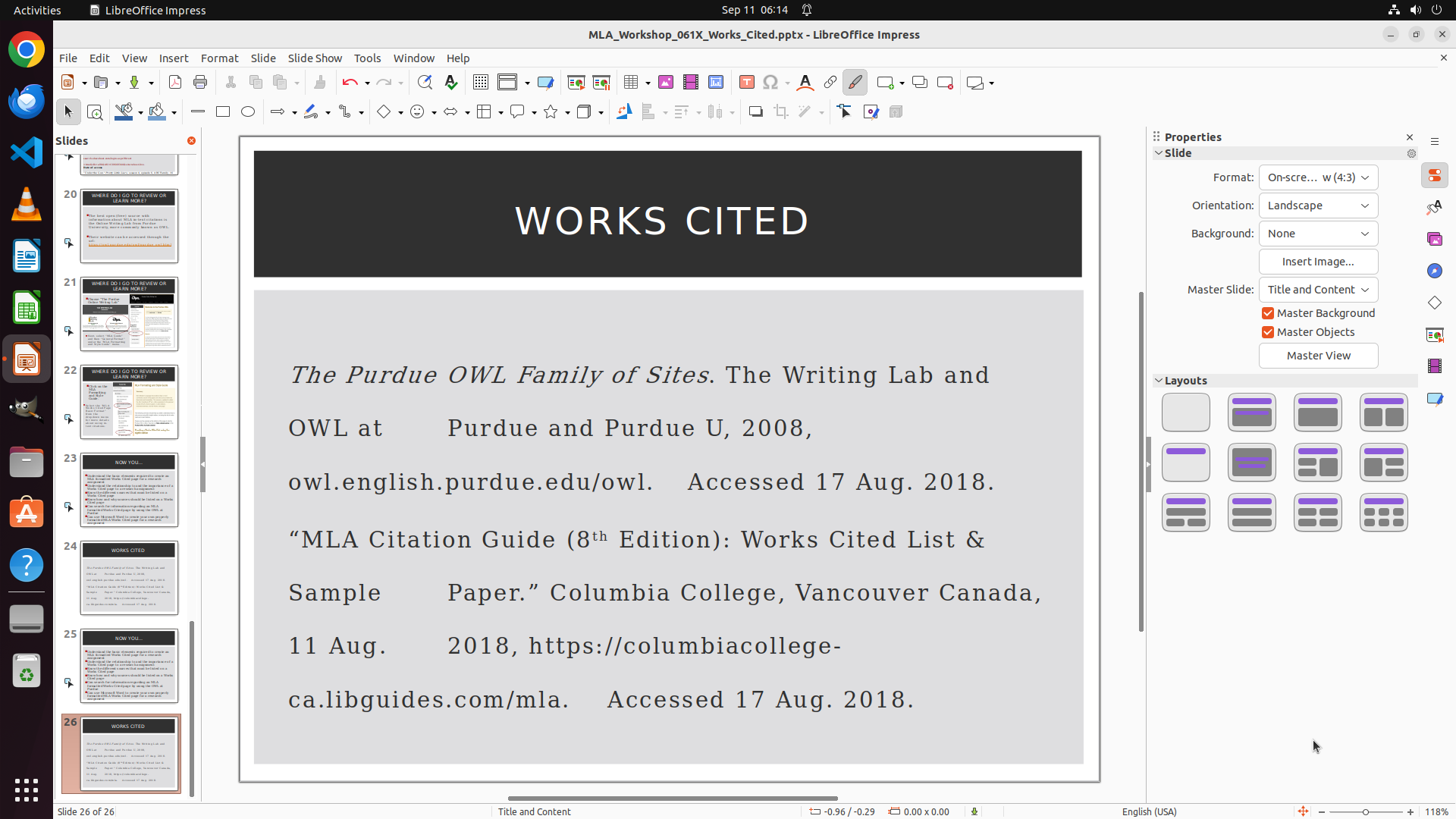}
        \end{minipage} &
    \end{tabular}

    \caption{Please duplicate the last two slides and insert the copies in alternating order, so the sequence becomes: original slide A, original slide B, then duplicated slide A, duplicated slide B.}
    \label{fig:mcp-impress-steps}
\end{figure}

The agent initially determined the positions of the last two slides (indexed 23 and 24) through tool calls and attempted to directly duplicate the slides. However, the operation failed at steps 2 and 4 due to the incorrect parameter of target position. After several attempts, the agent successfully duplicated the final two slides to positions 25 at step 5, and 26 at step 6. As shown in Figure \ref{fig:mcp-impress-steps}, the agent ultimately completing the task.

\newpage

\subsubsection{Case: Spreadsheet Copying}

The goal of the task is ``Copy the `Revenue' column along with the header to a new sheet named `Sheet2'\,''. Executed actions are as follows:

\begin{enumerate}
    \item \texttt{osworld\_mcp\_libreoffice\_calc.copy\_cells\_between\_sheets}
    \begin{itemize}
        \item \texttt{source\_range="A1:A20"}
        \item \texttt{target\_sheet\_name="Sheet2"}
        \item \texttt{target\_start\_cell="A1"}
    \end{itemize}

    \item \texttt{osworld\_mcp\_libreoffice\_calc.copy\_cells\_between\_sheets}
    \begin{itemize}
        \item \texttt{source\_range="A1:A20"}
        \item \texttt{target\_sheet\_name="Sheet2"}
        \item \texttt{target\_start\_cell="A1"}
    \end{itemize}

    \item \texttt{osworld\_mcp\_libreoffice\_calc.get\_workbook\_info}

    \item \texttt{osworld\_mcp\_libreoffice\_calc.copy\_cells\_between\_sheets}
    \begin{itemize}
        \item \texttt{source\_range="A1:A20"}
        \item \texttt{target\_sheet\_name="Sheet2"}
        \item \texttt{target\_start\_cell="A1"}
    \end{itemize}

    \item \texttt{osworld\_mcp\_libreoffice\_calc.switch\_active\_sheet}
    \begin{itemize}
        \item \texttt{sheet\_name="Sheet2"}
    \end{itemize}

    \item \texttt{osworld\_mcp\_libreoffice\_calc.switch\_active\_sheet}
    \begin{itemize}
        \item \texttt{sheet\_name="Sheet1"}
    \end{itemize}

    \item \texttt{osworld\_mcp\_libreoffice\_calc.copy\_cells\_between\_sheets}
    \begin{itemize}
        \item \texttt{source\_range="A1:A20"}
        \item \texttt{target\_sheet\_name="Sheet2"}
        \item \texttt{target\_start\_cell="A1"}
    \end{itemize}

    \item \texttt{osworld\_mcp\_libreoffice\_calc.switch\_active\_sheet}
    \begin{itemize}
        \item \texttt{sheet\_name="Sheet2"}
    \end{itemize}

    \item \texttt{computer\_use}
    \begin{itemize}
        \item \texttt{action="terminate"}
        \item \texttt{status="success"}
    \end{itemize}
\end{enumerate}

\begin{figure}[!htbp]
    \centering
    \setlength{\tabcolsep}{4pt}

    \begin{tabular}{cc}
        % Step 0
        \begin{minipage}{0.48\linewidth}
            \stepcaption{Initial state: {\texttt{reset()}}}
            \includegraphics[width=\linewidth]{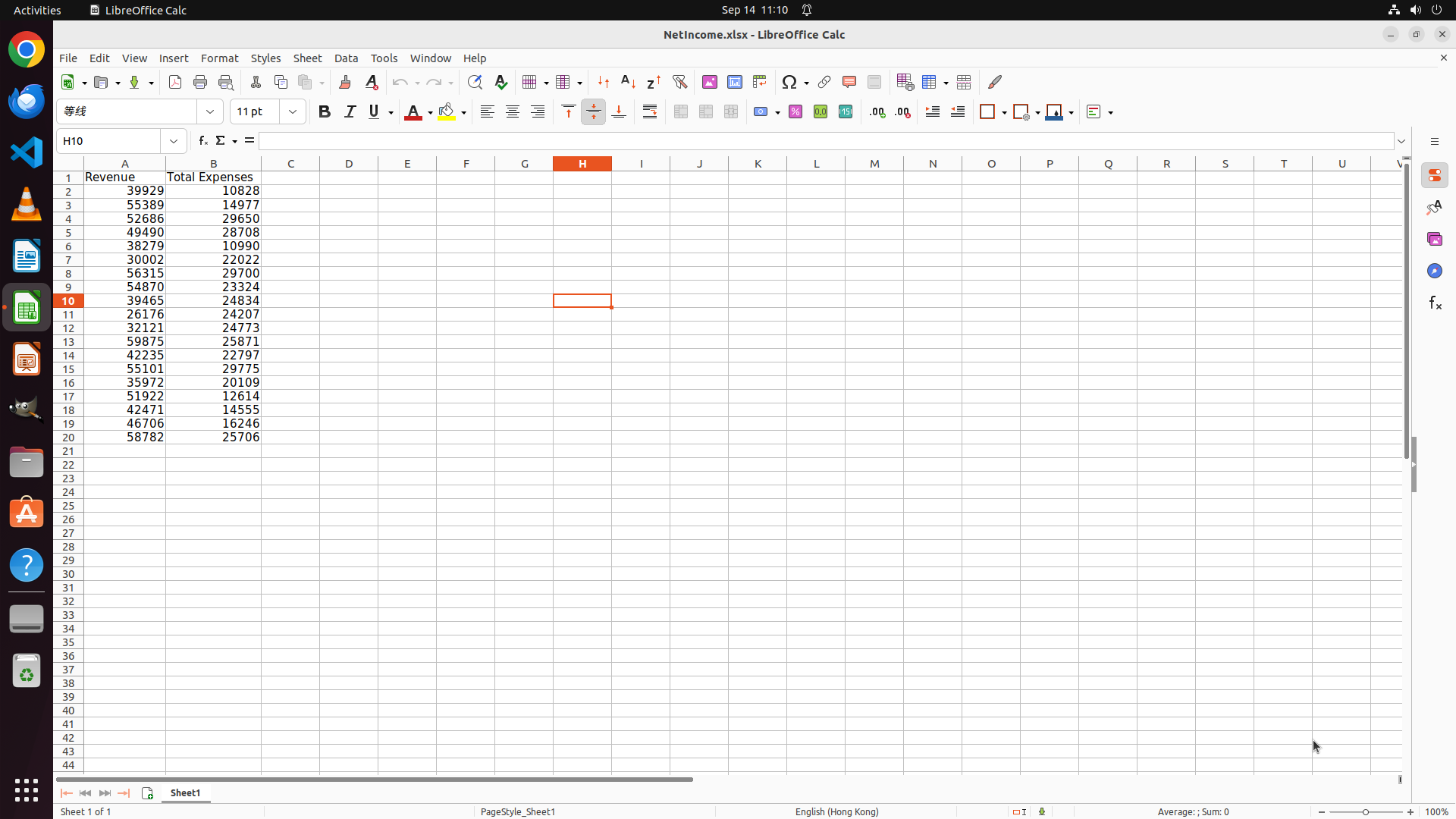}
        \end{minipage} &
        % Step 1
        \begin{minipage}{0.48\linewidth}
            \stepcaption{Step 1: {\texttt{copy\_cells\_between\_sheets(source\_range...}}}
            \includegraphics[width=\linewidth]{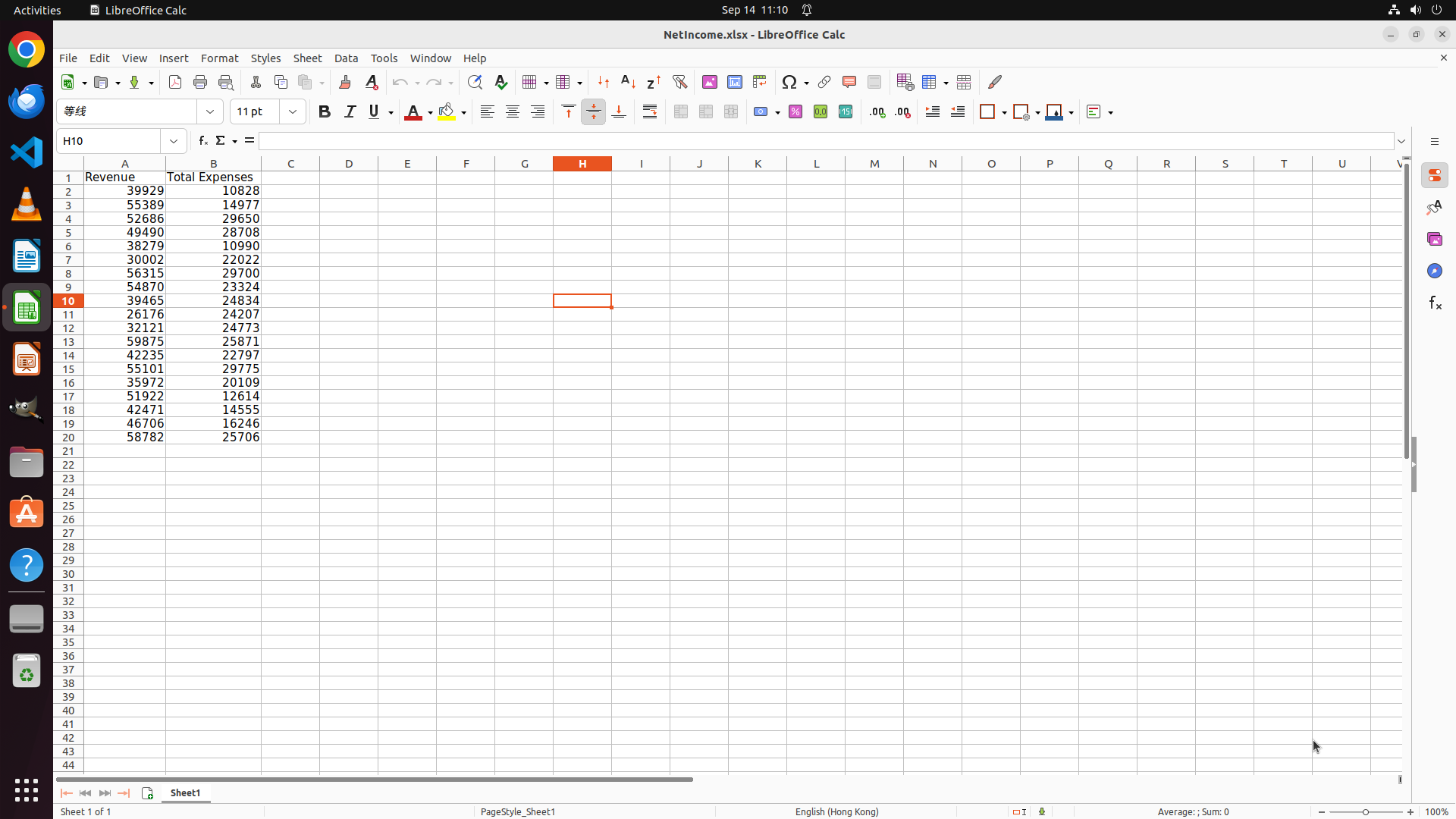}
        \end{minipage} \\
        % Step 2
        \begin{minipage}{0.48\linewidth}
            \stepcaption{Step 2: {\texttt{copy\_cells\_between\_sheets(source\_range...}}}
            \includegraphics[width=\linewidth]{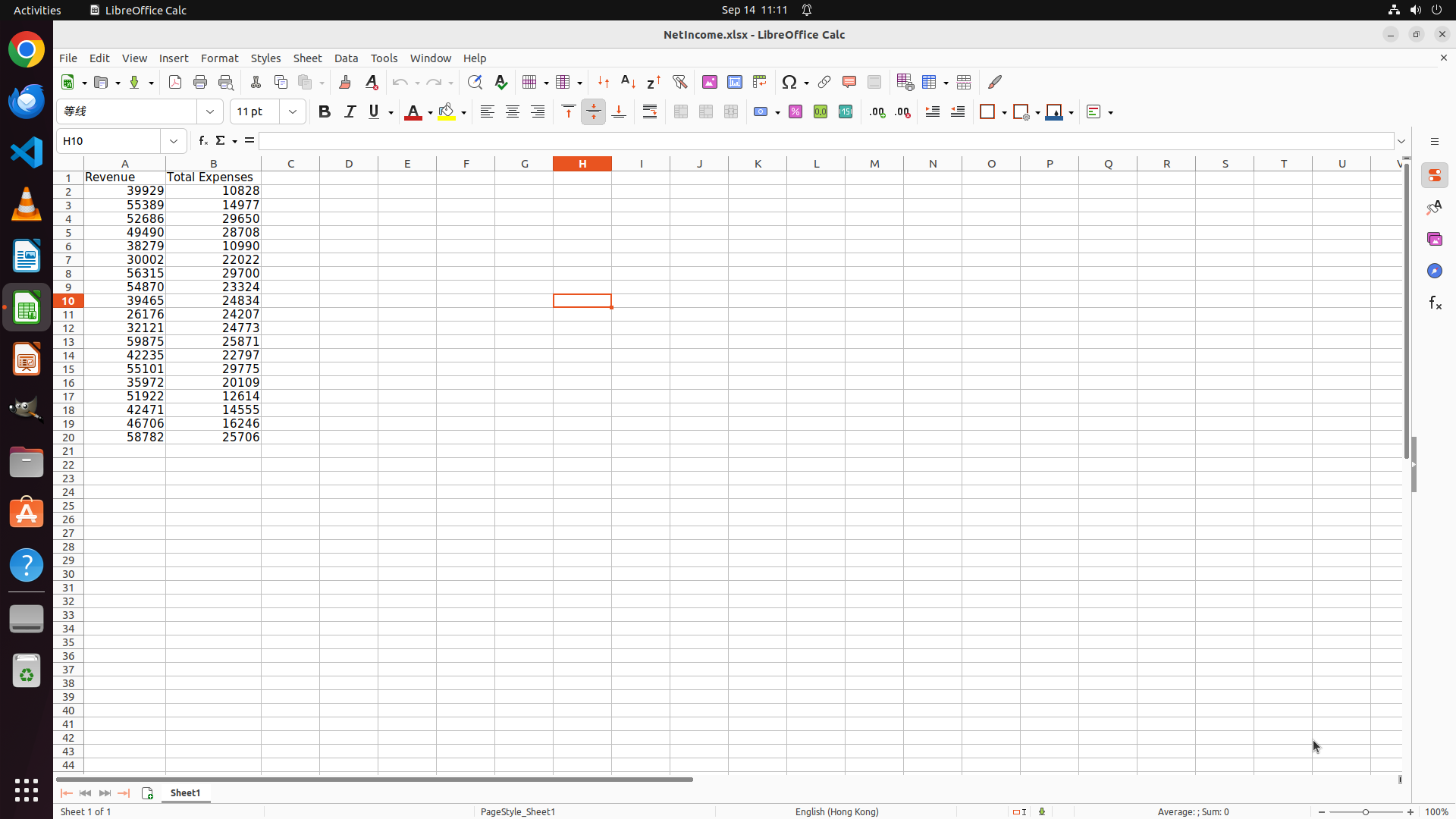}
        \end{minipage} &
        % Step 3
        \begin{minipage}{0.48\linewidth}
            \stepcaption{Step 3: {\texttt{get\_workbook\_info()}}}
            \includegraphics[width=\linewidth]{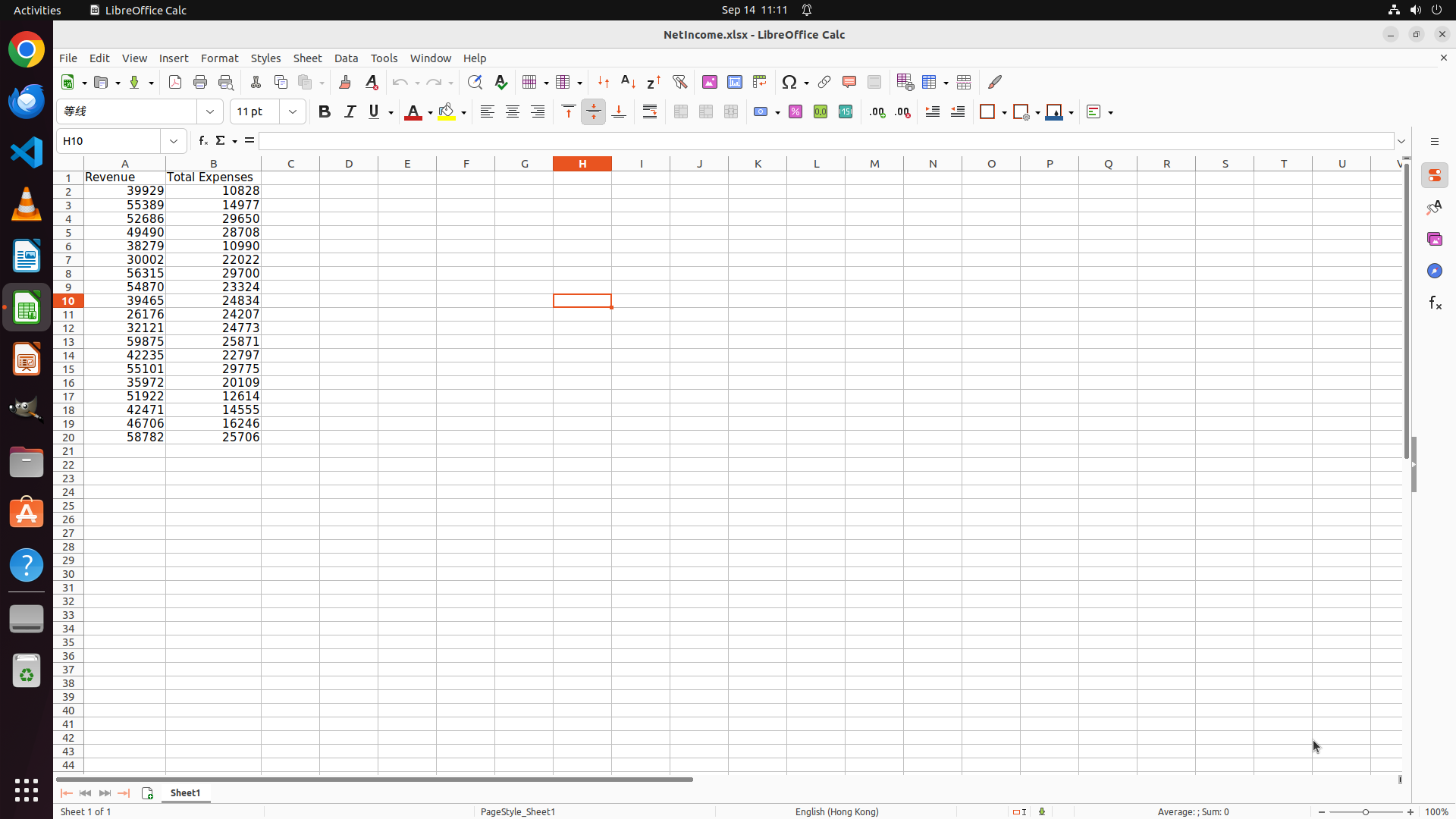}
        \end{minipage} \\
        % Step 4
        \begin{minipage}{0.48\linewidth}
            \stepcaption{Step 4: {\texttt{copy\_cells\_between\_sheets(source\_range...}}}
            \includegraphics[width=\linewidth]{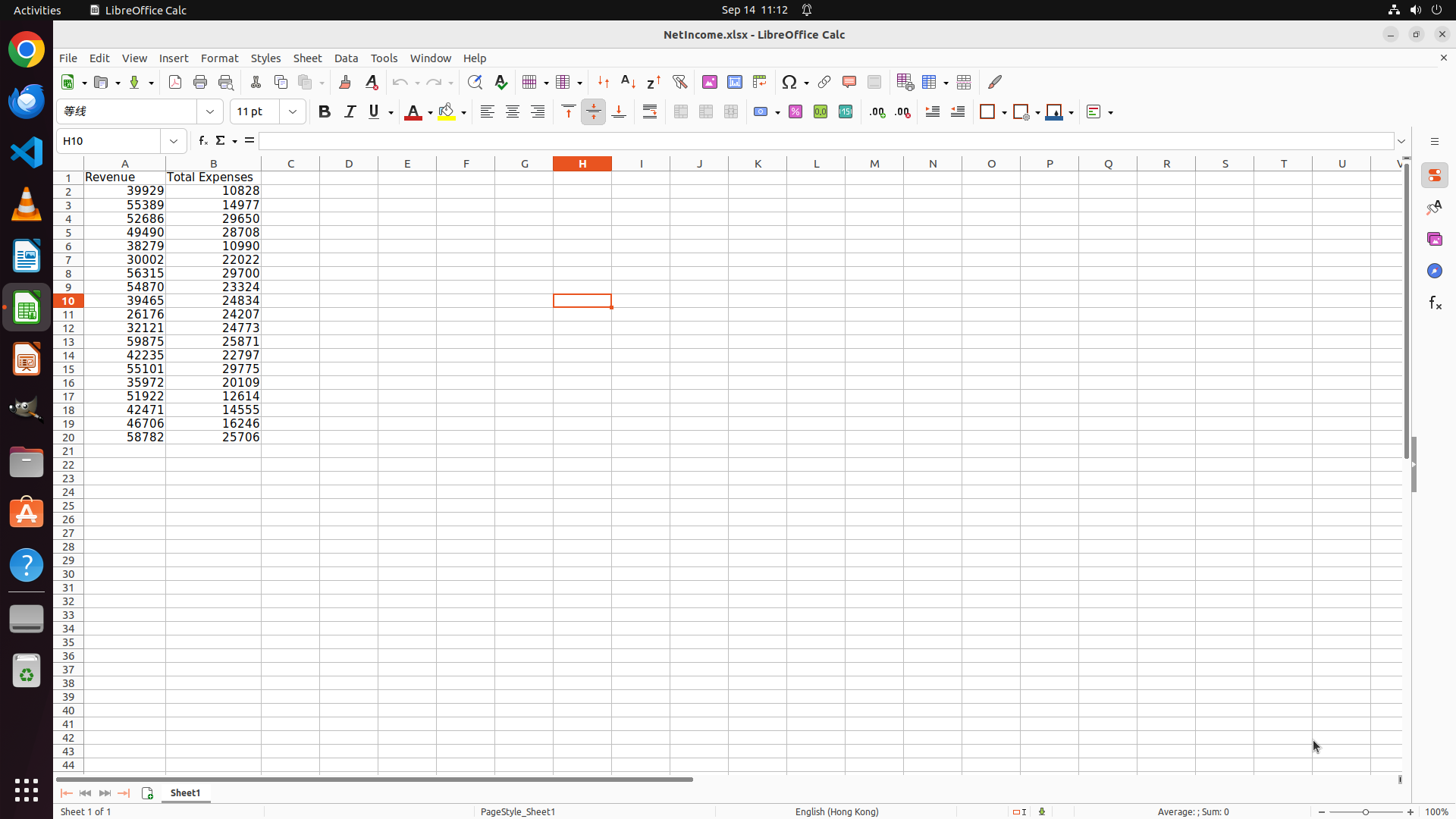}
        \end{minipage} &
        % Step 5
        \begin{minipage}{0.48\linewidth}
            \stepcaption{Step 5: {\texttt{switch\_active\_sheet(sheet\_name="Sheet2")}}}
            \includegraphics[width=\linewidth]{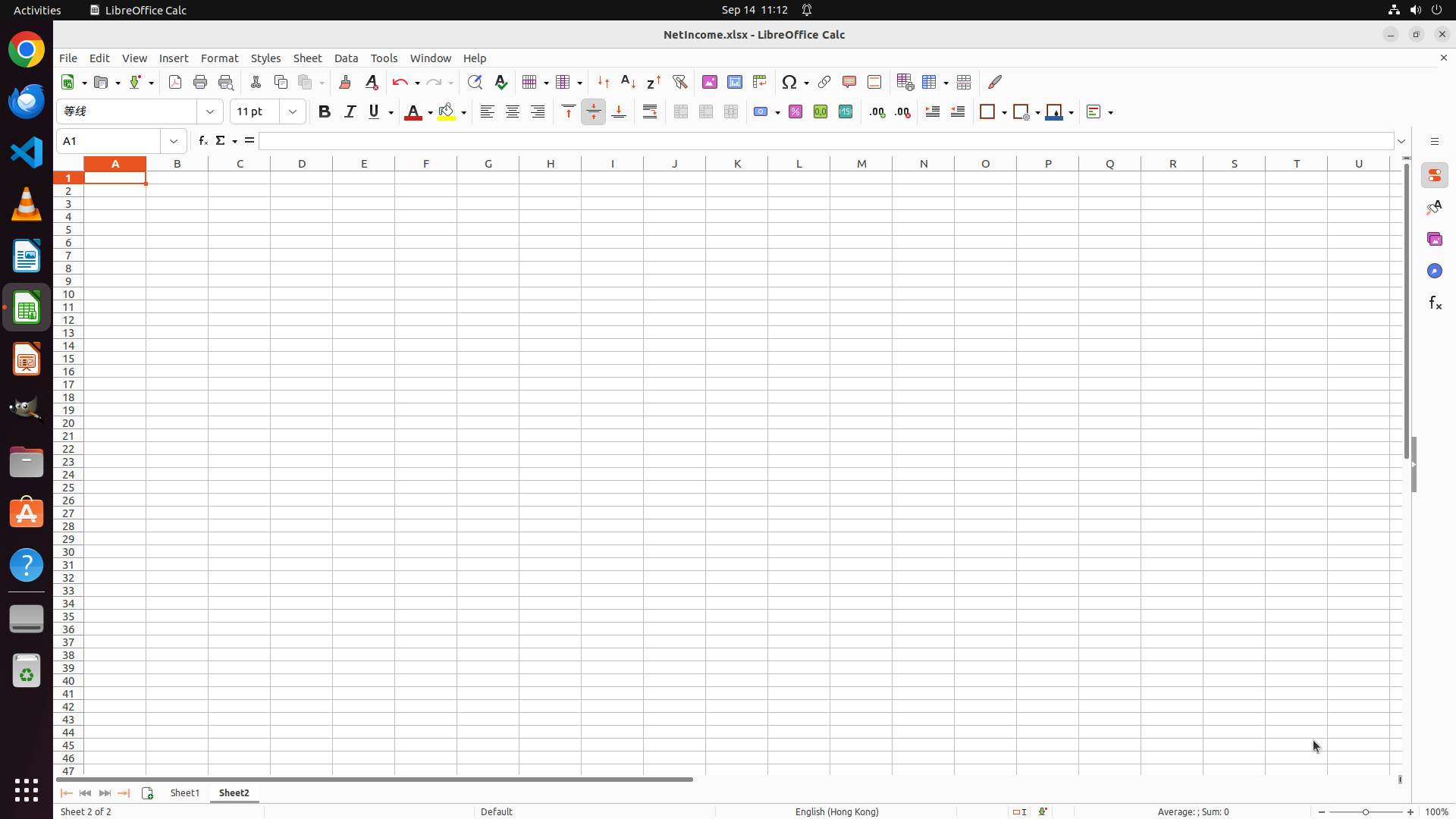}
        \end{minipage} \\
        % Step 6
        \begin{minipage}{0.48\linewidth}
            \stepcaption{Step 6: {\texttt{switch\_active\_sheet(sheet\_name="Sheet1")}}}
            \includegraphics[width=\linewidth]{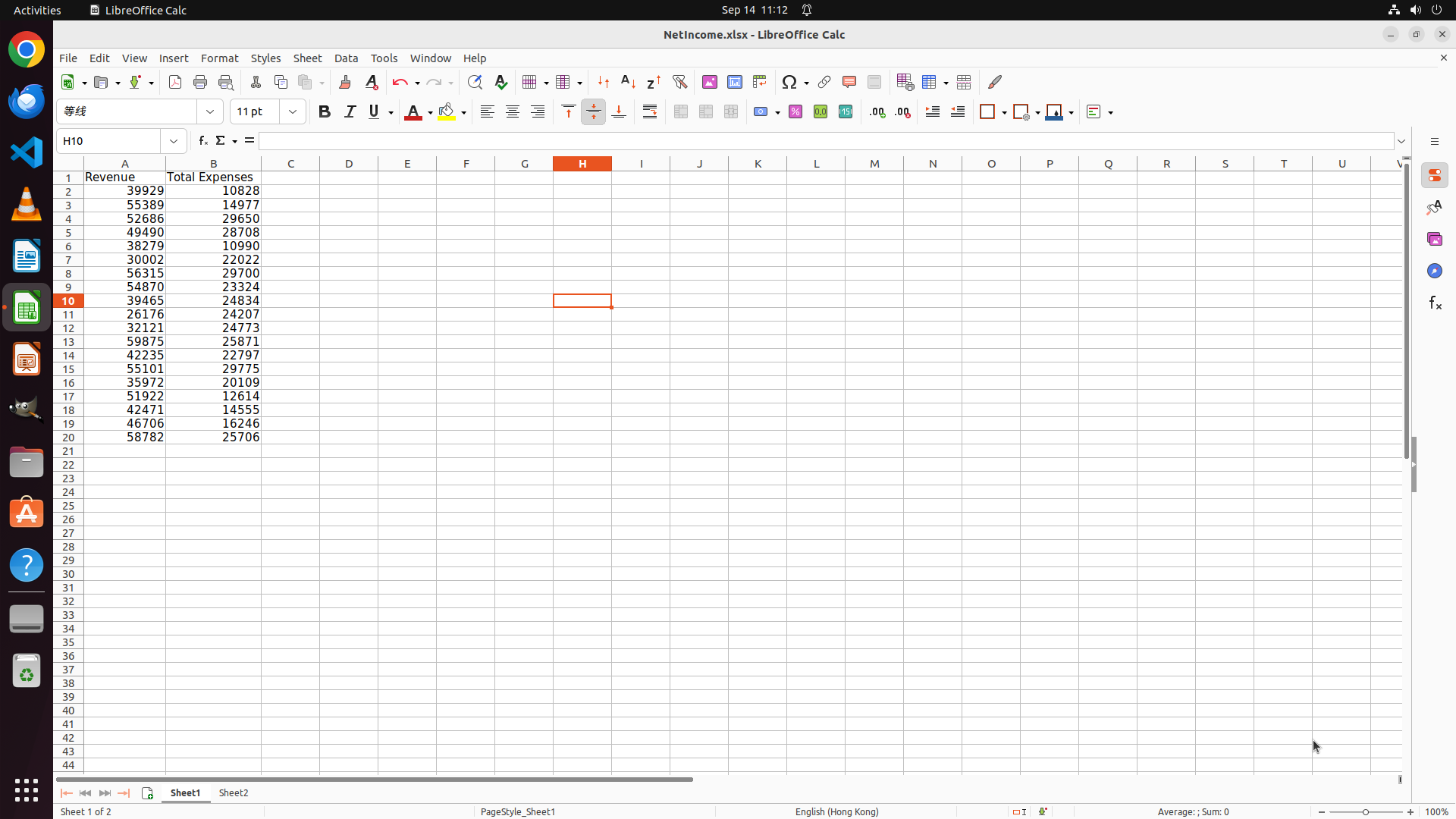}
        \end{minipage} &
        % Step 7
        \begin{minipage}{0.48\linewidth}
            \stepcaption{Step 7: {\texttt{copy\_cells\_between\_sheets(source\_range...}}}
            \includegraphics[width=\linewidth]{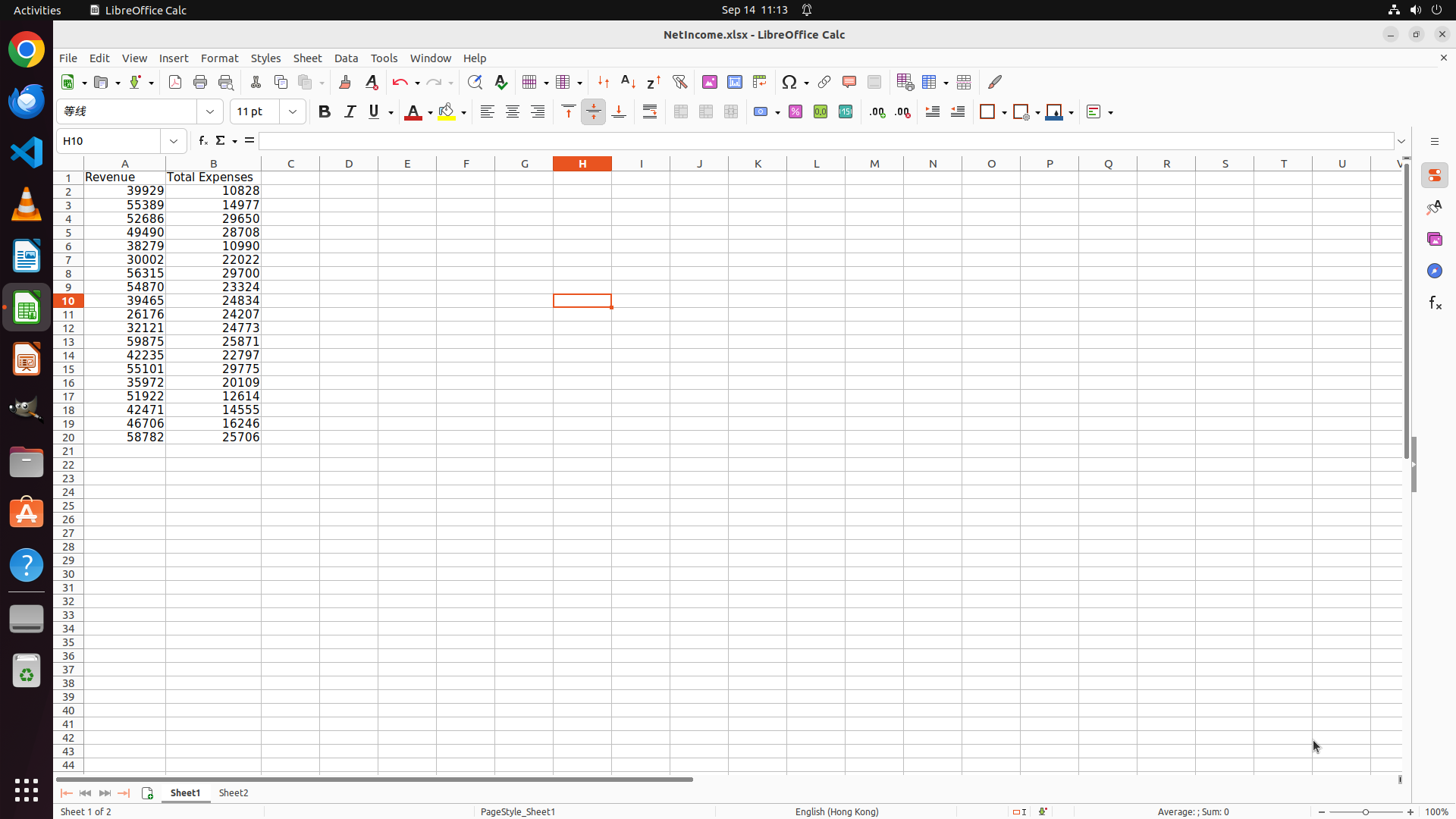}
        \end{minipage} \\
        % Step 8
        \begin{minipage}{0.48\linewidth}
            \stepcaption{Step 8: {\texttt{switch\_active\_sheet(sheet\_name="Sheet2")}}}
            \includegraphics[width=\linewidth]{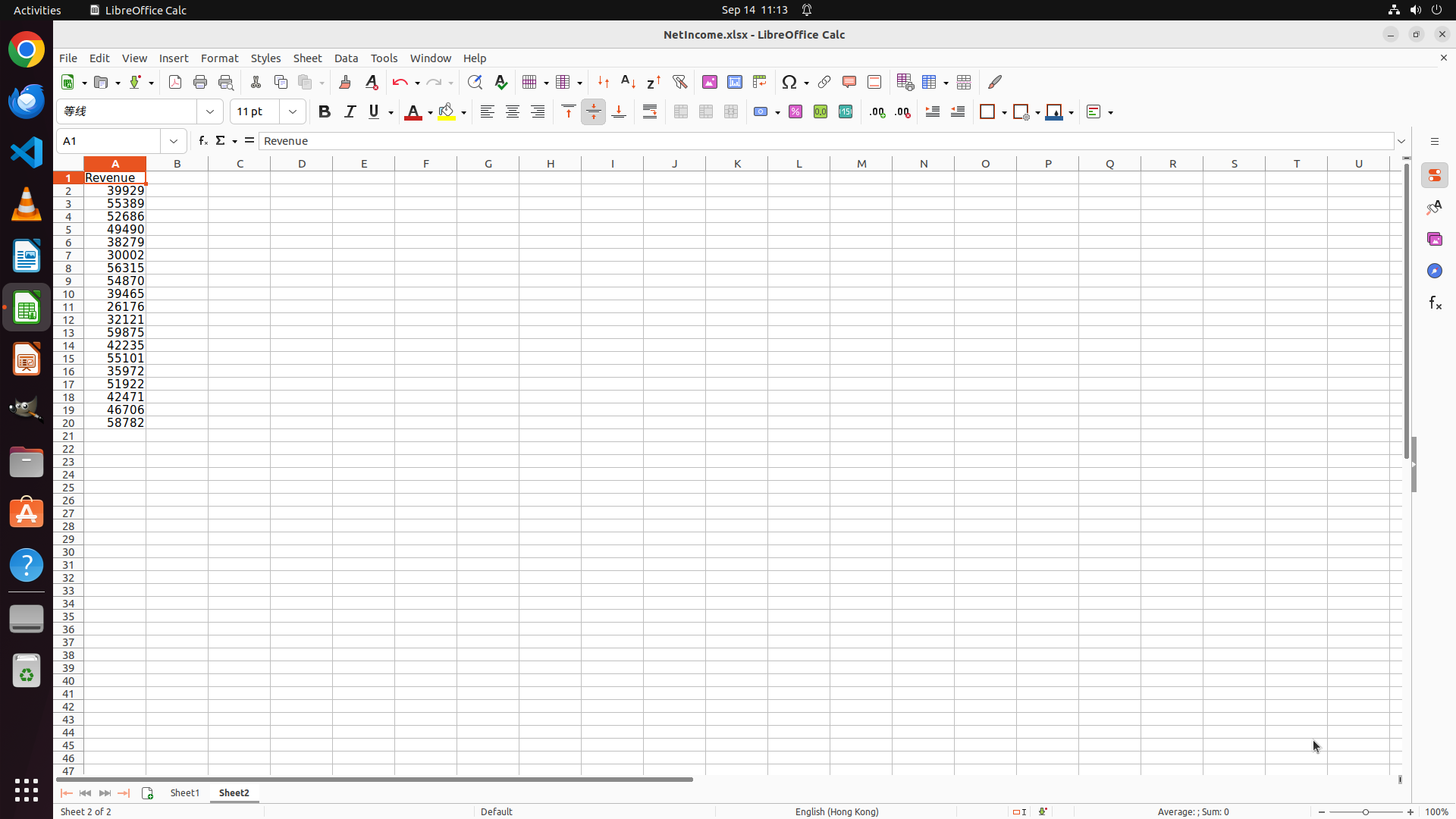}
        \end{minipage} &
        % Step 9
        \begin{minipage}{0.48\linewidth}
            \stepcaption{Step 9: {\texttt{computer\_use(action="terminate", sta...}}}
            \includegraphics[width=\linewidth]{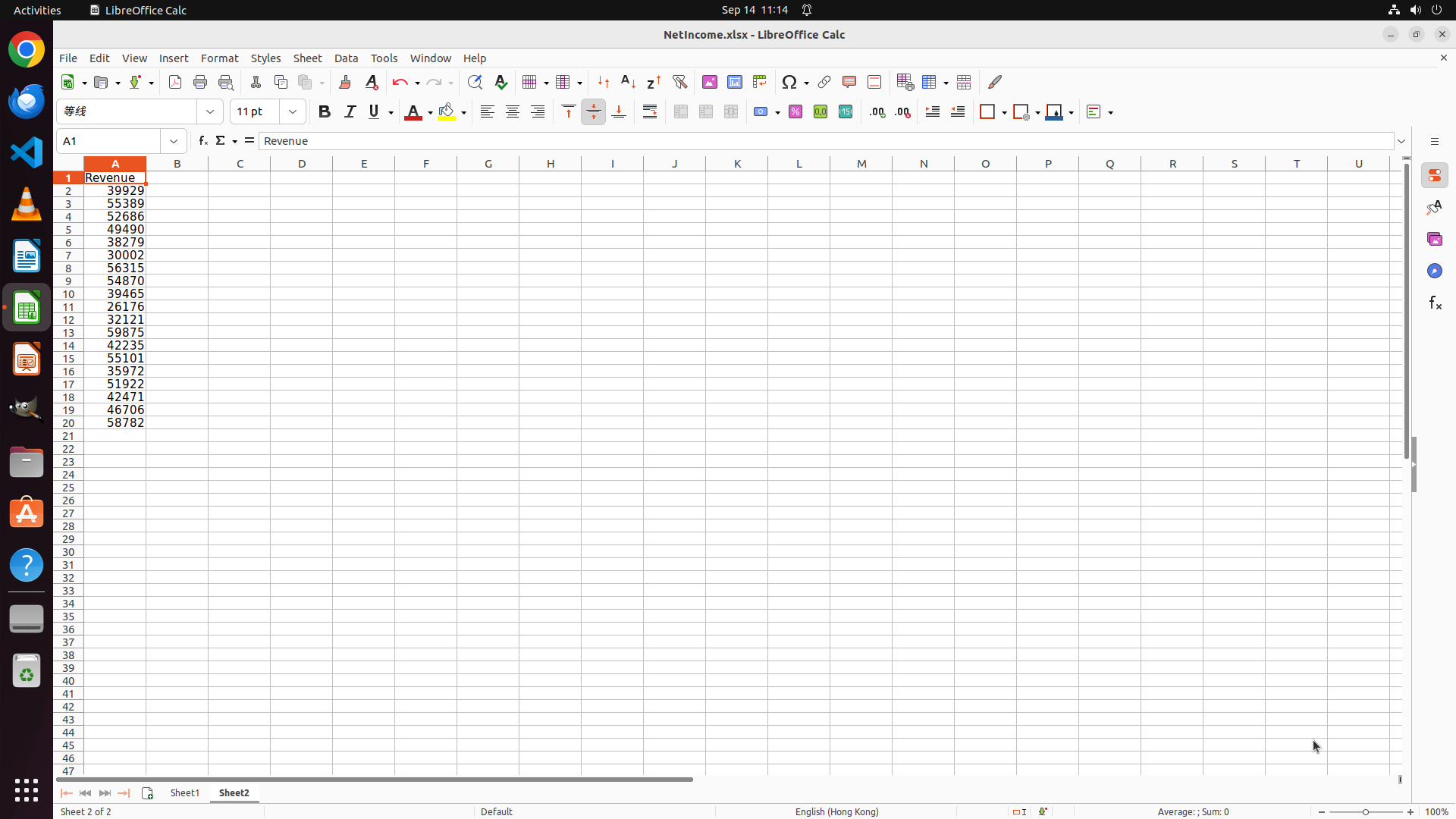}
        \end{minipage} \\
    \end{tabular}

    \caption{Copy the ``Revenue'' column along with the header to a new sheet named ``Sheet2''.}
    \label{fig:mcp-gemini-steps}
\end{figure}

According to Figure \ref{fig:mcp-gemini-steps}, the ``Revenue'' column has been copied to ``Sheet2'' successfully.

\newpage

% \subsection{More Results}
% \label{sec:more-result}

% We also conduct evaluation with a maximum step limit of 50 steps. The results are shown in Figure \ref{table:main-results-50}, with the same configuration as the 15-step case.

% \input{iclr2026/tables/main-result-50steps}
\subsection{MCP Tool List}

We present the complete list of 158 MCP tools as follows.

\begin{longtable}{cllp{6.5cm}}
\caption{Complete List of MCP Tools} \\
\toprule
\textbf{No.} & \textbf{MCP Server} & \textbf{Domain} & \textbf{MCP Tool} \\
\midrule
\endfirsthead

\toprule
\textbf{No.} & \textbf{MCP Server} & \textbf{Domain} & \textbf{MCP Tool} \\
\midrule
\endhead

\bottomrule
\endfoot

\bottomrule
\endlastfoot

1 & osworld\_mcp & VS Code & \texttt{add\_folder} \newline Adds a folder to the last active window in VSCode \\
2 & osworld\_mcp & VS Code & \texttt{compare\_files} \newline Compares two files in VSCode \\
3 & osworld\_mcp & VS Code & \texttt{disable\_extension} \newline Disables a specific extension for the next instance of VSCode \\
4 & osworld\_mcp & VS Code & \texttt{goto\_file} \newline Opens a file at a specific line and character position \\
5 & osworld\_mcp & VS Code & \texttt{install\_extension} \newline Installs an extension or updates it in VSCode \\
6 & osworld\_mcp & VS Code & \texttt{launch\_vscode} \newline Launches Visual Studio Code with the specified file path or directory \\
7 & osworld\_mcp & VS Code & \texttt{list\_extensions} \newline Lists installed extensions in VSCode \\
8 & osworld\_mcp & VS Code & \texttt{remove\_folder} \newline Removes a folder from the last active window in VSCode \\
9 & osworld\_mcp & VS Code & \texttt{toggle\_sync} \newline Toggles synchronization on or off in VSCode \\
10 & osworld\_mcp & VS Code & \texttt{uninstall\_extension} \newline Uninstalls an extension from VSCode \\
11 & osworld\_mcp & VS Code & \texttt{update\_extensions} \newline Updates all installed extensions in VSCode to the latest version \\
12 & osworld\_mcp & VS Code & \texttt{add\_files\_exclude} \newline Add a glob pattern to the Files: Exclude setting \\
13 & osworld\_mcp & VS Code & \texttt{replace\_text} \newline Open VSCode search and replace panel, input search and replacement text, and execute replacement \\
14 & osworld\_mcp & VS Code & \texttt{search\_text} \newline Open VSCode search panel, input the search term, and execute search \\
15 & osworld\_mcp & VS Code & \texttt{set\_auto\_save\_delay} \newline Set the auto save delay in milliseconds \\
16 & osworld\_mcp & VS Code & \texttt{set\_color\_theme} \newline Change the editor color theme \\
17 & osworld\_mcp & VS Code & \texttt{set\_focus\_editor\_on\_break} \newline Set the Debug: Focus Editor On Break setting to true or false \\
18 & osworld\_mcp & VS Code & \texttt{set\_python\_diagnostics\_override} \newline Override severity for a specific Python analysis diagnostic rule \\
19 & osworld\_mcp & VS Code & \texttt{set\_word\_wrap\_column} \newline Set the number of columns at which the editor will word wrap \\
20 & osworld\_mcp & VS Code & \texttt{set\_wrap\_tabs} \newline Enable or disable wrap tabs in the editor \\
21 & osworld\_mcp & Google Chrome & \texttt{bookmark\_page} \newline Bookmarks the current page in the browser (equivalent to Ctrl+D) \\
22 & osworld\_mcp & Google Chrome & \texttt{bring\_back\_last\_tab} \newline Restores the last-closed tab in the browser (equivalent to Ctrl+Shift+T) \\
23 & osworld\_mcp & Google Chrome & \texttt{chrome\_open\_tabs\_setup} \newline Opens the entered URL \\
24 & osworld\_mcp & Google Chrome & \texttt{delete\_browsing\_data} \newline Opens the 'Clear browsing data' dialog in the browser (equivalent to Ctrl+Shift+Del) \\
25 & osworld\_mcp & Google Chrome & \texttt{open\_appearance\_settings} \newline Opens the appearance settings page in the browser \\
26 & osworld\_mcp & Google Chrome & \texttt{open\_bookmarks} \newline Opens the bookmarks page in the browser \\
27 & osworld\_mcp & Google Chrome & \texttt{open\_extensions} \newline Opens the extensions management page in the browser \\
28 & osworld\_mcp & Google Chrome & \texttt{open\_password\_settings} \newline Opens the password/autofill settings page in the browser \\
29 & osworld\_mcp & Google Chrome & \texttt{open\_privacy\_settings} \newline Opens the privacy settings page in the browser \\
30 & osworld\_mcp & Google Chrome & \texttt{open\_profile\_settings} \newline Opens the profile settings page in the browser \\
31 & osworld\_mcp & Google Chrome & \texttt{open\_search\_engine\_settings} \newline Opens the search engine settings page in the browser \\
32 & osworld\_mcp & Google Chrome & \texttt{print} \newline Opens the print dialog for the current browser page (equivalent to Ctrl+P) \\
33 & osworld\_mcp & LibreOffice Calc & \texttt{adjust\_column\_width} \newline Adjust the width of specified columns \\
34 & osworld\_mcp & LibreOffice Calc & \texttt{adjust\_row\_height} \newline Adjust the height of specified rows \\
35 & osworld\_mcp & LibreOffice Calc & \texttt{copy\_sheet} \newline Create a copy of an existing worksheet in the workbook \\
36 & osworld\_mcp & LibreOffice Calc & \texttt{create\_chart} \newline Create a chart in the active worksheet based on the specified data range \\
37 & osworld\_mcp & LibreOffice Calc & \texttt{create\_pivot\_table} \newline Create a pivot table in the active worksheet based on data from the source sheet \\
38 & osworld\_mcp & LibreOffice Calc & \texttt{env\_info} \newline Get content of the specified or active sheet, including its name, headers, and data \\
39 & osworld\_mcp & LibreOffice Calc & \texttt{export\_to\_csv} \newline Export the current document to a CSV file with the same path and name as the original file \\
40 & osworld\_mcp & LibreOffice Calc & \texttt{export\_to\_pdf} \newline Export the current document or specified sheets to PDF \\
41 & osworld\_mcp & LibreOffice Calc & \texttt{format\_range} \newline Apply formatting to the specified range in the active worksheet \\
42 & osworld\_mcp & LibreOffice Calc & \texttt{freeze\_panes} \newline Freeze rows and/or columns in the active worksheet \\
43 & osworld\_mcp & LibreOffice Calc & \texttt{get\_column\_data} \newline Get all data from the specified column \\
44 & osworld\_mcp & LibreOffice Calc & \texttt{get\_workbook\_info} \newline Get workbook information, including file path, file name, sheets and active sheet \\
45 & osworld\_mcp & LibreOffice Calc & \texttt{hide\_row\_data} \newline Hide rows that contain the specified value \\
46 & osworld\_mcp & LibreOffice Calc & \texttt{highlight\_range} \newline Highlight the specified range with the specified color \\
47 & osworld\_mcp & LibreOffice Calc & \texttt{merge\_cells} \newline Merge cells in the specified range \\
48 & osworld\_mcp & LibreOffice Calc & \texttt{rename\_sheet} \newline Rename a worksheet in the workbook \\
49 & osworld\_mcp & LibreOffice Calc & \texttt{reorder\_columns} \newline Reorder the columns in the sheet according to the specified order \\
50 & osworld\_mcp & LibreOffice Calc & \texttt{reorder\_sheets} \newline Change the order of worksheets in the workbook \\
51 & osworld\_mcp & LibreOffice Calc & \texttt{save} \newline Save the current workbook to its current location \\
52 & osworld\_mcp & LibreOffice Calc & \texttt{set\_cell\_value} \newline Set a value to a specific cell in the active worksheet \\
53 & osworld\_mcp & LibreOffice Calc & \texttt{set\_column\_values} \newline Set values to the specified column, cannot be used to set formulas \\
54 & osworld\_mcp & LibreOffice Calc & \texttt{set\_number\_format} \newline Apply a specific number format to a range of cells \\
55 & osworld\_mcp & LibreOffice Calc & \texttt{set\_zoom\_level} \newline Adjust the zoom level of the current worksheet \\
56 & osworld\_mcp & LibreOffice Calc & \texttt{sort\_column} \newline Sort the data in the specified column in ascending or descending order \\
57 & osworld\_mcp & LibreOffice Calc & \texttt{switch\_active\_sheet} \newline Switch to the specified sheet and make it active. Creates new sheet if it doesn't exist \\
58 & osworld\_mcp & LibreOffice Calc & \texttt{transpose\_range} \newline Transpose the specified range and paste it to the target cell \\
59 & osworld\_mcp & LibreOffice Calc & \texttt{copy\_cells\_between\_sheets} \newline Copy cells from a specified rectangular source range to another sheet \\
60 & osworld\_mcp & LibreOffice Calc & \texttt{fill\_blank\_down} \newline Forward-fills blank cells in specified columns with value from cell above \\
61 & osworld\_mcp & LibreOffice Calc & \texttt{format\_numbers\_to\_human\_readable} \newline Convert numeric values to human-readable strings (M for millions, B for billions) \\
62 & osworld\_mcp & LibreOffice Calc & \texttt{scale\_first\_sheet\_and\_export\_pdf} \newline Scales the first sheet to specified pages and exports to PDF \\
63 & osworld\_mcp & LibreOffice Writer & \texttt{save} \newline Save the current document to its current location \\
64 & osworld\_mcp & LibreOffice Writer & \texttt{write\_text} \newline Write text at the current cursor position in the document \\
65 & osworld\_mcp & LibreOffice Writer & \texttt{set\_color} \newline Changes the color of matched text in the document \\
66 & osworld\_mcp & LibreOffice Writer & \texttt{find\_and\_replace} \newline Finds and replaces text in the document \\
67 & osworld\_mcp & LibreOffice Writer & \texttt{set\_font} \newline Changes the font of text in the document \\
68 & osworld\_mcp & LibreOffice Writer & \texttt{set\_line\_spacing} \newline Sets the line spacing for specified paragraphs \\
69 & osworld\_mcp & LibreOffice Writer & \texttt{insert\_formula\_at\_cursor} \newline Inserts a formula at the current cursor position \\
70 & osworld\_mcp & LibreOffice Writer & \texttt{insert\_image\_at\_cursor} \newline Inserts an image at the current cursor position \\
71 & osworld\_mcp & LibreOffice Writer & \texttt{set\_font\_size} \newline Changes the font size of specified text \\
72 & osworld\_mcp & LibreOffice Writer & \texttt{export\_to\_pdf} \newline Exports the current document to PDF format \\
73 & osworld\_mcp & LibreOffice Writer & \texttt{set\_paragraph\_alignment} \newline Sets the text alignment for specified paragraphs \\
74 & osworld\_mcp & LibreOffice Writer & \texttt{set\_default\_font} \newline Sets the default font for new text without changing existing text \\
75 & osworld\_mcp & LibreOffice Writer & \texttt{add\_page\_numbers} \newline Adds page numbers to the document at specified position \\
76 & osworld\_mcp & LibreOffice Writer & \texttt{insert\_page\_break} \newline Inserts a page break at current cursor position \\
77 & osworld\_mcp & LibreOffice Writer & \texttt{env\_info} \newline Retrieve all paragraphs, truncate each to at most 500 characters \\
78 & osworld\_mcp & LibreOffice Impress & \texttt{configure\_auto\_save} \newline Enables or disables auto-save functionality \\
79 & osworld\_mcp & LibreOffice Impress & \texttt{delete\_content} \newline Deletes the specified textbox from a slide \\
80 & osworld\_mcp & LibreOffice Impress & \texttt{duplicate\_slide} \newline Creates a duplicate of a specific slide \\
81 & osworld\_mcp & LibreOffice Impress & \texttt{env\_info} \newline Get the content of the specified pages \\
82 & osworld\_mcp & LibreOffice Impress & \texttt{export\_to\_image} \newline Exports the current presentation or a specific slide to an image file \\
83 & osworld\_mcp & LibreOffice Impress & \texttt{get\_slide\_count} \newline Gets the total number of slides in the current presentation \\
84 & osworld\_mcp & LibreOffice Impress & \texttt{go\_to\_slide} \newline Navigates to a specific slide in the presentation \\
85 & osworld\_mcp & LibreOffice Impress & \texttt{insert\_file} \newline Insects a video or audio file into the current or specified slide \\
86 & osworld\_mcp & LibreOffice Impress & \texttt{insert\_image} \newline Inserts an image to a specific slide \\
87 & osworld\_mcp & LibreOffice Impress & \texttt{position\_box} \newline Positions a textbox or image on a slide at a specific location \\
88 & osworld\_mcp & LibreOffice Impress & \texttt{save} \newline Save the current presentation to its current location \\
89 & osworld\_mcp & LibreOffice Impress & \texttt{save\_as} \newline Saves the current document to a specified location \\
90 & osworld\_mcp & LibreOffice Impress & \texttt{set\_background\_color} \newline Sets the background color for the specified textbox \\
91 & osworld\_mcp & LibreOffice Impress & \texttt{set\_slide\_background} \newline Sets the background color or image for a specific slide or all slides \\
92 & osworld\_mcp & LibreOffice Impress & \texttt{set\_slide\_font} \newline Sets the font style for all text elements in a specific slide \\
93 & osworld\_mcp & LibreOffice Impress & \texttt{set\_slide\_orientation} \newline Changes the orientation of slides between portrait and landscape \\
94 & osworld\_mcp & LibreOffice Impress & \texttt{set\_style} \newline Sets the style properties for the specified textbox \\
95 & osworld\_mcp & LibreOffice Impress & \texttt{set\_text\_color} \newline Sets the text color for the specified textbox \\
96 & osworld\_mcp & LibreOffice Impress & \texttt{set\_text\_strikethrough} \newline Applies or removes strike-through formatting to text \\
97 & osworld\_mcp & LibreOffice Impress & \texttt{set\_textbox\_alignment} \newline Sets the text alignment for the specified textbox \\
98 & osworld\_mcp & LibreOffice Impress & \texttt{write\_text} \newline writes text to a specific textbox on a slide \\
99 & osworld\_mcp & LibreOffice Impress & \texttt{convert\_to\_docx} \newline Transfers all text slide-by-slide into a new Writer document \\
100 & osworld\_mcp & OS & \texttt{change\_text\_scale} \newline Changes the text-scaling factor and returns the previous value \\
101 & osworld\_mcp & OS & \texttt{configure\_auto\_lock} \newline Configures the GNOME automatic-lock behaviour \\
102 & osworld\_mcp & OS & \texttt{copy\_matching\_files\_with\_hierarchy} \newline Copies all files matching a pattern preserving directory hierarchy \\
103 & osworld\_mcp & OS & \texttt{get\_do\_not\_disturb\_status} \newline Returns the current Do Not Disturb state \\
104 & osworld\_mcp & OS & \texttt{get\_text\_scale} \newline Returns the current GNOME text-scaling factor \\
105 & osworld\_mcp & OS & \texttt{get\_trash\_directory} \newline Returns the absolute path to the user Trash 'files' directory \\
106 & osworld\_mcp & OS & \texttt{get\_volume} \newline Returns the current output volume (percentage) \\
107 & osworld\_mcp & OS & \texttt{open\_shell} \newline Open a new terminal directly \\
108 & osworld\_mcp & OS & \texttt{rename\_directory} \newline Safely renames a folder on the local filesystem \\
109 & osworld\_mcp & OS & \texttt{restore\_file} \newline Restore the specified file from the user Trash back to its original location \\
110 & osworld\_mcp & OS & \texttt{search\_files} \newline Recursively search all files under the given folder \\
111 & osworld\_mcp & OS & \texttt{set\_default\_terminal\_size} \newline Persists the given number of columns × rows as the default GNOME-Terminal window size \\
112 & osworld\_mcp & OS & \texttt{set\_do\_not\_disturb} \newline Enable or disable the GNOME/Ubuntu Do Not Disturb mode \\
113 & osworld\_mcp & OS & \texttt{set\_volume} \newline Sets the output volume of the default PulseAudio / PipeWire sink \\
114 & osworld\_mcp & OS & \texttt{convert\_image\_format} \newline Convert an input image file to a specified format \\
115 & osworld\_mcp & OS & \texttt{ffmpeg\_video\_to\_gif} \newline Extract a portion of a video file and save it as an animated GIF \\
116 & osworld\_mcp & OS & \texttt{git\_operation} \newline Execute a git command (clone, add, commit, pull, push, etc.) \\
117 & osworld\_mcp & OS & \texttt{git\_set\_user\_info} \newline Configure git user.name and user.email \\
118 & osworld\_mcp & OS & \texttt{remove\_image\_background} \newline Remove the background of an input image and save with transparency \\
119 & osworld\_mcp & OS & \texttt{calculator} \newline Simple calculator. \\
120 & osworld\_mcp & VLC & \texttt{add\_to\_playlist} \newline Adds a media file to the VLC playlist \\
121 & osworld\_mcp & VLC & \texttt{get\_current\_time} \newline Gets the current playback time position in seconds \\
122 & osworld\_mcp & VLC & \texttt{get\_media\_duration} \newline Gets the total duration of the currently playing media file in seconds \\
123 & osworld\_mcp & VLC & \texttt{get\_media\_files} \newline Gets the media files for the specified path \\
124 & osworld\_mcp & VLC & \texttt{get\_playlist} \newline Gets the current VLC playlist with track information \\
125 & osworld\_mcp & VLC & \texttt{get\_settings} \newline Gets the current settings of the VLC player \\
126 & osworld\_mcp & VLC & \texttt{next} \newline Switches to the next media item in the VLC playlist \\
127 & osworld\_mcp & VLC & \texttt{pause} \newline Pauses the currently playing media in VLC player \\
128 & osworld\_mcp & VLC & \texttt{play} \newline Starts playing the current media in VLC player \\
129 & osworld\_mcp & VLC & \texttt{previous} \newline Switches to the previous media item in the VLC playlist \\
130 & osworld\_mcp & VLC & \texttt{set\_settings} \newline Sets the settings for the VLC player \\
131 & osworld\_mcp & VLC & \texttt{toggle\_fullscreen} \newline Toggles fullscreen mode for the currently playing video \\
132 & filesystem & OS & \texttt{read\_file} \newline Read the complete contents of a file as text (DEPRECATED) \\
133 & filesystem & OS & \texttt{read\_text\_file} \newline Read the complete contents of a file as text with encoding handling \\
134 & filesystem & OS & \texttt{read\_media\_file} \newline Read an image or audio file as base64 encoded data \\
135 & filesystem & OS & \texttt{read\_multiple\_files} \newline Read the contents of multiple files simultaneously \\
136 & filesystem & OS & \texttt{write\_file} \newline Create a new file or completely overwrite an existing file \\
137 & filesystem & OS & \texttt{edit\_file} \newline Make line-based edits to a text file \\
138 & filesystem & OS & \texttt{create\_directory} \newline Create a new directory or ensure a directory exists \\
139 & filesystem & OS & \texttt{list\_directory} \newline Get a detailed listing of all files and directories in a specified path \\
140 & filesystem & OS & \texttt{list\_directory\_with\_sizes} \newline Get a detailed listing with sizes of all files and directories \\
141 & filesystem & OS & \texttt{directory\_tree} \newline Get a recursive tree view of files and directories as JSON \\
142 & filesystem & OS & \texttt{move\_file} \newline Move or rename files and directories \\
143 & filesystem & OS & \texttt{search\_files} \newline Recursively search for files and directories matching a pattern \\
144 & filesystem & OS & \texttt{get\_file\_info} \newline Retrieve detailed metadata about a file or directory \\
145 & filesystem & OS & \texttt{list\_allowed\_directories} \newline Returns the list of directories that this server is allowed to access \\
146 & git & OS & \texttt{git\_status} \newline Shows the working tree status \\
147 & git & OS & \texttt{git\_diff\_unstaged} \newline Shows changes in the working directory that are not yet staged \\
148 & git & OS & \texttt{git\_diff\_staged} \newline Shows changes that are staged for commit \\
149 & git & OS & \texttt{git\_diff} \newline Shows differences between branches or commits \\
150 & git & OS & \texttt{git\_commit} \newline Records changes to the repository \\
151 & git & OS & \texttt{git\_add} \newline Adds file contents to the staging area \\
152 & git & OS & \texttt{git\_reset} \newline Unstages all staged changes \\
153 & git & OS & \texttt{git\_log} \newline Shows the commit logs \\
154 & git & OS & \texttt{git\_create\_branch} \newline Creates a new branch from an optional base branch \\
155 & git & OS & \texttt{git\_checkout} \newline Switches branches \\
156 & git & OS & \texttt{git\_show} \newline Shows the contents of a commit \\
157 & git & OS & \texttt{git\_init} \newline Initialize a new Git repository \\
158 & git & OS & \texttt{git\_branch} \newline List Git branches
\label{table:complete-mcp-tools}
\end{longtable}

\subsection{Use of LLMs}
Large language models (LLMs) are used solely to assist in the preparation of this manuscript. They help improve the clarity, coherence, and conciseness of the text, refine phrasing, and ensure that the language conforms to academic writing standards. All conceptual content, experimental design, data analysis, and conclusions are developed entirely by the authors.

\end{document}